\documentclass{article}
\usepackage{arxiv}

\title{\LARGE\bf Nonparametric Instrumental Regression via Kernel Methods is Minimax Optimal\vspace{1em}}

\author{ 
 Dimitri Meunier\footnotemark[1] $^{,}$\footnotemark[4] \\ {\footnotesize\em dimitri.meunier.21@ucl.ac.uk} \\ \and Zhu Li\thanks{Equal Contribution.} $^,$\thanks{Mingdu Technology, Hangzhou} \\ {\footnotesize\em liz@mingdutech.com} \and Timothy Christensen \thanks{Department of Economics, Yale University, New Haven.} \\{\footnotesize\em timothy.christensen@yale.edu}  \and  Arthur Gretton\thanks{Gatsby Computational Neuroscience Unit, University College London, London.} \\ {\footnotesize\em arthur.gretton@gmail.com} \\ $ $ \\  
}
\date{}
\begin{document}

\maketitle

\begin{abstract}
We study the \emph{kernel instrumental variable} (KIV) algorithm of \citet{singh2019kernel}, a kernel-based two-stage least-squares method for nonparametric instrumental variable regression. We provide a convergence analysis covering both identified and non-identified regimes: when the structural function is not identified, we show that the KIV estimator converges to the minimum-norm IV solution in the reproducing kernel Hilbert space associated with the kernel. Crucially, we establish convergence in the strong $L_2$ norm, rather than only in a pseudo-norm.
We quantify statistical difficulty through a link condition that compares the covariance structure of the endogenous regressor with that induced by the instrument, yielding an interpretable measure of ill-posedness.
Under standard eigenvalue-decay and source assumptions, we derive strong $L_2$ learning rates for KIV and prove that they are minimax-optimal over fixed smoothness classes.
Finally, we replace the stage-1 Tikhonov step by general spectral regularization, thereby avoiding saturation and improving rates for smoother first-stage targets.  The matching lower bound shows that instrumental regression induces an unavoidable slowdown relative to ordinary kernel ridge regression.
\end{abstract}


\section{Introduction}
We consider the \emph{nonparametric instrumental variable} (NPIV) estimation setting \citep{newey2003instrumental,ai2003efficient}. The problem is defined in terms of four random variables $X$, $Y$, $Z$, and $U$, where $X$ represents the endogenous variables, $Y$ is an outcome, $Z$ denotes instruments, and $U$ represents unmeasured confounding, so that $X$ may be correlated with $U$.  The model is
\begin{equation} \label{eq:npiv}
 Y = h_0(X) + U,~~~ \mathbb{E}\left[U|Z\right] = 0, 
\end{equation}
where $h_0$ is the  structural function, assumed to belong to the Hilbert space $L_2(X)$, that is, to be square-integrable with respect to the marginal distribution of $X$. Equivalently, NPIV can be characterized through the functional equation \citep{darolles2011nonparametric}
\begin{equation} \label{eq:npiv_integral}
\mathcal{T}h_0 = r_0,
\end{equation}
where $r_0(Z) \doteq \E [Y|Z]$ and $\mathcal{T}: L_2(X) \rightarrow L_2(Z)$ is the bounded linear operator mapping $h \in L_2(X)$ to $\E[h(X)| Z] \in L_2(Z)$. NPIV estimation can therefore be viewed as the problem of estimating the solution to this inverse problem from data.

\medskip \noindent
NPIV estimation has many applications, such as causal inference \citep{newey2003instrumental}, demand estimation in markets for differentiated products \citep{compiani2022market}, analysis of household spending \citep{blundell2007semi}, consumer demand and welfare analysis \citep{chen2018optimal}, missing data problems \citep{wang2014instrumental,sun2018semiparametric} and reinforcement learning \citep{uehara2021finite,xu2020learning,chen2022ivForPolicyEval,liao2024instrumental}. At the same time, NPIV is challenging for two distinct reasons. First, the structural function $h_0$ is identified if and only if the operator $\cT$ is injective \citep{newey2003instrumental}, an assumption often imposed in the literature, commonly referred to as completeness \citep{darolles2011nonparametric,chen2011rate,singh2019kernel}. Although completeness holds for a broad class of distributions \citep{andrews2017examples}, it may fail in practice and can also be difficult to test: \citet{canay2013testability} establish a general non-testability result for completeness, whereas \citet{freyberger2017completeness} obtains positive testability results under additional restrictions. Second, even when the solution is unique, NPIV remains an ill-posed inverse problem: small perturbations of the outcome can lead to arbitrarily large errors in estimating $h_0$, since $\cT^{-1}$ is typically not continuous \citep{carrasco2007linear}.

\medskip \noindent
In this paper, we study the kernel instrumental variable (KIV) estimator of \citet{singh2019kernel}, from the viewpoint of strong $L_2(X)$ risk. KIV is appealing in practice: it is stable, easy to implement, and empirically competitive with a range of existing NPIV methods, including Nadaraya--Watson IV \citep{carrasco2007linear,darolles2011nonparametric} and sieve-based IV \citep{chen2018optimal,newey2003instrumental}. However, existing analyses of KIV leave three central gaps: they assume identification, they control only a pseudo-metric, and they do not cleanly isolate how the instrument-induced ill-posedness enters the learning rate.


\medskip \noindent
We address these issues through the following contributions:
\begin{itemize}[itemsep=0pt]
\item \textbf{Strong $L_2(X)$ theory without identification.} We work with the minimum-norm solution of the IV equation in the underlying reproducing kernel Hilbert space (RKHS) as a canonical target, and show that KIV converges to this target in strong $L_2(X)$ norm, both in identified and non-identified regimes.

    \item \textbf{An interpretable notion of ill-posedness.} In two-stage least squares, Stage~2 does not learn from the endogenous regressor directly, but from features obtained by conditioning on the instrument. This conditioning smooths the signal, making estimation in the strong norm an ill-posed problem. We formalize this effect in a way that makes explicit how instrument strength shapes statistical rates, and yields smoothness assumptions that are easier to interpret than source conditions stated directly in terms of $\cT$ (as in \citet{singh2019kernel}).

    \item \textbf{Minimax-optimal strong-norm rates.} Under standard eigenvalue-decay and source assumptions, we derive strong $L_2(X)$ rates for KIV, and complement them with a matching minimax lower bound over fixed smoothness classes. These results show that instrumental regression induces an unavoidable penalty relative to ordinary kernel ridge regression. The lower bound shows that this penalty is not an artifact of the analysis or of KIV itself, but a fundamental feature of NPIV.

    \item \textbf{Spectral Regularization.} We replace the Tikhonov regularization used by \citet{singh2019kernel} with general spectral regularization in Stage 1. This avoids saturation and yields faster rates for smoother first-stage targets.

\end{itemize}

\subsection{Related Works}

NPIV is an important estimation problem and the subject of a substantial number of studies. In the econometrics literature, NPIV solutions are historically obtained using series-based estimators or estimators based on kernel density estimation \citep{newey2003instrumental,hall2005nonparametric,blundell2007semi,chen2007large,darolles2011nonparametric,florens2011identification,horowitz2011applied,chen2012estimation}. \citet{chen2011rate,chen2018optimal} demonstrate that these estimators can achieve the minimax optimal rate. However, kernel-density and series-based estimators are difficult to train and have suboptimal empirical performance \citep{singh2019kernel}. As a result, recent developments in NPIV employ modern machine learning methods to address such problems, including kernel-based estimators \citep{singh2019kernel,zhang2023instrumental} and neural network-based estimators \citep{hartford2017deep,bennett2019deep,xu2020learning,petrulionyte2024functional,sun2025spectral,meunier2025demystifying,meunier2025outcome,bruns2025two}. 

Modern NPIV learning methods can be largely categorized into two classes: \emph{two-stage estimation methods} \citep{hartford2017deep,singh2019kernel,xu2020learning,li2024regularized,petrulionyte2024functional} and \emph{min--max optimization methods} \citep{bennett2019deep,dikkala2020minimax,liao2020provably,bennett2023minimax,zhang2023instrumental,chen2024stochastic}. Min--max optimization methods consider a saddle-point problem of the form $\min_{h \in \cH} \max_{g \in \cG} \mathcal{L}(h, g)$ for some function classes $\cH$ and $\cG$, where $\mathcal{L}$ typically involves the moment $\E[(Y - h(X))g(Z)]$ together with regularizers, e.g., $\mathcal{L}(h, g) = \E[(Y - h(X))g(Z)] + \Phi(h) - \Psi(g)$. 
The optimization can be unstable and may fail to converge, especially when deep neural networks are used as function classes. On the other hand, two-stage methods split NPIV estimation into the following steps. First, depending on the specific method, Stage~1 estimates either the conditional expectation operator $\cT$ or the conditional density $P(X\mid Z)$. Second, Stage~2 performs a regression of the outcome on the estimator obtained in Stage~1. When both stages involve a least squares problem, a two-stage method is called a two-stage least squares (2SLS) regression. Compared to min--max optimization methods, two-stage estimation provides more stable algorithms since it avoids saddle-point optimization. 
Furthermore, two-stage methods offer valuable flexibility regarding data collection: they do not strictly require a fully paired dataset of $(X,Y,Z)$. Instead, they can naturally leverage two separate datasets, using $m$ observations of $(X,Z)$ to train Stage~1 and $n$ observations of $(Y,Z)$ to train Stage~2.

While there are many algorithms proposed to address the NPIV estimation problem, theoretical understanding of these algorithms remains a topic of active research. In the domain of modern NPIV learning methods, \citet{bennett2023minimax} and \citet{li2024regularized} address consistency under our general setting: namely, in the strong $L_2(X)$ norm, and without unique identification. \citet{bennett2023minimax} show convergence in $L_2$ norm but do not adapt to different degrees of smoothness. \citet{li2024regularized} offer smoothness-adaptive $L_2$ rates that can accommodate the non-identified setting; however, they resort to density estimation in Stage~1, which proves difficult in complex high-dimensional scenarios. Regarding NPIV methods based on series estimators or kernel density estimation, \citet{florens2011identification,chen2012estimation,babii2017completeness} consider convergence to a minimum-norm solution in the absence of identification. \citet{hall2005nonparametric} obtain the first lower bound in $L_2$ norm for NPIV in the mildly ill-posed setting, under a source condition with respect to the operator $\cT^*\cT$, and show that an estimator based on density estimation combined with Tikhonov regularization can achieve this lower bound. \citet{chen2011rate} obtain a lower bound in $L_2$ norm in both the mildly and severely ill-posed settings, under a source condition (with respect to a user-defined hypothesis class rather than $\cT^*\cT$) and a link condition measuring the smoothness of $\cT$ relative to the hypothesis class; they then show that a sieve minimum distance estimator can achieve the lower bound. Finally, \citet{chen2018optimal} obtain a lower bound in $L_{\infty}$ norm for the structural function and its derivatives and show that a sieve-based estimator achieves the lower bound. Recent research has extended the kernel IV framework to the more complex regime of IV with observed covariates \citep{shen2025nonparametric}. This work 
builds directly on our analysis of KIV, but addresses the distinct ``partial identity'' structure of the conditional expectation operator that arises in the presence of observed covariates. Due to the resulting difficulty of handling anisotropic smoothness, their analysis relies on Gaussian kernels and modified link conditions, which differ from the general framework we analyze here. 
In related work by some of the present authors, \citet{kimoptimality} study
\emph{deep feature instrumental variable} regression (DFIV), where the feature
representations used in the two stages are learned from data using neural networks.
Their analysis establishes minimax convergence rates (in a strong $L_2$ metric) for
Besov-type structural classes while requiring only a \emph{linear} first-stage sample size
$m \asymp n$, under (i) a two-sided link condition (involving both a forward and a reverse
link) and (ii) a \emph{maximal-smoothness} compatibility assumption on the conditional
expectation operator that controls the intrinsic difficulty of Stage~1.
Outside this regime, the rates in \citet{kimoptimality} deteriorate---in particular,
when the maximal-smoothness condition fails, Stage~1 becomes harder and the resulting
upper bound slows down, and when the reverse link is substantially looser than the forward
link their upper and lower bounds no longer match, so minimax optimality is not guaranteed.
In contrast, the strong-norm upper bounds developed in the present paper are derived under
a \emph{one-sided} (lower) link condition and do not rely on a maximal-smoothness assumption.
An interesting open direction is to understand whether the sufficient allocation threshold for
fixed-feature KIV can be sharpened to the linear regime $m \asymp n$, or whether achieving
$m \asymp n$ fundamentally requires adaptive first-stage procedures (e.g.\ representation learning),
as in \citet{kimoptimality}.

\section{Background}\label{sec:bg}
We introduce notation and recall the RKHS background needed in the paper. This material is developed in \cite{lietal2022optimal,li2024towards,meunier2024optimal}; we include it here for convenience and ease of reference.

\subsection{Notation and Tensor Product of Hilbert Spaces}
Throughout the paper, we consider three random variables: $X$ (the endogenous variable), $Y$ (the outcome) and $Z$ (the instrument). $Y$ is defined on $\mathbb{R}$ while $X$ and $Z$ are defined respectively on the second countable locally compact Hausdorff spaces $E_X$ and $E_Z$ endowed with their respective Borel $\sigma$-fields $\mathcal{F}_{E_X}$ and $\mathcal{F}_{E_Z}$. We let $(\Omega, \mathcal{F},\mathbb{P})$ be the underlying probability space with expectation operator $\mathbb{E}$. Let $P$ be the pushforward of $\P$ under $(X,Y,Z)$, and let $\pi_W$, $W \in \{X,Y,Z,(X,Y),(Z,Y),(X,Z)\}$, denote the marginal distributions. We use a Markov kernel $p: E_Z \times \mathcal{F}_{E_X} \rightarrow [0,1]$ to represent the conditional distribution of $X$ given $Z$, i.e., \[\mathbb{P}[X \in A|Z = z] = p(z,A),\] for all $z \in E_Z$ and event $A \in \mathcal{F}_{E_X}$. We denote the space of real-valued Lebesgue square-integrable functions on $(E_X,\mathcal{F}_{E_X})$ with respect to $\pi_X$ as $L_2(E_X,\mathcal{F}_{E_X},\pi_X)$, abbreviated $L_2(X)$, and similarly for $\pi_Z$ as $L_2(E_Z,\mathcal{F}_{E_Z},\pi_Z)$, abbreviated $L_2(Z)$. We introduce some notation related
to linear operators on Hilbert spaces and vector-valued integration; 
formal definitions can be found in Appendix~\ref{sec:bg_appendix}
for completeness, or we refer the reader to \citet{weidmann80linear,diestel77vector}. Let $H$ be a separable real Hilbert space with inner product $\langle \cdot, \cdot \rangle_{H}$.  $L_2(E_Z,\mathcal{F}_{E_Z},\pi_Z; H)$, abbreviated $L_2(Z; H)$, is the space of strongly measurable and Bochner $2$-integrable functions from $E_Z$ to $H$ equipped with the norm $\| \cdot \|_{L_2(Z;H)}^2 = \int_{E_Z} \|\cdot  \|_{H}^2 \, \mathrm{d}\pi_Z$. We write $\mathcal{L}(H,H')$ as the Banach space of bounded linear operators from $H$ to another Hilbert space $H'$, equipped with the operator norm $\|\cdot\|_{H \rightarrow H'}$. When $H=H'$, we simply write $\mathcal{L}(H)$ instead. We write $S_2(H,H')$ as the Hilbert space of Hilbert-Schmidt operators from $H$ to $H'$ and $S_1(H,H')$ as the Banach space of trace class operators (see Appendix~\ref{sec:bg_appendix} for a complete definition). 
For two Hilbert spaces $H,H'$, we say that $H$ is (continuously) embedded in $H'$ and denote it as 
$H \hookrightarrow H'$ if $H$ can be interpreted as a vector subspace of $H'$ and the inclusion operator $i: H \to H'$ 
performing the change of norms with $i x = x$ for $x \in H$ is continuous; and we say that $H$ is isometrically isomorphic to $H'$ and denote it as $H \simeq H'$ if there is a linear isomorphism between $H$ and $H'$ which is an isometry. For any linear operator $A$, $\cR(A)$ denotes its range and $\cN(A)$ its null space. For any bounded linear operator $A$, $A^*$ denotes its adjoint. For any subspace $M \subseteq H$, $M^{\perp}$ denotes its orthogonal complement.

\medskip
\noindent {\textbf{Tensor Product of Hilbert Spaces} (\citealp{aubin2000applied}, Section 12)\textbf{.}} Denote by $H\otimes H'$  the tensor product of Hilbert spaces $H$, $H'$. The element $x\otimes x' \in H \otimes H'$ is treated as the linear rank-one operator $x\otimes x': H' \rightarrow H$ defined by $y' \rightarrow \langle y',x' \rangle_{H'}x $ for $y' \in H'$. 
Based on this identification, the tensor product space $H\otimes H'$ is isometrically isomorphic to the space of Hilbert-Schmidt operators from $H'$ to $H$, i.e., $H\otimes H' \simeq S_2(H',H)$. We will hereafter not make the distinction between these two spaces, and treat them as being identical.

\begin{remark}[\citealp{aubin2000applied}, Theorem 12.6.1]\label{rem:tensor_product}
Consider the Bochner space $L_2(Z;H)$ where $H$ is a separable Hilbert space. One can show that $L_2(Z;H)$ is isometrically identified with the tensor product space $H \otimes L_2(Z)$, and we denote as $\Psi$ the isometric isomorphism between the two spaces. See Appendix~\ref{sec:bg_appendix} for more details. 
\end{remark}

\subsection{Reproducing Kernel Hilbert Spaces and Conditional Mean Embedding}
{\bf Scalar-valued Reproducing Kernel Hilbert Space (RKHS).} 
We let $k_{X}: E_X \times E_X \rightarrow \mathbb{R}$ be a symmetric and positive definite kernel function and $\mathcal{H}_{X}$ be a vector space of functions from $E_X$ to $\mathbb{R}$, endowed with a Hilbert space structure via an inner product $\langle\cdot, \cdot\rangle_{\mathcal{H}_{X}}$. We say $k_{X}$ is a reproducing kernel of $\mathcal{H}_{X}$ if and only if for all $ x \in E_X$ we have $k_{X}(\cdot,x) \in \mathcal{H}_{X}$ and for all $x \in E_X$ and $f \in \mathcal{H}_{X}$, we have $f(x)=\left\langle f, k_{X}(\cdot,x)\right\rangle_{\mathcal{H}_{X}}$. A space $\mathcal{H}_{X}$ which possesses a reproducing kernel is called a reproducing kernel Hilbert space (RKHS; \citealp{berlinet2011reproducing}). We denote the canonical feature map of $\mathcal{H}_{X}$ as $\phi_{X}(x) = k_{X}(\cdot,x)$.  Similarly for $E_Z$, we consider a RKHS $\mathcal{H}_{Z}$ with symmetric and positive definite kernel $k_{Z}: E_Z \times E_Z \rightarrow \mathbb{R}$ and canonical feature map denoted as $\phi_{Z}$.

\begin{asst} \label{asst:rkhs} The RKHSs $\cH_X,\cH_Z$ and kernels $k_X,k_Z$ satisfy:
\begin{enumerate}[itemsep=0em]
    \item \label{assump:separable} $\mathcal{H}_{X}$ and $\mathcal{H}_{Z}$ are separable, this is satisfied
    if $k_X, k_Z$ are continuous, given that $E_X, E_Z$ are separable;\footnote{This follows from \citet[Lemma 4.33]{steinwart2008support}. Note that the lemma requires separability of $E_X, E_Z$, which is satisfied since we assume that $E_X, E_Z$ are second countable locally compact Hausdorff spaces.}
    \item \label{assump:measurable} $k_{X}(\cdot,x)$ and $k_{Z}(\cdot,z)$ are measurable for $\pi_X$-almost all $x \in E_X$ and $\pi_Z$-almost all $z \in E_Z$;
    \item \label{assump:bounded} $k_X(x,x) \leq 1$ for $\pi_X$-almost all $x \in E_X$ and $k_Z(z,z) \leq 1$ for $\pi_Z$-almost all $z \in E_Z$.
\end{enumerate}
\end{asst}
\noindent
The above assumptions are not restrictive in practice, as well-known kernels such as the Gaussian, Laplace and Mat{\'e}rn kernels satisfy them on $\mathbb{R}^d$ \citep{sriperumbudur2011universality}. The uniform bound in Assumption~\ref{asst:rkhs}.\ref{assump:bounded} is without loss of generality: if $k_X(x,x)\le \kappa_X^2$ a.s., we can replace $k_X$ by the rescaled kernel $\tilde k_X \doteq k_X/\kappa_X^2$, which satisfies $\tilde k_X(x,x)\le 1$ and only changes universal multiplicative constants in the sequel, without affecting rates. Similarly for $k_Z$.\\
We now introduce some facts about the interplay between $\cH_X$ and $L_{2}(X),$ which has been extensively studied by \citet{smale2004shannon,smale2005shannon}, \citet{de2006discretization} and \citet{steinwart2012mercer}. We first define the (not necessarily injective) embedding $\cI_{X}: \cH_X \rightarrow L_{2}(X)$, mapping a function $f \in \cH_X$ to its $\pi_X$-equivalence class $[f]_{X}$. The embedding is a well-defined compact operator since its Hilbert-Schmidt norm can be bounded as \citep[][Lemma 2.2 \& 2.3]{steinwart2012mercer}
$\left\|\cI_{X}\right\|_{S_{2}\left(\cH_X, L_{2}(X)\right)} \leq 1.$ The adjoint operator $S_{X}\doteq\cI_{X}^{*}: L_{2}(X) \rightarrow \cH_X$ is an integral operator with respect to the kernel $k_X$, i.e. for $f \in L_{2}(X)$ and $x \in E_X$ we have \citep[][Theorem 4.27]{steinwart2008support}
$$
\left(S_{X} f\right)(x)=\int_{E_X} k_X\left(x, x^{\prime}\right) f\left(x^{\prime}\right) \mathrm{d} \pi_X\left(x^{\prime}\right).
$$
Next, we define the self-adjoint, positive semi-definite and trace class operators
$$
L_{X}\doteq \cI_{X} S_{X}: L_{2}(X) \rightarrow L_{2}(X) \quad \text { and } \quad C_{X}\doteq S_{X} \cI_{X}: \cH_X \rightarrow \cH_X.
$$
By the spectral theorem for self-adjoint compact operators, there exist a non-increasing sequence $(\mu_{X,i})_{i \geq 1} > 0$, and a family $(e_{X,i})_{i \geq 1} \subseteq \cH_X$, such that $\left([e_{X,i}]_X\right)_{i \geq 1}$ is an orthonormal basis (ONB) of $\overline{\cR(\cI_{X})} \subseteq L_2(X)$, $(\mu_{X,i}^{1/2}e_{X,i})_{i \geq 1}$ is an ONB of $\cN(\cI_{X})^{\perp} \subseteq \cH_X$, and we have 
\begin{equation} \label{eq:SVD}
    L_{X} = \sum_{i \geq 1} \mu_{X,i} \langle\cdot, [e_{X,i}]_X \rangle_{L_2(X)}[e_{X,i}]_X, \qquad C_{X} = \sum_{i \geq 1} \mu_{X,i} \langle\cdot, \mu_{X,i}^{\frac{1}{2}}e_{X,i} \rangle_{\cH_X} \mu_{X,i}^{\frac{1}{2}}e_{X,i}.
\end{equation}
We similarly define $\cI_Z, S_Z, L_Z, C_Z, (\mu_{Z,i})_{i \geq 1}, (e_{Z,i})_{i \geq 1}$ for the RKHS $\cH_Z$. $C_X$ and $C_Z$ admit the covariance-operator representation
\[
C_X = \E\big[\phi_X(X)\otimes \phi_X(X)\big] \in S_1(\cH_X), \qquad C_Z = \E\big[\phi_Z(Z)\otimes \phi_Z(Z)\big] \in S_1(\cH_Z)
\]
in the sense that for all $f\in \cH_X$,
$C_X f = \E\big[\phi_X(X)\langle \phi_X(X),f\rangle_{\cH_X}\big]$, and similarly for $C_Z$. We also define the cross-covariance operator
\[
C_{XZ}\doteq \E\big[\phi_X(X)\otimes \phi_Z(Z)\big]\in S_2(\cH_Z,\cH_X).
\]

\medskip
\noindent {\bf Vector-valued Reproducing Kernel Hilbert Space (vRKHS).} Let $K: E_Z \times E_Z \rightarrow \mathcal{L}(\cH_X)$ be an operator valued positive definite kernel \citep[][Definition 2.2]{carmeli2006vector}. Fix $z \in E_Z$, and $h \in \cH_X$, then $\left(K_z h\right)(\cdot)\doteq K(\cdot, z) h$
defines a function from $E_Z$ to $\cH_X$. The completion of 
$
\mathcal{G}_{\text {pre }}\doteq\operatorname{span}\left\{K_z h \mid z \in E_Z, h \in \cH_X\right\}
$
with inner product defined on the elementary elements as $\left\langle K_z h, K_{z^{\prime}} h^{\prime}\right\rangle_{\mathcal{G}}\doteq\left\langle h, K\left(z, z^{\prime}\right) h^{\prime}\right\rangle_{\cH_X}$, defines a vRKHS denoted as $\mathcal{G}$. For a more complete overview of the vector-valued reproducing kernel Hilbert space, we refer the reader to \cite{carmeli2006vector}, \cite{carmeli2010vector} and \citet[][Section 2]{li2024towards}. In the following, we will denote $\mathcal{G}$ as the vRKHS induced by the kernel $K: E_Z \times E_Z \rightarrow \mathcal{L}(\cH_X)$ with 
\begin{equation*} 
K(z,z') \doteq k_Z(z,z')\operatorname{Id}_{\cH_X}, \quad z,z' \in E_Z.
\end{equation*}
We emphasize that this family of kernels is the de facto standard for high- and infinite-dimensional applications 
\citep{grunewalder2012conditional,grunewalder2012modelling,park2020measure,ciliberto2016consistent,ciliberto2020general,singh2019kernel,mastouri2021proximal,kostic2022learning,kostic2024sharp} due to the crucial \textit{representer theorem} which gives a closed-form solution to estimators derived from this family of kernels.

\begin{remark}[General multiplicative kernel]
\label{rem:general_kernel}
    Without loss of generality, we provide our results for the vRKHS
    $\mathcal{G}$ induced by the operator-valued kernel given by
    $K = k_Z\operatorname{Id}_{\cH_X}$.
    However, with suitably adjusted constants in the assumptions, 
    our results transfer directly to the more general
    vRKHS $\widetilde{\mathcal{G}}$ induced by the more general operator-valued kernel 
    \(
    \widetilde K(z,z') \doteq k_Z(z,z')Q, z,z' \in E_Z,
    \)
    where $Q: \cH_X \to \cH_X$
    is any positive-semidefinite self-adjoint operator. The characterization of the adjusted constants is given by \citet{li2024towards}.
\end{remark}
\noindent
An important property of $\mathcal{G}$ is that it is isometrically isomorphic to the space of Hilbert-Schmidt operators between $\cH_Z$ and $\cH_X$ \citep[Corollary 1]{li2024towards}. Similarly to the scalar case, we can map every element in $\cG$ into its $\pi_Z-$equivalence class in $L_2(Z;\cH_X)$ and we use the shorthand notation $[F] = [F]_{Z;X}$ for some $F \in \cG$ (see Appendix~\ref{sec:bg_appendix} for more details).

\begin{theorem}[vRKHS isomorphism, Corollary 1, \cite{li2024towards}] \label{theo:operep}
For every function $F\in \mathcal{G}$ there exists a unique operator $C \in S_2(\cH_Z, \mathcal{H}_X)$ such that $F(\cdot) = C\phi_Z(\cdot) \in \mathcal{H}_X$ with $\|C\|_{S_2(\cH_Z, \mathcal{H}_X)} = \|F\|_{\mathcal{G}}$ and vice versa. Hence $\cG \simeq S_2(\cH_Z, \mathcal{H}_X)$ and $\cG$ can be written as $\mathcal{G}= \left\{F: E_Z \to \mathcal{H}_X \mid F = C\phi_Z(\cdot),\, C \in S_2(\cH_Z, \mathcal{H}_X)\right\}$. 
\end{theorem}
\noindent
{\bf Conditional Mean Embedding.} A particular advantage of kernel methods is their convenience for working with probability distributions \citep[see e.g.,][]{smola2007hilbert,sejdinovic2013equivalence,tolstikhin2017minimax}. In particular, kernel methods allow us to deal with conditional distributions through the conditional mean embedding, as defined in \citet{song2009hilbert,grunewalder2012conditional,park2020measure,klebanov2020rigorous}.

\begin{definition}[Conditional Mean Embedding (CME)] 
The $\mathcal{H}_{X}$-valued conditional mean embedding (CME) for the conditional distribution of $X$ given $Z$, is defined as 
\begin{equation*}
F_{*}(\cdot)\doteq\int_{E_X}\phi_{X}(x)p(\cdot,dx) = \mathbb{E}\left[\phi_{X}(X)|Z = \cdot \right] \in L_2(Z; \mathcal{H}_{X}). 
\end{equation*}
\end{definition}
\noindent
By the reproducing property, we have $\mathbb{E}[f(X)|Z=z] = \langle f, F_*(z) \rangle_{\mathcal{H}_{X}}$, for all $f \in \mathcal{H}_{X}$ and for $\pi_Z$-almost all $z \in E_Z$. Therefore, the CME allows us to easily compute conditional expectations through an inner product, which as we will see below is a crucial step in KIV. The approximation of $F_*$ with Tikhonov regularization (also known as vector-valued kernel ridge regression) is a key concept in kernel methods and is extensively studied in \citet{park2020measure, lietal2022optimal, li2024towards}, where learning $F_*$ is formulated as the following optimization problem at the population level:
\begin{equation}\label{eqn:cme_pop}
F_{\xi}\doteq \argmin_{F \in \mathcal{G}} \E_{X,Z} \left\|\phi_X(X) - F(Z)\right\|^2_{\mathcal{H}_X} + \xi \|F\|_{\mathcal{G}}^2 = C_{\xi}\phi_Z(\cdot),
\end{equation}
where $C_{\xi} \doteq C_{XZ}\left(C_{Z} + \xi \operatorname{Id}_{\cH_Z}\right)^{-1} \in S_2(\cH_Z, \cH_X)$.
Tikhonov regularization in \cref{eqn:cme_pop} is known to exhibit the \emph{saturation effect} where it fails to benefit from high smoothness of the target function $F_*$. This was recently verified by \citet{meunier2024optimal} in the context of vector-valued regression. To avoid saturation, we therefore generalize Tikhonov regularization to general spectral regularization (see also \citet{mollenhauer2020nonparametric, mollenhauer2022learning}). General spectral regularization typically starts with defining a filter function, i.e., a function on an interval which is applied on self-adjoint operators to each individual eigenvalue via the spectral calculus.
\begin{definition}[Filter function]
    \label{def:genreg_reg_fn}
Let $\Lambda\subseteq \mathbb{R}^{+}$. A family of functions $g_{\xi}: [0,\infty)\to [0,\infty)$ indexed by $\xi \in \Lambda$ is called a filter with qualification $\rho\geq0$ if it satisfies the following two conditions:
\begin{enumerate}[itemsep=0em]
    \item There exists a positive constant $E$ such that, for all $\xi \in \Lambda$
    \begin{equation*}
        \sup_{\theta \in[0,1]}\sup_{x\in \left[0,1\right]}\xi^{1-\theta}x^{\theta}g_{\xi}(x)\le E.
        \end{equation*}
    \item There exists a positive constant $\omega_{\rho}<\infty$ such that
    \begin{equation*}
            \sup_{\theta\in[0,\rho]}\sup_{\xi \in\Lambda}\sup_{x\in \left[0,1\right]}|1 - g_{\xi}(x)x|x^{\theta}\xi^{-\theta}\le \omega_{\rho}.
        \end{equation*} 
\end{enumerate}
\end{definition}
\noindent
One may think of the above definition as a class of functions approximating the inversion map $x \mapsto 1/x$ while still being defined for $x=0$ in a reasonable way. As examples, with $g_{\xi}(x) = (x+\xi)^{-1}$, we retrieve ridge regression with $\rho = 1$ and with $g_\xi(x) = x^{-1} \mathds{1}[x \geq\xi]$ we obtain kernel principal component regression with $\rho = +\infty$. We refer to Appendix~\ref{sec:bg_appendix} for other examples of spectral methods such as gradient descent, iterated Tikhonov and gradient flow. Given a filter function $g_{\xi}$, we call $g_{\xi}(C_{Z})$ the regularized inverse of $C_{Z}$. We may think of the regularized inverse as approximating the \textit{pseudoinverse}
of $C_Z$ (see e.g.\ \citet{EnglHankeNeubauer2000}) when $\xi \to 0$. We define the regularized population solution with filter function $g_{\xi}$ as
\begin{equation*} 
    C_{\xi} \doteq C_{XZ}g_{\xi}(C_{Z}) \in S_2(\cH_Z, \cH_X), \qquad F_{\xi}(\cdot) \doteq C_{\xi} \phi_Z(\cdot) \in \cG.
\end{equation*}
{\bf Empirical solution}: Given a dataset $\cD_1 = \{(\tilde{z}_i, \tilde{x}_i)\}_{i=1}^m$, the empirical analogue to $C_{\xi}$ is
\begin{equation} \label{eqn:vkrr_func}
\hat{C}_{\xi} \doteq \hat{C}_{XZ}g_{\xi}(\hat{C}_{Z}) , \qquad \hat{F}_{\xi}(\cdot) \doteq \hat{C}_{\xi}\phi_Z(\cdot) \in \cG,
\end{equation}
where $\hat{C}_{XZ}$, $\hat{C}_{Z}$ are empirical covariance operators defined as
$$
\hat{C}_{Z} \doteq \frac{1}{m} \sum_{i=1}^m \phi_Z(\tilde{z}_i) \otimes \phi_Z(\tilde{z}_i) \qquad \hat{C}_{XZ} \doteq \frac{1}{m} \sum_{i=1}^m \phi_X(\tilde{x}_i) \otimes \phi_Z(\tilde{z}_i). 
$$
The formula is obtained in \cite{meunier2024optimal} using a generalization of the \emph{representer theorem}. 

\section{Instrumental Variable Estimation with RKHS} \label{sec:IV_RKHS}
In this section, we illustrate how kernel-based algorithms can be used to solve the NPIV problem. Recall that the structural function $h_0$ can be written as a solution to the functional equation \(\mathcal{T}h = r_0,\) where $r_0(z)=\E[Y\mid Z=z]$ and $(\cT h)(z)=\E[h(X)\mid Z=z]$. A standard \emph{solvability} condition in the NPIV literature is $r_0 \in \cR(\cT)$, which ensures that the integral equation admits at least one solution in $L_2(X)$. However, this condition alone does not specify a unique target: when $\cT$ is not injective, the solution set is an affine space of the form $\tilde{h} + \mathcal N(\cT)$, where $\tilde{h}$ is any solution of $\cT h = r_0$.
This failure of injectivity can occur in practice, for example, when both $X$ and $Z$ are discrete and $|\operatorname{supp}(X)| > |\operatorname{supp}(Z)|$, $\cT$ cannot be injective. In such cases it is not clear a priori which solution a given estimator converges to, and obtaining guarantees in the strong $L_2$ norm becomes more delicate. Motivated by recent work on unidentified NPIV \citep[see e.g.,][]{chen2021robust,bennett2023minimax,li2024regularized}, we therefore fix a canonical target by selecting a minimum-norm solution. Specifically, let $\tilde \cT \doteq \cT \circ \cI_X: \cH_X \to L_2(Z)$ and denote \(\tilde \cT ^{-1}({r_0}) = \{h \in \cH_X: \tilde{\cT} h = r_0\}.\)

\begin{asst}[Well-specifiedness of solutions] \label{asst:exist_rkhs} $\tilde \cT ^{-1}({r_0}) \ne \emptyset$.
\end{asst}
\noindent
The RKHS $\cH_X$ encompasses our a priori belief on the properties that $h_0$ should satisfy. Assumption~\ref{asst:exist_rkhs} states that there is at least one function in $\cH_X$ satisfying the integral equation. Note that Assumption~\ref{asst:exist_rkhs} is stronger than $r_0 \in \cR(\cT)$ as $\cH_X$ can be seen as a subset of $L_2(X)$. However, for a universal RKHS, $\cH_X$ is dense in $L_2(X)$ under mild conditions \citep[see e.g.,][]{sriperumbudur2011universality}. Since $\cT$ is not guaranteed to be injective, $\tilde \cT$ is also not guaranteed to be injective. The minimum RKHS norm solution is then defined as
\begin{equation*} 
h_* \doteq \argmin_{h \in \tilde \cT ^{-1}({r_0})} \|h\|_{\cH_X}. 
\end{equation*}
$h_*$ is the pseudo-inverse of the linear system: $h_* = \tilde{\cT}^{\dagger}r_0$ \citep{EnglHankeNeubauer2000}. The next proposition shows that $h_*$ is uniquely defined; the proof is postponed to Appendix~\ref{sec:proof_minimum_norm_sol}.

\begin{proposition}\label{prop:uni_mini}
Under Assumption~\ref{asst:exist_rkhs}, $h_*$ exists uniquely and $\{h_*\} = \cN(\tilde \cT)^{\perp} \cap \tilde \cT ^{-1}({r_0})$.
\end{proposition}

\begin{remark}
Our construction yields a unique target $h_*$ both when $\cT$ is injective and when it is not. We will see that the kernel-based estimator converges to $h_*$ in either case. Related minimum-norm targets and convergence analyses under identification failure appear in \citet{florens2011identification,chen2012estimation,babii2017completeness}.
\end{remark}

\subsection{The KIV Estimator} \label{sec:Kernel_NPIV}
Under Assumption~\ref{asst:exist_rkhs}, $h_* \in \cH_X$, and using the reproducing property, for $\pi_Z$-almost all $z \in E_Z$,
$
(\tilde{\mathcal{T}}h_*)(Z)=\E\left[\langle h_*, \phi_X(X)\rangle_{\cH_X} \mid Z\right] = \langle h_*, F_*(Z) \rangle_{\cH_X}.
$
\cite{singh2019kernel} then suggest a two-stage least squares estimation procedure, where we use a Stage 1 sample of size $m$, $\cD_1 \doteq\{(\tilde z_i, \tilde x_i)\}_{i=1}^m$, for the Stage~1 regression, and an independent Stage 2 sample of size $n$, $\cD_2\doteq\{(z_i,y_i)\}_{i=1}^n$, for the Stage~2 regression.
\begin{enumerate}
    \item estimate $F_*$ with vector-valued regression using dataset $\cD_1$ and spectral regularization;
    \item estimate $h_*$ through regressing $Y$ on $F_*(Z)$ using dataset $\cD_2$. 
\end{enumerate}
Instead of using Tikhonov regularization, as in \cite{singh2019kernel}, below we employ a learning procedure with general spectral algorithms for Stage~1. Strong convergence guarantees are preserved for this general regularization scheme thanks to the results of \cite{meunier2024optimal} for vector-valued regression with spectral regularization.

\medbreak
\noindent {\bf Stage~1.} Using $\cD_1$ we apply Eq.~\eqref{eqn:vkrr_func} to obtain the empirical estimator $\hat{F}_{\xi}$.

\medbreak
\noindent {\bf Stage~2.} The algorithm at the population level can be written as
\begin{equation}\label{eqn:iv_pop}
\begin{aligned}
h_{\lambda}\doteq  \argmin_{h \in \mathcal{H}_X} \E[(Y - \langle h, F_*(Z)\rangle_{\cH_X} )^2] + \lambda \|h\|_{\mathcal{H}_{X}}^2 
\end{aligned} 
\end{equation}
Empirically, we use the estimated $\hat{F}_{\xi}$ from Stage~1 to learn $h_*$ with $\cD_2$
\begin{equation*}
\hat{h}_{\lambda,\xi}\doteq \argmin_{h \in \mathcal{H}_X} \frac{1}{n} \sum_{i = 1}^n \left(y_i - \left\langle h , \hat{F}_{\xi}(z_i)\right\rangle_{\mathcal{H}_X} \right)^2 + \lambda \|h\|_{\mathcal{H}_{X}}^2. 
\end{equation*}
KIV with Tikhonov regularization for both stages admits a closed-form solution as derived in \cite{singh2019kernel}. We provide a new version with spectral algorithms in Appendix \ref{sec:explt}. We also introduce $\bar{h}_{\lambda}$, a theoretical estimator for Stage~2 with access to the true CME,
\begin{equation}
\bar{h}_{\lambda} \doteq \argmin_{h \in \mathcal{H}_X} \frac{1}{n} \sum_{i = 1}^n \left(y_i - \left\langle h , F_{*}(z_i)\right\rangle_{\mathcal{H}_X} \right)^2 + \lambda \|h\|_{\mathcal{H}_{X}}^2. \label{eq:iv_bar}
\end{equation}

\begin{remark}[Spectral Algorithm] 
One could also attempt to employ spectral regularization for Stage~2, instead of Tikhonov regularization. However, the interplay between the qualification of the filter function with our smoothness assumptions (see \eqref{asst:src} below) is far from trivial. We therefore leave this investigation for future work. A first step in that direction appears in \cite{bennett2023source} where they study how iterated Tikhonov regularization can be incorporated into a conditional moment model.
\end{remark}
\section{Ill-posedness} \label{sec:measure_subspace}

Our next step is to characterize the behavior of the KIV estimator, by bounding $\|\hat h_{\lambda,\xi} - h_*\|_{L_2(X)}$. Stage~1 in 2SLS estimates the conditional mean embedding $F_*$, and $F_*$ determines which directions in $\cH_X$ are statistically visible from the instrument. To make this precise, define the covariance operator
\begin{equation*}
C_F \doteq \E[F_*(Z) \otimes F_*(Z)].
\end{equation*}
Since $\tilde \cT h(Z) = \langle h, F_*(Z) \rangle_{\cH_X}$ for $h \in \cH_X$, we have the operator identity
\(
C_F = \tilde \cT^* \tilde \cT
\)
on $\cH_X$. Moreover, because $F_*(Z) = \E[\phi_X(X) \mid Z]$, Jensen's inequality yields $C_F \preceq C_X$, and therefore $\cN(C_X) \subseteq \cN(C_F)$. By \Cref{prop:uni_mini}, the minimum-norm target satisfies
\[
h_* \in \cN(\tilde \cT)^{\perp} = \cN(C_F)^{\perp} \subseteq \cN(C_X)^{\perp}.
\]
We refer to $\cN(C_F)^{\perp}$ as the \emph{identified component}: it is the instrument-induced subspace of $\cH_X$ over which Stage~2 effectively learns. Thus, although the estimator is optimized over all of $\cH_X$, the relevant object is the
spectral geometry of the identified component inside $\cN(C_X)^\perp$. This motivates the link condition below, 
which quantifies ill-posedness on the identified component in terms of the ambient geometry encoded by $C_X$.

\begin{asst}[Link condition]
Let $\gamma \in [1,+\infty)$. We assume there exists a constant $R > 0$ such that
\[
R\,\|C_X^{\gamma/2} f\|_{\cH_X}
\leq
\|C_F^{1/2} f\|_{\cH_X},
\qquad \forall f \in \cN(C_F)^\perp.
\tag{LINK}\label{eq:link_up}
\]
\end{asst}
\noindent
To simplify notation, we absorb $R$ into universal constants throughout and write \eqref{eq:link_up} with $R = 1$. We focus on polynomial-link, or mildly ill-posed, settings. In particular, there are situations where no finite $\gamma$ can satisfy \eqref{eq:link_up}: for example, if $C_X$ and $C_F$ share an eigenbasis with eigenvalues $\mu_{X,i} = i^{-2}$ and $\mu_{F,i} = e^{-i}$, then \eqref{eq:link_up} would require $i^{-\gamma} \lesssim e^{-i/2}$, which fails for every fixed $\gamma < \infty$. This is the severely ill-posed regime familiar in inverse problems; see, e.g., \citet[Theorem~1]{chen2011rate}. \eqref{eq:link_up} is equivalent to the operator inequality
\(
P_F C_X^\gamma P_F \preceq C_F,
\)
where $P_F$ denotes the orthogonal projection onto $\cN(C_F)^\perp$. Further consequences are collected in Appendix~\ref{sec:proof_subsize}.

\begin{remark}[$\gamma$ as an ill-posedness parameter]
In kernel regression, the spectrum of $C_X$ quantifies the statistical complexity of the
RKHS $\cH_X$: faster eigenvalue decay means smaller effective dimension and hence easier
learning. The operator $C_F$ plays the analogous role for the component of $\cH_X$ that is
statistically visible through the instrument.

Assumption~3 states that, on the identified component $\cN(C_F)^\perp$, the weak quantity
$\|C_F^{1/2}f\|_{\cH_X}$ controls the stronger norm
$\|C_X^{\gamma/2}f\|_{\cH_X}$. 
Thus $\gamma$ measures how much additional smoothing is induced by the channel $X \mapsto Z$, through the comparison between the operator scales generated by $C_F$ and $C_X$.

When $\gamma=1$, the instrument does not introduce additional ill-posedness on the identified component. As we will see in the next section, the resulting strong $L_2(X)$ rate for estimating the minimum-norm solution $h_\ast$ matches that of ordinary kernel ridge regression. A simple special case is $Z = X$, for which $\mathcal T = Id$ and $h_\ast = h_0 = \E[Y \mid X]$, so the problem reduces exactly to ordinary regression. 
When $\gamma>1$, estimating $h_\ast$ in strong $L_2(X)$ norm requires inverting additional smoothing induced by the instrument, and the rate deteriorates accordingly. In this precise sense, $\gamma$ is the ill-posedness parameter of NPIV in our framework.
\end{remark}
\begin{remark}[Relation to classical link conditions]
Classical inverse-problem analyses often assume a two-sided comparison between the forward
operator and the hypothesis geometry: for some index function $\omega : [0,\infty)\to[0,\infty)$
and constants $R_0,R_1>0$,
\[
R_0\|\omega(C_X)^{1/2}f\|_{\cH_X}
\le
\|C_F^{1/2}f\|_{\cH_X}
\le
R_1\|\omega(C_X)^{1/2}f\|_{\cH_X},
\qquad
f\in\cH_X,
\]
see, e.g., \citet{nair2005regularization,chen2011rate}. In the polynomial case $\omega(t)=t^\gamma$, this becomes the two-sided comparison
$C_F \asymp C_X^\gamma$.

Our upper-bound analysis needs only the lower half of this relation, namely \eqref{eq:link_up},
and only on the identified component $\cN(C_F)^\perp$. This is important for two reasons.
First, when $C_F$ is not injective, a global condition on all of $\cH_X$ is unnatural because
Stage~2 only learns on the identified component. Second, the upper comparison is not needed
for the strong $L_2(X)$ upper bound.

\end{remark}


\section{Minimax Optimal Learning Rates} \label{sec:minimax_rates}

In this section, we establish the minimax optimality of the KIV estimator. Before stating our assumptions, we briefly recall the interpolation spaces associated with the integral operators. These spaces provide a convenient way to express regularity conditions via fractional powers of $L_X$ and $L_Z$, and will be used to state smoothness assumptions for both scalar- and vector-valued regression. Readers are referred to Appendix~\ref{sec:bg_appendix} for full details.  We start with scalar-valued functions. Given $\beta \geq  0$ and a squared-integrable scalar-valued function $f \in L_2(Z)$, the $\beta-$interpolation norm is defined as 
$$
 \left\|f\right\|_\beta \doteq \|L_Z^{-\beta / 2}f\|_{L_2(Z)}.
$$
The subset of $ f \in L_2(Z)$ for which $\left\|f\right\|_\beta < +\infty$ is denoted $[\cH_Z]^\beta$. $[\cH_X]^\beta \subseteq L_2(X)$ is defined similarly with $L_X$. Vector-valued interpolation norms and spaces introduced by \cite{lietal2022optimal} generalize the above interpolation space definitions to spaces of vector-valued functions. Given $\beta \geq  0$ and a vector-valued function $F \in L_2(Z; \cH_X)$ since $L_2(Z; \cH_X)$ is isometric to $S_2(L_2(Z), \cH_X)$ (see Remark~\ref{rem:tensor_product}), there is an operator $C \in S_2(L_2(Z), \cH_X)$ such that $\|F\|_{L_2(Z; \cH_X)} = \|C\|_{S_2(L_2(Z), \cH_X)}$. The vector-valued $\beta-$interpolation norm is then defined as 
\begin{equation} \label{eq:vv_norm}
\|F\|_{\beta} \doteq \|C\|_{\beta} \doteq \|CL_Z^{-\beta / 2}\|_{S_2(L_2(Z), \mathcal{H}_X)}.
\end{equation}
The space of $F \in L_2(Z; \cH_X)$ such that $\|F\|_{\beta} < +\infty$ is denoted $[\cG]^{\beta}$. For details regarding vector-valued interpolation spaces, we refer to Appendix~\ref{sec:bg_appendix}.

\subsection{Assumptions for Stage~1} \label{sec:asst_stage_1}
The analysis of Stage~1 convergence is studied in \cite{lietal2022optimal,li2024towards} for Tikhonov regularization; and later generalized to spectral generalization by \cite{meunier2024optimal}. We summarize their results, which rely on the following assumptions:

\begin{asst}[Eigenvalue decay for Stage~1]
There exist constants $\overline{c}_Z >0$ and $p_Z \in (0,1]$ such that for all
\begin{equation}\label{asst:evdz}
        \mu_{Z,i} \leq  \overline{c}_Z i^{-1/p_Z}. \tag{EVDZ}
\end{equation}
\end{asst}

\begin{asst}[Embedding into $L_{\infty}$ for Stage~1]
There exists $\alpha_Z \in [p_Z, 1]$ such that the inclusion map $\mathcal{I}^{\alpha_Z, \infty}_{Z}: [\mathcal{H}]_{Z}^{\alpha_Z} \hookrightarrow L_{\infty}(Z)$ is continuous and there is a constant $A_Z > 0$ such that,
    \begin{equation}\label{asst:embz}
    \|\mathcal{I}^{\alpha_Z, \infty}_{Z}\|_{[\mathcal{H}_Z]^{\alpha_Z} \rightarrow L_{\infty}(Z)} = A_Z. \tag{EMBZ}
    \end{equation}
\end{asst}

\begin{asst}[Source condition for Stage~1]
There exists $\beta_Z > \alpha_Z$, and a constant $B_Z \geq 0$ such that,
    \begin{equation}\label{asst:srcz}
    \|F_*\|_{\beta_Z} = \|C_*L_Z^{-\frac{\beta_Z}{2}}\|_{S_2(L_2(Z), \mathcal{H}_X)} \leq B_Z, \tag{SRCZ}
    \end{equation}
    where $C_{*} \doteq \Psi^{-1}(F_*) \in S_2(L_2(Z), \mathcal{H}_X)$ (see Remark~\ref{rem:tensor_product} for the definition of $\Psi$). 
\end{asst}
\noindent
\eqref{asst:evdz} is a classical assumption that characterizes the size of the RKHS $\cH_Z$ equipped with the marginal distribution $\pi_Z$. \eqref{asst:srcz} characterizes the smoothness of the target function $F_*$. Property \eqref{asst:embz} is referred to as the \textit{embedding property} in \cite{fischer2020sobolev}. It can be shown that it holds if and only if there exists a constant $A_Z \geq  0$ with $\sum_{i \geq 1} \mu_i^{\alpha} e_{Z,i}^2(z) \leq  A_Z^2$ for $\pi_Z$-almost all $z \in E_Z$ \citep[Theorem 9]{fischer2020sobolev}. Since $k_Z$ is bounded, \eqref{asst:embz} always holds with $\alpha_Z = 1$. Moreover, \eqref{asst:embz} implies a polynomial eigenvalue decay \eqref{asst:evdz} of order $1/\alpha_Z$ (in particular, one may take $p_Z = \alpha_Z$). Hence we assume $0< p_Z \leq  \alpha_Z \leq 1$. For an in-depth discussion of these assumptions, we refer the reader to \cite{li2024towards}. Under \eqref{asst:evdz}, \eqref{asst:srcz}, \eqref{asst:embz}, \cite{meunier2024optimal} demonstrate that the estimator in Eq.~\eqref{eqn:vkrr_func} converges to $F_*$.\\
The following (informal) result is adapted from \citet[Theorem 4]{meunier2024optimal}; we refer to Theorem~\ref{theo:cme_rate} in Appendix~\ref{sec:auxiliary} for the formal statement. The $L_2-$rate is minimax-optimal, matching the lower bound of \cite{li2024towards}. Moreover, the same tuning achieves the minimax rate even when the target lies outside the hypothesis space: in particular, when $\alpha_Z \le \beta_Z < 1$, one has \(F_*\notin \cG\).

\begin{theorem}\label{theo:cme}
Let $g_{\xi}$ be a filter function with qualification $\rho \geq 1$ used to build the estimator $\hat F_{\xi}$ on $\cD_1$ with Eq.~\eqref{eqn:vkrr_func}. Let Assumptions~\ref{asst:rkhs}, \eqref{asst:evdz}, \eqref{asst:srcz} and \eqref{asst:embz} hold with $\beta_Z \in (\alpha_Z,2\rho]$ and $0< p_Z \leq  \alpha_Z \leq  1$. Taking $\xi_m = \Theta \left(m^{-\frac{1}{\beta_Z + p_Z}}\right)$, there are constants $J,J' > 0$ such that with high probability, 
$$
\begin{aligned}
    \left\|\hat{F}_{\xi_m} - F_*\right\|^2_{L_2(Z; \cH_X)} &\leq  J m^{-\frac{\beta_Z}{\beta_Z + p_Z}} \qquad \& \qquad \left\|\hat{F}_{\xi_m} - F_*\right\|^2_{L_{\infty}(Z; \cH_X)} \leq  J' m^{-\frac{\beta_Z-\alpha_Z}{\beta_Z + p_Z}}.
\end{aligned}
$$
\end{theorem}
\noindent
For readability we state Theorem~\ref{theo:cme} in $L_2(Z;\cH_X)$ and $L_\infty(Z;\cH_X)$ norms. 
Appendix~\ref{sec:auxiliary} provides a stronger bound in general interpolation norms, which in particular recovers control in the vRKHS norm $\|\cdot\|_{\cG}$ used by \citet{singh2019kernel}.\\
We compare Theorem~\ref{theo:cme} with the Stage~1 analysis of \citet{singh2019kernel} for Tikhonov regularization. 
First, Theorem~\ref{theo:cme} allows for targets outside the Stage~1 hypothesis space, i.e., it permits misspecification such as $F_*\notin \cG$. 
Second, it provides rates directly in the norms $L_2(Z;\cH_X)$ and $L_\infty(Z;\cH_X)$, whereas \citet{singh2019kernel} states its Stage~1 guarantees in the vRKHS norm $\|\cdot\|_{\cG}$ norm. 
Additionally, the rates explicitly adapt to the eigenvalue decay of $C_Z$ through \eqref{asst:evdz}; in our notation, analyses that do not exploit such decay correspond to the worst-case choice $p_Z=1$. Finally, Theorem~\ref{theo:cme} demonstrates the benefit of general spectral regularization. While standard Tikhonov regression ($\rho=1$) saturates for highly smooth targets ($\beta_Z > 2$), spectral filters with higher qualification ($\rho > 1$) avoid this saturation, exploiting smoothness up to $2\rho$ for faster rates.

\subsection{Assumptions for Stage~2} 

\begin{asst}[Eigenvalue decay for Stage~2]
For some constants $\overline{c}_X >0$ and $p_X \in (0,1]$ and for all $i \geq 1$,
\begin{equation}\label{asst:evd}
        \mu_{X,i} \leq  \overline{c}_X i^{-1/p_X}.\tag{EVDX}
\end{equation}
\end{asst}

\begin{asst}[Source condition for Stage~2]
    There exists $\beta_X \geq 1$ and a constant $B_X \geq  0$ such that
    \begin{equation}\label{asst:src}
        \|h_*\|_{\beta_X} = \left\|L_X^{-\frac{\beta_X}{2}} [h_*]_X \right\|_{L_2(X)}\leq  B_X. \tag{SRCX}
    \end{equation}
\end{asst}

\begin{asst}[MOM]
    There are constants $\sigma, L>0$ such that for all integers $q\ge 2$,
    \begin{equation}\label{asst:mom}
        \E\left[|Y - \E[h_*(X) \mid Z]|^q \mid Z\right] \leq  \frac{1}{2} q ! \sigma^2 L^{q-2}. \tag{MOM}
    \end{equation}
\end{asst}
\noindent
\eqref{asst:evd} and \eqref{asst:src} play the same role as for Stage~1; the former characterizes the size of the space $\cH_X$ equipped with the marginal distribution $\pi_X$ while the latter characterizes the smoothness of the target function $h_*$. Note that \eqref{asst:src} can be equivalently stated as $\|C_X^{-\frac{\beta_X-1}{2}} h_* \|_{\cH_X}\leq  B_X$. Finally, \eqref{asst:mom} is a Bernstein moment condition used to control the noise of the observations (see \citealp{caponnetto2007optimal,fischer2020sobolev} for more details).
Under these assumptions, the next theorem provides an upper bound on the learning risk $\|\hat h_{\lambda} - h_*\|_{L_2(X)}$. The proof is in Appendix~\ref{sec:proof_ub}.

\begin{theorem}\label{theo:iv_l2} Let the assumptions of \Cref{theo:cme} hold and let Assumption~\ref{asst:exist_rkhs},  \eqref{eq:link_up}, \eqref{asst:evd}, \eqref{asst:src} and \eqref{asst:mom} hold with $p_X \in (0,1]$ and $1 \leq \beta_X \leq \gamma + 1$. Fix $\tau\ge 1$, $\lambda \in (0,1]$ and $\xi_m$ as in Theorem~\ref{theo:cme}. Condition on the first-stage sample $\cD_1$ used to construct $\hat F_{\xi_m}$.
Then, for $n,m$ large enough, with probability at least $1-12e^{-\tau}$ over the draw of the second-stage sample $\cD_2$, we have
\begin{equation*}
    \begin{aligned}
    \|\hat h_{\lambda,\xi_m} - h_*\|_{L_2(X)} &\leq J_0 \tau \lambda^{\frac{c_F}{2\gamma} -1} \left(\left\|\hat F_{\xi_m} - F_*\right\|_{L_2(Z;\cH_X)} +\frac{\left\| \hat F_{\xi_m} - F_*\right\|_{\alpha_Z}}{\sqrt{n}} \right)\bigl(\|\bar h_\lambda\|_{\cH_X}+1\bigr)\\
    & + J_1 \tau\left(\lambda^{\frac{\beta_X}{2\gamma}} + 
    \frac{1}{n \lambda^{1 - \frac{1}{2 \gamma}}} + \frac{1}{\sqrt{n} \lambda^{\frac{\gamma + p_X - 1}{2 \gamma}}}
\right) 
    \end{aligned}
\end{equation*}
where $J_0, J_1$ depend on $\sigma, L, A_Z, B_Z, \alpha_Z, \beta_Z, p_X, B_X$, $\|h_*\|_{\cH_X}$ and $c_F \doteq \mathrm{1}_{\cN(C_F) = \{0\}}$.
\end{theorem}
\noindent
Theorem~\ref{theo:iv_l2} gives a finite-sample upper bound on $\|\hat h_{\lambda,\xi_m}-h_*\|_{L_2(X)}$.
The first term on the right-hand side quantifies how the Stage~1 CME estimation error propagates into Stage~2,
while the second term is the intrinsic Stage~2 error (bias--variance trade-off given $\cT$).
The explicit “$n,m$ large enough” condition is stated in Appendix~\ref{sec:proof_ub}, Theorem~\ref{theo:iv_l2_full}.

\begin{remark}[Comparison to pseudo-metric guarantees]
\citet{singh2019kernel} establish minimax-optimal convergence guarantees for KIV with Tikhonov regularization in both stages, but their analysis is in the pseudo-metric
\(
\|\cT(\hat h_{\lambda,\xi} - h_*)\|_{L_2(Z)}.
\)
Since conditional expectation is an $L_2$-contraction, $\|\cT (\hat h_{\lambda,\xi} - h_*)\|_{L_2(Z)} \leq \|\hat h_{\lambda,\xi} - h_*\|_{L_2(X)}$. Therefore, convergence in the pseudo-metric does not imply convergence in strong $L_2(X)$; it can go to zero even when $\|\hat{h}_{\lambda,\xi}-h_*\|_{L_2(X)}$ does not. In contrast, we work directly with $\|\hat h_{\lambda,\xi} - h_*\|_{L_2(X)}$, which yields a strong-norm guarantee.
\end{remark}

\begin{corollary}
\label{cor:iv_rate_main}
Assume the conditions of Theorem~\ref{theo:iv_l2} and Assumption~\eqref{asst:evdz} hold. Let
\[
m=n^a
\qquad\text{for some } a>0,
\qquad
\xi_m=\Theta\!\left(m^{-1/(\beta_Z+p_Z)}\right),
\]
and define
\[
c_F \doteq \mathbf 1_{N(C_F)=\{0\}},
\qquad
D \doteq \beta_X+\gamma+p_X-1,
\qquad
\Delta \doteq \beta_X+2\gamma-c_F,
\]
\[
\delta \doteq \bigl(1-\beta_X+(\gamma-1)p_X\bigr)_+,
\qquad
\widetilde\Delta \doteq \Delta+\delta,
\]
\[
\widetilde a_A \doteq \frac{\beta_Z+p_Z}{\beta_Z}\frac{\widetilde\Delta}{D},
\qquad
\widetilde a_B \doteq \frac{\beta_Z+p_Z}{\beta_Z-\alpha_Z}\frac{\widetilde\Delta-D}{D},
\qquad
\widetilde a_\ast \doteq \max\{\widetilde a_A,\widetilde a_B\}.
\]
If \(a \ge \widetilde a_\ast\), then taking
\(
\lambda_n=\Theta\!\left(n^{-\gamma/D}\right)
\)
yields
\[
\|\hat h_{\lambda_n,\xi_m}-h_*\|_{L_2(X)}^2
=
O_P\!\left(
n^{-\beta_X/D}
\right)
=O_P\!\left(
n^{-\frac{\beta_X}{\beta_X+\gamma+p_X-1}}
\right).
\]
The full set of regimes (including Stage~1--limited rates when $a<a_\ast$) is given in Appendix~\ref{sec:proof_cor}.
\end{corollary}
\noindent
We now complement the upper bound with a minimax lower bound in the strong $L_2(X)$ norm. We fix the pair $(k_X,\pi_X)$, which determines the covariance operator $C_X$ and hence the smoothness class appearing in Assumption~\eqref{asst:src}. Unlike in classical NPIV lower bounds, the adversary may vary not only the structural function $h_\ast$ but also the instrument channel, that is, the conditional law of $X$ given $Z$. In the hardest case, the channel saturates the allowed ill-posedness, so that $C_F \asymp C_X^\gamma$. Assumption~\ref{ass:link_plus_exist} formalizes the existence of such a channel.

\begin{asst}[Existence of a $\gamma$-ill-posed channel]
\label{ass:link_plus_exist}
Let $\pi_X$ be the fixed marginal law of $X$. We assume that there exists at least one joint
distribution $\pi_{X,Z}$ on $E_X\times E_Z$ with $X$-marginal $\pi_X$ such that the
associated instrument-induced covariance operator $C_F$ satisfies
\begin{equation}\label{asst:link+}
R_0 C_X^\gamma \preceq C_F \preceq R_1 C_X^\gamma \tag{LINK+}
\end{equation}
for some constants $R_0,R_1>0$. 
\end{asst}
\noindent
We additionally require the following two-sided regularity conditions to make sure that the eigenvalues of the marginal covariance operator do not decay faster than \eqref{asst:evd} guarantees.

\begin{asst}
There exist constants $\underline{c}_X,\overline{c}_X >0$ and $p_X \in (0,1)$ such that for all $i \geq 1$,
    \begin{equation}\label{asst:evd+}
      \underline{c}_X i^{-1/p_X} \leq  \mu_{X,i} \leq \overline{c}_X i^{-1/p_X}. \tag{EVDX+}
    \end{equation}
\end{asst}

\begin{theorem}[Minimax lower bound]
\label{theo:lower_bound}
Let $\pi_X$ and $k_X$ be a probability distribution and a kernel on $E_X$ respectively such that \Cref{asst:rkhs}, \eqref{asst:evd+} and \eqref{asst:link+} hold. Fix $\beta_X\ge 1$ and constants
$B_X,\sigma,L>0$.

Then there exist constants $J_0,J,r>0$, depending only on the fixed parameters and the
constants in the assumptions, such that for every learning method
\(
(\cD_1,\cD_2)\mapsto \widehat h(\cD_1,\cD_2),
\)
every $\tau>0$, every $m\ge 1$, and all sufficiently large $n$, there exists an NPIV model $P$ over
$(X,Z,Y)$ such that:
\begin{enumerate}
\item the $X$-marginal of $P$ is $\pi_X$;
\item the associated instrument-induced covariance operator $C_F$ satisfies \eqref{asst:link+} with
exponent $\gamma$;
\item the target $h_*$ satisfies \eqref{asst:src} with parameters $(B_X,\beta_X)$;
\item \eqref{asst:mom} holds with parameters $(\sigma,L)$;
\item if
\[
\cD_1=((\tilde Z_i,\tilde X_i))_{i=1}^m\sim P_{Z,X}^{\otimes m},
\qquad
\cD_2=((Z_i,Y_i))_{i=1}^n\sim P_{Z,Y}^{\otimes n},
\]
independently, then
\[
\bigl(\pi_{Z,X}^{\otimes m}\otimes \pi_{Z,Y}^{\otimes n}\bigr)
\left(
\|\widehat h(\cD_1,\cD_2)-h_*\|_{L_2(X)}^2
\ge \tau J\, n^{-\frac{\beta_X}{\beta_X+\gamma-1+p_X}}
\right)
\ge
1-J_0\tau^{1/r}.
\]
\end{enumerate}
\end{theorem}
\noindent
To show that \eqref{asst:link+} is non-vacuous and covers
classical econometric constructions, we record the following Sobolev example.

\begin{example}[Noisy instrument model]
\label{ex:noisy_proxy}
Let $E_X=E_Z=\mathbb T^d$, the Torus on $[0,1]^d$, let $\pi_X$ be the uniform measure on $\mathbb T^d$,
and let $k_X$ be a periodic Sobolev (equivalently, periodic Mat\'ern-type)
kernel whose RKHS is equivalent to $H^{\nu}(\mathbb T^d)$ for some
$\nu>d/2$. Suppose
\[
Z = X + \varepsilon \pmod 1,
\]
where $\varepsilon$ is independent of $X$ and has density $q$ on $\mathbb T^d$.
Assume that the Fourier coefficients of $q$ satisfy
\begin{equation}
\label{eq:ordinary_smooth_torus}
 c\,(1+|\ell|^2)^{-\xi}
 \;\le\;
 |\widehat q(\ell)|^2
 \;\le\;
 C\,(1+|\ell|^2)^{-\xi},
 \qquad \ell\in\mathbb Z^d,
\end{equation}
for some constants $\xi\ge 0$ and $0<c\le C<\infty$.

\medskip \noindent
Then the conditional expectation operator $\cT :h\mapsto \E[h(X)\mid Z]$ is the
convolution operator $\cT h = q*h$, and both $C_X$ and $\cT^*\cT$ are diagonal in the
Fourier basis $(e_{\ell})_{\ell\in\mathbb Z^d}$. More precisely,
\[
\mu_{\ell}(C_X) \asymp (1+|\ell|^2)^{-\nu},
\qquad
\cT^*\cT e_{\ell} = |\widehat q(\ell)|^2 e_{\ell}.
\]
Since $C_F = I_X^* \cT^*\cT I_X$, it follows that
\[
\mu_{\ell}(C_F)
\asymp
(1+|\ell|^2)^{-(\nu+\xi)}
\asymp
\mu_{\ell}(C_X)^{\,1+\xi/\nu}.
\]
Therefore Assumption~\eqref{asst:link+} holds with
\(
\gamma = 1 + \frac{\xi}{\nu}.
\)
Moreover, for this fixed Sobolev class one has $p_X = d/(2\nu)$, and the source
condition~\eqref{asst:src} with parameter $\beta_X=s/\nu$ corresponds to Sobolev
smoothness $h_*\in H^{s}(\mathbb T^d)$ \citep{fischer2020sobolev}. Consequently, \Cref{theo:lower_bound} yields the concrete lower rate
\[
 n^{-\frac{s}{s+\xi+d/2}}.
\]
Under the sample-allocation regime of \Cref{cor:iv_rate_main} ($a \geq \widetilde a_\ast$), the upper bound matches the same
exponent. Thus this example provides a transparent benchmark in which the upper and
lower rates coincide. This matches the classical minimax $L_2$ rate for NPIV over Sobolev classes in the mildly ill-posed regime with exponent $\xi$; see \cite{chen2011rate}.
\end{example}
\begin{remark}[Comparison with standard kernel ridge regression]
\label{rem:comparison_krr}
The Sobolev rate in Example~\ref{ex:noisy_proxy}
should be compared to the classical nonparametric regression rate on the same Sobolev scale \citep{stone1982optimal},
\[
n^{-\,\frac{s}{s+d/2}},
\]
which is recovered by our general exponent as soon as the problem is
\emph{well-posed} (no additional smoothing by the conditional expectation),
i.e.\ $\gamma=1$ in \Cref{theo:lower_bound}, equivalently $\xi=0$
in \Cref{eq:ordinary_smooth_torus}.
In the convolution model, $\xi=0$ corresponds to a ``flat'' transfer function
$|\widehat q(\ell)|^2\asymp 1$, so that conditioning on $Z$ does not dampen high
frequencies beyond what is already dictated by $(k_X,\pi_X)$.

More generally, for standard kernel ridge regression without instrumental variables, the minimax rate under \eqref{asst:evd} and \eqref{asst:src} is \citep{fischer2020sobolev}
\[
n^{-\,\frac{\beta_X}{\beta_X + p_X}}.
\]
The role of the ill-posedness parameter is transparent: NPIV incurs an additional penalty of size $\gamma-1$ in the
denominator of the minimax exponent compared to the standard regression case. Again, this rate is achieved by KIV when there is no ill-posedness ($\gamma=1$).
\end{remark}
\noindent
The preceding comparison also clarifies why the link parameter governs the
difficulty of NPIV: even when the hypothesis class is held fixed (Sobolev
smoothness), weaker instruments strictly slow down learning in the strong
$L_2(X)$ norm.

\begin{remark}[How many first-stage samples are needed?]
\label{rem:allocation_exponent}
To streamline the discussion, we focus on the \emph{identified} case (i.e.\ $\cN(C_F)=\{0\}$, so
$c_F=1$ in \Cref{cor:iv_rate_main}).
Recall that the stage-2-optimal rate in \Cref{cor:iv_rate_main} is
\(
n^{-\beta_X/(\beta_X+\gamma+p_X-1)} .
\)
The corollary shows that this exponent is achieved as soon as $m=n^{a}$ with
\(a \ge \widetilde a_\ast \doteq \max\{\widetilde a_A,\widetilde a_B\}\),
where
\(
\delta \doteq \bigl(1-\beta_X+(\gamma-1)p_X\bigr)_+,
\)
\[
\widetilde a_A =
\left(
1+\frac{p_Z}{\beta_Z}
\right)
\frac{\beta_X+2\gamma-1+\delta}{\beta_X+\gamma+p_X-1},
\qquad
\widetilde a_B =
\frac{\beta_Z+p_Z}{\beta_Z-\alpha_Z}
\frac{\gamma-p_X+\delta}{\beta_X+\gamma+p_X-1}.
\]
Thus the sufficient allocation threshold factors into a purely first-stage term and
a stage-2 transfer term.

\smallskip
\noindent\textbf{Stage~1 quantities ($\beta_Z,p_Z,\alpha_Z$).}
The parameter $\beta_Z$ is the Stage~1 smoothness index for the conditional mean
embedding. The parameter $p_Z$ quantifies the eigenvalue decay of $C_Z$ and therefore the effective dimension of the $Z$-geometry. The parameter $\alpha_Z$ is the sup-norm embedding index of $\cH_Z$ and allows a tight control of the sup-norm of the conditional mean embedding which is necessary to propagate the stage-1 error into stage-2. Stage~1 becomes easier when the target is smoother (larger $\beta_Z$), when the geometry
is simpler (smaller $p_Z$), and when the sup-norm is smaller (smaller $\alpha_Z$); decreasing the required allocation.

\smallskip
\noindent\textbf{Stage~2 quantities ($\beta_X,\gamma,p_X$).}
The parameter $\beta_X$ is the smoothness index of $h_\star$ relative to
$C_X$; smoother structural functions (larger $\beta_X$) reduce both transfer factors and
therefore reduce the required Stage~1 sample size. The ill-posedness index $\gamma\ge 1$ controls how strongly the conditional expectation operator smooths directions of $h_\star$; larger $\gamma$ amplifies estimation error and therefore
increases both transfer factors.
Finally, $p_X$ quantifies the eigenvalue decay of $C_X$ (effective dimension of the endogenous regressor geometry).
Larger $p_X$ corresponds to slower eigenvalue decay and hence a \emph{slower} stage-2-optimal
rate; this in turn relaxes the requirement that Stage~1 error be negligible at the target rate.

\smallskip
\noindent\textbf{Consequences and open problem.}
Since $\gamma\ge 1$ and $p_X\in(0,1)$, we have $\beta_X+2\gamma-1+\delta > \beta_X+\gamma+p_X-1$, hence $\widetilde a_A>1$ and therefore $\widetilde a_\star>1$.
Thus, according to the present analysis, achieving the stage-2-optimal exponent requires more
$(X,Z)$ pairs than $(Y,Z)$ pairs.
Importantly, $\widetilde a_\star$ is a \emph{sufficient} allocation threshold.
Determining whether one can attain the stage-2-optimal exponent with fewer first-stage samples
(i.e.\ whether $\widetilde{a}_\star$ is sharp) would require lower bounds for the \emph{two-sample} learning
problem that explicitly depend on both sample sizes $(m,n)$.
Such bounds would quantify, in a minimax sense, how much information about the reduced form
(or conditional law of $X$ given $Z$) must be learned from $\cD_1$ in order to reach the
stage-2-optimal rate based on $\cD_2$.
To the best of our knowledge, allocation-dependent minimax lower bounds of this type are
largely missing in the general literature on two-stage procedures, including NPIV.

\smallskip \noindent \textbf{Sobolev Case.}
We now instantiate $\widetilde a_\star$ explicitly in \Cref{ex:noisy_proxy}. We have
\[
p_X=\frac{d}{2\nu},
\qquad
\beta_X=\frac{s}{\nu},
\qquad
\gamma=1+\frac{\xi}{\nu}.
\]
Moreover, since \(|\widehat q(\ell)|>0\) for all \(\ell\), the convolution operator
\(\cT:h\mapsto q*h\) is injective, so \(c_F=1\). 
Assume in addition that Stage~1 is also built on a periodic Sobolev scale over
\(E_Z=\mathbb T^d\): let \(k_Z\) be a periodic Sobolev kernel of order \(\nu_Z>d/2\),
and suppose that \eqref{asst:srcz} holds with \(\beta_Z=t/\nu_Z\), $t > d/2$ (\citet{li2024towards} shows that this is equivalent to \(F_*\in H^t(\mathbb T^d; \cH_X)\), the vector-valued Sobolev space of order \(t\)). Then \eqref{asst:embz} and \eqref{asst:evdz} hold with \(\alpha_Z=p_Z=d/(2\nu_Z)\) \citep{fischer2020sobolev}. Define
\[
\delta_{\mathrm{Sob}} \doteq \left(
\nu - s + \frac{\xi d}{2\nu}
\right)_+.
\]
The sufficient stage-1 allocation exponent is
\[
\widetilde{a}_\star
=
\max\left\{\left(\frac{2t+ d}{2t}\right)\frac{\nu+s+2\xi + \delta_{\mathrm{Sob}}}{s+\xi+d/2},\left(\frac{2t+d}{2t-d}\right)\frac{\nu+\xi-d/2+\delta_{\mathrm{Sob}}}{s+\xi+d/2} \right\}
\]
In the limit of a very smooth first stage (\(t\to\infty\)),
\[
\widetilde{a}_\star \downarrow \frac{\nu+s+2\xi + \delta_{\mathrm{Sob}}}{s+\xi+d/2},
\]
which lies in \((1,3)\) under the standing assumption \(s\ge \nu\) (equivalently,
\(\beta_X\ge 1\)).
\end{remark}

\section{Conclusion}
We have studied the kernel instrumental variable (KIV) estimator as a two-stage
least-squares method for nonparametric instrumental variable (NPIV) regression.
Our analysis covers both identified and non-identified regimes. When the NPIV
operator $\cT$ is not injective, we show that KIV still converges---in the strong
$L_2(X)$ norm---to a canonical target: the minimum-$\cH_X$-norm solution of the
integral equation. This resolves the ambiguity as to which solution is learned
under lack of identification, and yields guarantees in a metric of direct
statistical interest.

A central message of the paper is that statistical rates are governed by the
geometry of the instrument-induced component. We formalize this through a
polynomial link condition comparing the covariance operator $C_X$ to the induced operator $C_F$ (identified geometry), leading to the ill-posedness index $\gamma$. Under standard eigenvalue-decay
and source assumptions, we derive finite-sample $L_2(X)$ rates for KIV (with
general spectral regularization in Stage~1, avoiding saturation of Tikhonov
regularization), and we complement them with a minimax lower bound. Together,
these results show that the obtained rates are minimax-optimal over the
considered model class.

Finally, our bounds clarify the precise sense in which NPIV is harder than ordinary kernel ridge regression. In the well-posed case \(\gamma = 1\), our strong \(L_2(X)\) rate reduces to the classical kernel regression rate. In contrast, when \(\gamma > 1\), the instrumental-variable structure induces additional smoothing, so recovering \(h_\ast\) in strong \(L_2(X)\) norm becomes statistically harder and the minimax rate slows down accordingly (cf.\ Remark~\ref{rem:comparison_krr}). This slowdown is not an artifact of the analysis or of KIV, but a fundamental feature of NPIV.

There are several directions for future work. It would
be valuable to extend the Stage~2 analysis to more general spectral
regularization, and to develop sharper conditions under which the Stage~1 error
becomes negligible with minimal sample size. On the modeling side, understanding
how representation learning or adaptive first-stage procedures (e.g.\ deep
feature learning) can reduce constants or improve robustness in complex
high-dimensional settings remains an important practical question, although the
minimax lower bound indicates that ill-posedness-driven slowdowns cannot be
eliminated in worst case. Related work by some of the present authors \citep{kimoptimality} provides one concrete step in this
direction by analyzing deep feature IV estimators and showing that, under additional compatibility assumptions,
a balanced sample regime $m \asymp n$ can suffice for minimax-optimal rates.

\bibliography{references}
\newpage
\appendix
\section*{Appendices}
\noindent Section~\ref{sec:explt} provides a closed-form expression for the KIV estimator. Section~\ref{sec:bg_appendix} provides additional background material. Section~\ref{sec:proof_minimum_norm_sol} provides the proof of Proposition~\ref{prop:uni_mini}. Section~\ref{sec:proof_subsize} provides additional consequences to \eqref{eq:link_up}. Section~\ref{sec:proof_ub} provides a proof sketch (Section~\ref{sec:sketch}) followed by the detailed proof (Section~\ref{sec:detailed}) for the upper bound presented in Theorem~\ref{theo:iv_l2}, and finally the proof of Corollary~\ref{cor:iv_rate_main} (Section~\ref{sec:proof_cor}). Section~\ref{sec:proof_lb} proves the lower bound given in 
Theorem~\ref{theo:lower_bound}. Section~\ref{sec:more_bounds} provides additional bounds used in the main proofs. Finally, in Section~\ref{sec:auxiliary}, we collect some technical supporting results.

\section{Explicit Solutions of KIV} \label{sec:explt}
The closed-form solution for KIV with Tikhonov regularization for both stages is given in \citet[Algorithm 1,][]{singh2019kernel}. We provide the closed-form solution where we allow a general regularization scheme for Stage 1. 
We use $[m]=\{1,\dots,m\}$ and $[n]=\{1,\dots,n\}$.

\medskip \noindent
\textbf{Stage 1.} In Stage 1, given $\cD_1$ and $\xi > 0$ (\Cref{eqn:vkrr_func}), we estimate $\hat{F}_{\xi}(\cdot) = \hat{C}_{\xi}\phi_Z(\cdot)$, with

\begin{equation}
\hat{C}_{\xi} = \frac{1}{m}\mathbf{\Phi}_{\tilde X}^*\mathbf{\Phi}_{\tilde Z}g_{\xi}\left( \frac{1}{m}\mathbf{\Phi}_{\tilde Z}^*\mathbf{\Phi}_{\tilde Z} \right), \label{eqn:cme_sol}
\end{equation}
with
\begin{equation*}
    \begin{aligned}
\mathbf{\Phi}_{\tilde Z}: \cH_Z \rightarrow \R^m &\qquad \mathbf{\Phi}_{\tilde Z} = [\phi_Z(\tilde z_1),\dots, \phi_Z(\tilde z_m)]^{*} \\
\mathbf{\Phi}_{\tilde X}: \cH_X \rightarrow \R^m &\qquad \mathbf{\Phi}_{\tilde X} = [\phi_X(\tilde x_1),\dots, \phi_X(\tilde x_m)]^{*} 
    \end{aligned}
\end{equation*}
The solution can also be written in the following dual form (see \citealp{meunier2024optimal}):
\begin{equation*}
\hat{C}_{\xi} = \frac{1}{m}\mathbf{\Phi}_{\tilde X}^*g_{\xi}\left(\frac{\mathbf{K}_{\tilde{Z}\tilde{Z}}}{m} \right)\mathbf{\Phi}_{\tilde Z},
\end{equation*}
where we introduce the Gram matrix
$
\mathbf{K}_{\tilde Z \tilde Z} = \mathbf{\Phi}_{\tilde Z}\mathbf{\Phi}_{\tilde Z}^*, [\mathbf{K}_{\tilde Z \tilde Z}]_{ij} = k_Z(\tilde z_i, \tilde z_j), i,j \in [m]. 
$

\medskip \noindent
\textbf{Stage 2.} $\hat h_{\lambda,\xi}$ admits the following closed-form expression, 
$$
\hat h_{\lambda,\xi} = \left( \frac{1}{n}\mathbf{\Phi}_{\hat F}^*\mathbf{\Phi}_{\hat F} + \lambda \operatorname{Id}_{\cH_X}\right)^{-1}\frac{1}{n}\mathbf{\Phi}_{\hat F}^*Y =\left( \hat C_{\hat F} + \lambda \operatorname{Id}_{\cH_X}\right)^{-1}\frac{1}{n}\mathbf{\Phi}_{\hat F}^*Y,
$$
where 
\begin{equation*}
\begin{aligned}
&\mathbf{\Phi}_{\hat F}: \cH_X \rightarrow \R^n \qquad \mathbf{\Phi}_{\hat F} = [\hat{F}_{\xi}(z_1),\dots, \hat{F}_{\xi}(z_n)]^{*},\\
&\hat C_{\hat F} = \frac{1}{n}\mathbf{\Phi}_{\hat F}^*\mathbf{\Phi}_{\hat F} = \frac{1}{n}\sum_{i=1}^n \hat{F}_{\xi}(z_i) \otimes \hat{F}_{\xi}(z_i),
\end{aligned}
\end{equation*}
which we can write in dual form as follows:
$$
\hat{h}_{\lambda,\xi}= \mathbf{\Phi}_{\hat F}^* \left[\mathbf{F}+n\lambda \operatorname{Id}_n\right]^{-1}Y, \qquad Y = [y_1,\dots, y_n]^T \in \mathbb{R}^n
$$
$$
\begin{aligned}
\mathbf{F}_{ij} & = [\mathbf{\Phi}_{\hat F}\mathbf{\Phi}_{\hat F}^*]_{ij} = \langle \hat{F}_{\xi}(z_i) , \hat{F}_{\xi}(z_j)\rangle_{\cH_X}\qquad i,j \in [n]
\end{aligned}
$$
Define
\(
\mathbf{\Phi}_{Z}: \cH_Z \rightarrow \R^n, \mathbf{\Phi}_{Z} = [\phi_Z(z_1),\dots, \phi_Z(z_n)]^{*}.
\)
By Eq.~\eqref{eqn:cme_sol} and using \\$\hat{F}_{\xi}(\cdot) = \hat{C}_{\xi}\phi_Z(\cdot)$, we obtain the closed-form expressions for $\mathbf{F}$ and $\mathbf{\Phi}_{\hat F}^*$:
$$
\begin{aligned}
\mathbf{\Phi}_{\hat F}^* &= \frac{1}{m}\mathbf{\Phi}_{\tilde X}^*g_{\xi}\left(\frac{\mathbf{K}_{\tilde{Z}\tilde{Z}}}{m} \right)\mathbf{K}_{\tilde{Z}Z} \\
\mathbf{F} &= \frac{1}{m^2}\mathbf{K}_{\tilde{Z}Z}^{\top}g_{\xi}\left(\frac{\mathbf{K}_{\tilde{Z}\tilde{Z}}}{m} \right)\mathbf{K}_{\tilde{X} \tilde{X}}g_{\xi}\left(\frac{\mathbf{K}_{\tilde{Z}\tilde{Z}}}{m} \right)\mathbf{K}_{\tilde{Z}Z},
\end{aligned}
$$
where,
$$
\begin{aligned}
\mathbf{K}_{\tilde Z Z} &= \mathbf{\Phi}_{\tilde Z}\mathbf{\Phi}_{Z}^* \in \R^{m\times n}, \qquad [\mathbf{K}_{\tilde Z Z}]_{ij} =k_Z(\tilde z_i, z_j) \qquad i \in [m], j \in [n] \\
\mathbf{K}_{\tilde X \tilde X}  &= \mathbf{\Phi}_{\tilde X}\mathbf{\Phi}_{\tilde X}^*\in \R^{m\times m}, \qquad [\mathbf{K}_{\tilde X\tilde X}]_{ij} = k_X(\tilde x_i,\tilde x_j) \qquad i,j \in [m].
\end{aligned}
$$
Therefore, introducing $\mathbf{J} \doteq \frac{1}{m}g_{\xi}\left(\frac{\mathbf{K}_{\tilde{Z}\tilde{Z}}}{m} \right)\mathbf{K}_{\tilde{Z}Z}$, for a new test point $x \in E_X$, we have
$$
\hat{h}_{\lambda,\xi}(x) = \langle \hat{h}_{\lambda}, \phi_X(x) \rangle_{\cH_X} =\mathbf{K}_{\tilde{X}x}^{\top}\mathbf{J}\left[\mathbf{J}^{\top}\mathbf{K}_{\tilde{X} \tilde{X}}\mathbf{J}+n\lambda \operatorname{Id}_n\right]^{-1}Y,
$$
where
$
\mathbf{K}_{\tilde{X}x}= [k_X(\tilde{x}_1,x),\dots,k_X(\tilde{x}_m,x)]^{\top}.
$

\section{Additional Background}\label{sec:bg_appendix}
\paragraph{Hilbert spaces and linear operators.} $H,H'$ are separable Hilbert spaces. 

\begin{definition}[Bochner $L_q-$spaces, e.g. \cite{diestel77vector}] For $1 \leq q \leq \infty$, $L_q(E_Z,\mathcal{F}_{E_Z},\pi_Z; H)$, abbreviated $L_q(Z; H)$, is the space of strongly measurable and Bochner $q$-integrable functions from $E_Z$ to $H$, with the norms
\[
\| F \|_{L_q(Z;H)}^q 
= \int_{E_Z} \|F \|_H^q \, \mathrm{d}\pi_Z,~q<\infty,~~~\| F \|_{L_{\infty}(Z;H)} = \inf\left\{C \geq0: \pi_Z\{\| F \|_{H}> C\}=0\right\}.
\]
\end{definition}

\begin{definition}[$q$-Schatten class] For $1 \leq q \leq \infty$, $S_q(H,H')$, abbreviated $S_q(H)$ if $H=H'$, is the Banach space of all compact operators $Q$ from $H$ to $H'$ such that $\|Q\|_{S_q(H,H')} \doteq \left\|\left(\sigma_i(Q)\right)_{i \geq 1}\right\|_{\ell_q}$ is finite. Here $\|\left(\sigma_i(Q)\right)_{i \geq 1}\|_{\ell_q}$ is the $\ell_q-$sequence space norm of the (at most countable) sequence of singular values $(\sigma_i(Q))_{i \geq 1}$. For $q = 2$, we retrieve the space of Hilbert-Schmidt operators, for $q = 1$ we retrieve the space of Trace Class operators, and for $q=+\infty$, $\|\cdot\|_{S_\infty(H,H')}$ corresponds to the operator norm $\|\cdot\|_{H \to H'}$.
\end{definition}

\begin{definition}[Tensor Product] \label{def:appendix_tensor} The Hilbert space $H\otimes H'$ is the completion of the algebraic tensor product with respect to the norm induced by the inner product $\langle x_1\otimes x_1', x_2\otimes x_2'\rangle_{H\otimes H'} = \langle x_1,x_2 \rangle_H \langle x_1', x_2'\rangle_{H'}$ for $x_1,x_2 \in H$ and $x_1', x_2' \in H'$ defined on the elementary tensors
of $H\otimes H'$ \citep{aubin2000applied}. This definition extends to
$\operatorname{span}\{x\otimes x'| x\in H, x'\in H'\}$ and finally to
its completion. 
If $\{e_i\}_{i \geq 1}$ and $\{e'_j\}_{j \geq 1}$ are orthonormal basis in $H$ and $H'$ respectively, $\{e_i\otimes e'_j\}_{i \geq 1, j \geq 1}$ is an orthonormal basis in $H \otimes H'$.
\end{definition}

\begin{theorem}[Isomorphism] \label{th:def_psi} The Bochner space $L_2(Z;H)$ is isometrically isomorphic to $S_2(L_2(Z), H)$ and the isometric isomorphism is realized by the map $\Psi:S_2(L_2(Z), H) \to L_2(Z;H)$ acting on elementary tensors as $\Psi(f \otimes h)  = (\omega \to f(\omega)h)$ \citep{aubin2000applied}. 
\end{theorem}
    
\paragraph{RKHS embedding into $L_2$.} Recall that $\cI_{Z}: \cH_Z \to L_2(Z)$ is the embedding that maps every function in $\cH_Z$ into its $\pi_Z$-equivalence class in $L_2(Z)$ and that we use the shorthand notation $[f]_Z = \cI_{Z}(f)$ for all $f \in \cH_Z$. We define similarly $\mathcal{I}_{Z;X}: \cG \to L_2(Z; \cH_X)$ as the embedding that maps every function in $\cG$ into their $\pi_Z$-equivalence class in $L_2(Z; \cH_X)$.

\begin{definition}
    \label{def:embedding} Let $\mathcal{I}_{Z;X}\doteq \operatorname{Id}_{\cH_X} \otimes \cI_{Z}$ be the tensor product of the operator $\operatorname{Id}_{\cH_X}$ with the operator $\cI_{Z}$ (see \citet[Definition 12.4.1.]{aubin2000applied} for the definition of tensor product of operators). $\mathcal{I}_{Z;X}$ maps every function in $\cG$ into their $\pi_Z$-equivalence class in $L_2(Z; \cH_X)$. We then use the shorthand notation $[F]_{Z;X} = \mathcal{I}_{Z;X}(F)$ for all $F \in \cG$. 
\end{definition}

\medskip \noindent
\textbf{Spectral regularization.} 

\medskip \noindent
1. \textit{Ridge regression.} From the Tikhonov filter function $g_{\xi}(x) = (x+\xi)^{-1}$, we obtain the ridge regression algorithm introduced in Eq.~\eqref{eqn:cme_pop}. In this case, we have 
$E = \rho = \omega_\rho = 1$.

\medskip
\noindent
2. \textit{Gradient Descent.} 
From the Landweber iteration filter function
given by
\begin{equation*}
    g_k(x) \doteq \tau \sum_{i=0}^{k-1} 
    (1-\tau x)^i
    \text{ for } k \doteq 1/\xi, k \in \mathbb{N}
\end{equation*}
we obtain the gradient descent scheme with constant step size $\tau >0$, which corresponds to the population gradient iteration given by
$F_{k+1} \doteq F_{k} - \frac{\tau}{2} \nabla_F\left( \E_{X,Z} \left\|\phi_X(X) - F(Z)\right\|^2_{\mathcal{H}_X}\right)(F_k)$ for $k \in \mathbb{N}$. In this case, we have $E = 1$ and
arbitrary qualification with 
$\omega_\rho = 1$
whenever $0 < \rho \leq 1$ and
$\omega_\rho = \rho^\rho$ otherwise.
Gradient schemes with more complex update rules
can be found for example in 
\citet{mucke2019beating,lin2018optimal,lin2020optimal}.

\medskip
\noindent
3. \textit{Kernel principal component regression.} 
The truncation filter function
$g_\xi(x) = x^{-1} \mathds{1}[x \geq\xi]$ yields kernel principal component regression, corresponding to
a hard thresholding of eigenvalues at a
truncation level $\xi$. We have
$E = \omega_\rho = 1$ for arbitrary qualification $\rho$.

\medskip
\noindent
4. \textit{Iterated Tikhonov.} Mixture between Landweber iteration and Tikhonov regularization. Unlike Tikhonov regularization which has finite qualification and cannot exploit the regularity of the solution beyond a certain regularity level, iterated Tikhonov overcomes this problem by means of the following regularization: $g_{\xi, \nu}(x) = \frac{(x+\xi)^{\nu} - \xi^{\nu}}{x(x+\xi)^{\nu}}$ with $\nu > 0$. In this case we have $E = \omega_\rho = 1$ and $\rho = \nu$. For $\nu = 1$, we retrieve the standard Tikhonov regularization and for $\nu \in \mathbb{N}$ we can show that applying $g_{\xi, \nu}$ corresponds to the following iterative procedure: $g_{\xi, 1} = (x + \xi)^{-1}$ and $g_{\xi, k} = (1+\xi g_{\xi, k-1})g_{\xi, 1}, k \geq 2$.


\medskip
\noindent
\textbf{Interpolation spaces.} 

\medskip
\noindent
The interpolation spaces $[\cH_Z]^{\beta}$, $[\cH_X]^{\beta}$ and $[\cG]^{\beta}$ introduced previously correspond to the Hilbert scale generated by the operator $L_Z$, $L_X$ and $\operatorname{Id}_{\cH_X} \otimes L_Z$ respectively. We now give more details on their construction. 
For $\beta \geq 0$, we define the $\beta$-interpolation space \citep{steinwart2012mercer} by
\[[\mathcal{H}_Z]^{\beta}\doteq\left\{\sum_{i \geq 1} a_i\mu_{Z,i}^{\beta / 2}[e_{Z,i}]_Z:\left(a_{i}\right)_{i \geq 1} \in \ell_{2}\right\} \subseteq L_{2}(Z),\]
equipped with the inner product
\[\left\langle \sum_{i \geq 1}a_i(\mu_{Z,i}^{\beta/2}[e_{Z,i}]_Z), \sum_{i \geq 1}b_i(\mu_{Z,i}^{\beta/2}[e_{Z,i}]_Z) \right\rangle_{\beta} = \sum_{i \geq 1} a_i b_i.\]
The $\beta$-interpolation space is a separable Hilbert space with ONB $\left(\mu_{Z,i}^{\beta/2}[e_{Z,i}]_Z\right)_{i \geq 1}$. For $\beta=0$, we have $[\mathcal{H}_Z]^{0}=\overline{\cR (\mathcal{I}_{Z})} \subseteq L_{2}(Z)$ with $\|\cdot\|_{0}=\|\cdot\|_{L_{2}(Z)}$. For $\beta=1$, we have $[\mathcal{H}_Z]^{1}=\cR (\mathcal{I}_{Z})$ and $[\mathcal{H}_Z]^{1}$ is isometrically isomorphic to the closed subspace $\left(\cN \left(\mathcal{I}_{Z}\right)\right)^{\perp}$ of $\mathcal{H}_{Z}$ via $\mathcal{I}_{Z}$, i.e. $\left\|[f]_Z\right\|_{1}=\|f\|_{\mathcal{H}_{Z}}$ for $f \in \left(\cN \left(\mathcal{I}_{Z}\right)\right)^{\perp}$. For $0<\beta<\alpha$, we have
\begin{equation*}
[\mathcal{H}_Z]^{\alpha} \hookrightarrow [\mathcal{H}_Z]^{\beta}  \hookrightarrow [\mathcal{H}_Z]^{0} \subseteq L_{2}(Z). 
\end{equation*}
For $\beta > 0$ and $f \in \overline{\cR (\mathcal{I}_{Z})}$, the $\beta$-interpolation space is given by the image of the fractional integral operator, $[\cH_Z]^\beta = \cR(L_Z^{\beta / 2})$ and $\left\|f\right\|_\beta=\|L_Z^{-\beta / 2}f\|_{L_2(Z)}.$
For a vector-valued function $F \in L_2(Z; \cH_X)$ since $L_2(Z; \cH_X)$ is isometric to $S_2(L_2(Z), \cH_X)$, there is an operator $C \in S_2(L_2(Z), \cH_X)$ such that $\|F\|_{L_2(Z; \cH_X)} = \|C\|_{S_2(L_2(Z), \cH_X)}$. For $C \in S_2(\overline{\cR (\mathcal{I}_{Z})}, \cH_X)$, we define the vector-valued $\beta-$interpolation norm as 
\begin{equation}\label{eq:vv_interpolation_norm}
\|F\|_{\beta} \doteq \|C\|_{\beta} \doteq \|CL_Z^{-\beta / 2}\|_{S_2(L_2(Z), \mathcal{H}_X)}.
\end{equation}
The interpolation space $[\mathcal{H}_X]^{\beta}$ is defined similarly to $[\mathcal{H}_Z]^{\beta}$. For details regarding vector-valued interpolation spaces, we refer to \cite{lietal2022optimal,li2024towards}.

\section{Proof of Proposition~\ref{prop:uni_mini}}\label{sec:proof_minimum_norm_sol}

By Assumption~\ref{asst:exist_rkhs}, the solution set
\(
S\doteq \tilde{\cT}^{-1}(r_0)=\{h\in\cH_X:\tilde{\cT}h=r_0\}
\)
is nonempty. Fix any \(\tilde{h} \in S\). Using the orthogonal decomposition
\(\cH_X=\cN(\tilde{\cT})\oplus \cN(\tilde{\cT})^\perp\), write uniquely $\tilde{h}=h_*+u_0$, $h_*\in \cN(\tilde{\cT})^\perp,\; u_0\in \cN(\tilde{\cT}).$ Since \(\tilde{\cT}u_0=0\), we have \(\tilde{\cT}h_*=\tilde{\cT}\tilde{h}=r_0\), hence
\(h_*\in S\cap \cN(\tilde{\cT})^\perp\).
Now take any \(h\in S\). Then \(\tilde{\cT}(h-\tilde{h})=0\), so \(h-\tilde{h}\in \cN(\tilde{\cT})\).
Therefore \(h=\tilde{h}+u=h_*+(u_0+u)\) for some \(u\in \cN(\tilde{\cT})\), i.e.
\(
S = h_* + \cN(\tilde{\cT}).
\)
In particular, every solution has the same \(\cN(\tilde{\cT})^\perp\)-component \(h_*\).
Finally, for any \(h=h_*+u\in S\) with \(u\in \cN(\tilde{\cT})\), orthogonality yields
\(
\|h\|_{\cH_X}^2=\|h_*\|_{\cH_X}^2+\|u\|_{\cH_X}^2 \ge \|h_*\|_{\cH_X}^2,
\)
with equality if and only if \(u=0\). Hence \(h_*\) is the unique minimum-norm element of \(S\), and
\(
\{h_*\}= S\cap \cN(\tilde{\cT})^\perp.
\)

\section{Link Condition}\label{sec:proof_subsize}

The following basic consequences of \eqref{eq:link_up} are used repeatedly in the analysis. 




\begin{proposition}\label{prop:subsize_appendix}
Let $P_F$ be the orthogonal projection onto $\cN(C_F)^\perp$.
\begin{itemize}
\item[a).] \eqref{eq:link_up} is equivalent to the operator inequality
\(P_F C_X^{\gamma} P_F \preceq C_F.\)

\item[b).] For any $\tilde{\theta} \in[0,1]$,
\(
P_F C_X^{\tilde{\theta}\gamma} P_F \preceq C_F^{\tilde{\theta}}.
\)
\end{itemize}
\end{proposition}

\begin{proof} \textbf{Part (a).} For any $f\in\cN(C_F)^\perp$,
\(
\|C_X^{\gamma/2}f\|_{\cH_X}\le \|C_F^{1/2}f\|_{\cH_X}
\iff
\langle f, C_X^{\gamma}f\rangle_{\cH_X}\le \langle f, C_F f\rangle_{\cH_X}.
\)
Since $P_F f=f$ on $\cN(C_F)^\perp$, the latter is equivalent to
\[
\langle f,\, P_F C_X^{\gamma} P_F f\rangle_{\cH_X}\le \langle f, C_F f\rangle_{\cH_X}
\qquad \forall f\in\cH_X,
\]
which is exactly $P_F C_X^{\gamma} P_F \preceq C_F$.

\medskip\noindent
\textbf{Part (b).} Start from $P_F C_X^{\gamma}P_F\preceq C_F$ and apply the L\"owner--Heinz theorem \citep{heinz1951beitrage}: for $\tilde{\theta} \in [0,1]$, the function $t\mapsto t^{\tilde{\theta}}$ is operator monotone on $[0,\infty)$, and
\(
\big(P_F C_X^{\gamma}P_F\big)^{\tilde{\theta}} \preceq C_F^{\tilde{\theta}}.
\)
It remains to relate $P_F C_X^{\tilde{\theta}\gamma}P_F$ to $\big(P_F C_X^{\gamma}P_F\big)^{\tilde{\theta}}$. 
Since $t\mapsto t^{\tilde{\theta}}$ is operator concave on $[0,\infty)$ for $\tilde{\theta}\in[0,1]$, Jensen's operator inequality \citep[Theorem~V.2.3]{bhatia2013matrix} gives \[ P_F C_X^{\tilde{\theta}\gamma} P_F  = P_F \big(C_X^{\gamma}\big)^{\tilde{\theta}} P_F \preceq \big(P_F C_X^{\gamma} P_F\big)^{\tilde{\theta}}. \qedhere \]
\end{proof}
\section{Proof of \Cref{theo:iv_l2}} \label{sec:proof_ub}

In this section we prove \Cref{theo:iv_l2}, which we give in full detail in \Cref{theo:iv_l2_full} below. 

\begin{theorem} \label{theo:iv_l2_full} For $\tau \geq 1$ and $\lambda \in (0,1]$, we define
\begin{equation} \label{eq:cte_fisher}
\begin{aligned}
\mathcal{N}_F(\lambda) &\doteq \operatorname{Tr}\left(C_F\left(C_F+\lambda \operatorname{Id}_{\mathcal{H}_X}\right)^{-1}\right), \\
g_\lambda & \doteq\log \left(2 e \mathcal{N}_F(\lambda) \frac{\left\|C_F\right\|_{\cH_X \to \cH_X}+\lambda}{\left\|C_F\right\|_{\cH_X \to \cH_X}}\right), \qquad A_{\lambda, \tau} & \doteq8 \tau g_\lambda \lambda^{-1}.\\
\end{aligned}
\end{equation}
Let Assumptions~\ref{asst:rkhs}, \ref{asst:exist_rkhs}, \eqref{eq:link_up} , \eqref{asst:evd}, \eqref{asst:src} and \eqref{asst:mom} hold with $p_X \in (0,1]$ and $1 \leq \beta_X \leq \gamma + 1$ and let Assumptions~\eqref{asst:srcz} and \eqref{asst:embz} hold with $\alpha_Z < \beta_Z$. Condition on the first-stage sample $\cD_1$ used to construct $\hat F_\xi$, sufficiently large $m$ and $n$ such that
\begin{equation} \label{eq:m_con}
\begin{aligned}
    &n \geq A_{\lambda, \tau}, \quad \frac{J\sqrt{\tau}\|F_*-\hat F_{\xi}\|_{\alpha_Z}}{\lambda \sqrt{n}} \leq \frac{1}{12}, \quad\frac{J\|F_*-\hat F_{\xi}\|_{L_2(Z;\cH_X)}}{\lambda} \leq \frac{1}{12}, \\
    &\|F_* - \hat F_{\xi}\|_{L_2(Z;\cH_X)} \leq 1 \qquad \|F_* - \hat F_{\xi}\|_{\alpha_Z} \leq 1,
\end{aligned}
\end{equation}
where $J$ depends on $A_Z,B_Z,\alpha_Z,\beta_Z$, we have with $P^n(\cdot\mid \cD_1)$-probability at least $1-12e^{-\tau}$,
\begin{equation*}
\begin{aligned}
    \|\hat h_{\lambda,\xi} - h_*\|_{L_2(X)} &\leq J_0 \tau  \lambda^{\frac{c_F}{2\gamma} -1} \left(\left\|F_*-  \hat F_{\xi}\right\|_{L_2(Z;\cH_X)} +\frac{\left\| \hat F_{\xi} - F_*\right\|_{\alpha_Z}}{\sqrt{n}} \right)\bigl(\|\bar h_\lambda\|_{\cH_X}+1\bigr)\\
    & + J_1\left(\lambda^{\frac{\beta_X}{2\gamma}} + \sqrt{\frac{\tau}{n}}\lambda^{-\frac{\gamma+p_X-1}{2\gamma}}
+
\,\frac{\tau}{n}\,\lambda^{-1+1/(2\gamma)}\right) 
\end{aligned}
\end{equation*}
where $J_0, J_1$ depend on $\sigma, L, A_Z, B_Z, \alpha_Z, \beta_Z, p_X, B_X$, $\|h_*\|_{\cH_X}$ and $c_F =  \mathrm{1}_{\cN(C_F) = \{0\}}$.
\end{theorem}

\subsection{Analysis Outline} \label{sec:sketch} We start from the decomposition
\begin{equation*}
\|\hat h_{\lambda,\xi} - h_*\|_{L_2} \leq \underbrace{\|\hat h_{\lambda,\xi} - \bar h_{\lambda}\|_{L_2}}_{\text{Stage~1~error}} + \underbrace{\|\bar h_{\lambda} - h_*\|_{L_2}}_{\text{Stage~2~error}},
\end{equation*}
with $\bar h_{\lambda}$ the ideal stage-2 estimator defined in \Cref{eq:iv_bar}. The stage 1 error measures additional error incurred by using features $\hat{F}_{\xi}$ instead of $F_*$. This quantity will be bounded by a function of $m$ (the number of samples for stage 1), via the difference $\hat{F}_{\xi}-F_*$, and $n$ (the number of samples for stage 2). On the other hand, stage 2 error only depends on $n$ and measures how well we approximate $h_*$ by regressing $Y$ on $F_*(Z)$.

\subsubsection{Stage 1 Error}
We start with the observation that we always have
\begin{equation}\label{eq:outline_1}
\|\hat h_{\lambda,\xi} - \bar h_{\lambda} \|_{L_2} =  \|C_X^{\frac{1}{2}}(\hat h_{\lambda,\xi} - \bar h_{\lambda}) \|_{\cH_X} \leq \lambda^{-1/2}\|(C_F+\lambda \operatorname{Id})^{1/2}(\hat h_{\lambda,\xi} - \bar h_{\lambda})\|_{\cH_X},
\end{equation}
where we used $\|C_X\|_{\cH_X \to \cH_X} \leq 1$ (as by \Cref{asst:rkhs}, $k_X(X,X)\leq 1$ a.e.) and $\|(C_F+\lambda \operatorname{Id})^{-1/2}\|_{\cH_X \to \cH_X} \leq \lambda^{-1/2}$. Alternatively, we would like to use \eqref{eq:link_up}, however, we generally cannot ensure that $\hat{h}_{\lambda,\xi} \in \cN(C_F)^{\perp}$ unless $C_F$ is injective, i.e. $\cN(C_F)^{\perp} = \cH_X$. In that case, it follows that $\hat{h}_{\lambda,\xi} \in \cN(C_F)^{\perp}$ and by \eqref{eq:link_up} combined with \Cref{prop:subsize_appendix}-$b)$ applied to $\tilde{\theta} = 1/\gamma \in [0,1]$, we have
\begin{equation}\label{eq:outline_2}
\|C_X^{\frac{1}{2}}(\hat h_{\lambda,\xi} - \bar h_{\lambda}) \|_{\cH_X} \leq  \|C_F^{\frac{1}{2\gamma}}(\hat h_{\lambda,\xi} - \bar h_{\lambda}) \|_{\cH_X} \leq  \lambda^{\frac{1}{2\gamma} - \frac{1}{2}}\|(C_F+\lambda \operatorname{Id})^{1/2}(\hat h_{\lambda,\xi} - \bar h_{\lambda}) \|_{\cH_X},
\end{equation}
where we used \Cref{lma:supl} to obtain $\|C_F^{\frac{1}{2\gamma}}(C_F+\lambda \operatorname{Id})^{-1/2}\|_{\cH_X \to \cH_X} \leq \lambda^{\frac{1}{2\gamma} - \frac{1}{2}}.$
To go further, we use that $\hat h_{\lambda,\xi},\bar h_{\lambda}$, admit the following closed-form expressions (see \Cref{sec:explt}): 
\begin{align}
\hat h_{\lambda,\xi} &= \left( \frac{1}{n}\mathbf{\Phi}_{\hat F}^*\mathbf{\Phi}_{\hat F} + \lambda \operatorname{Id}\right)^{-1}\frac{1}{n}\mathbf{\Phi}_{\hat F}^*Y =\left( \hat C_{\hat F} + \lambda \operatorname{Id}\right)^{-1}\frac{1}{n}\mathbf{\Phi}_{\hat F}^*Y \label{eqn:iv_sol_hat}\\
\bar h_{\lambda} &= \left(\frac{1}{n}\mathbf{\Phi}_{F_*}^*\mathbf{\Phi}_{F_*} + \lambda \operatorname{Id}\right)^{-1}\frac{1}{n}\mathbf{\Phi}_{F_*}^*Y =  \left( \hat C_F + \lambda \operatorname{Id}\right)^{-1}\frac{1}{n}\mathbf{\Phi}_{F_*}^*Y, \label{eqn:iv_sol_bar}
\end{align}
where
$\mathbf{\Phi}_{F_*}:\cH_X \rightarrow \R^n, \mathbf{\Phi}_{F_*} = [F_{*}(z_1), \dots, F_{*}(z_n)]^{*},$
and
$\hat C_{F} = \frac{1}{n}\mathbf{\Phi}_{F_*}^*\mathbf{\Phi}_{F_*} = \frac{1}{n}\sum_{i=1}^n F_{*}( z_i) \otimes F_{*}( z_i).$
Let us define $c_{F} \doteq \mathrm{1}_{\cN(C_F) = \{0\}}$. Combining Eq.~\eqref{eq:outline_1}, Eq.~\eqref{eq:outline_2}, Eq.~\eqref{eqn:iv_sol_hat} and Eq.~\eqref{eqn:iv_sol_bar} yields
\begin{equation*}
\begin{aligned}
\|\hat h_{\lambda,\xi} - \bar h_{\lambda} \|_{L_2} &\leq  \lambda^{\frac{c_F}{2\gamma}-\frac{1}{2}}\left\|(C_F+\lambda \operatorname{Id})^{1/2}\left(\left( \hat C_{\hat F} + \lambda \operatorname{Id}\right)^{-1}\frac{\mathbf{\Phi}_{\hat F}^*Y}{n} -  \left( \hat C_F + \lambda \operatorname{Id}\right)^{-1}\frac{\mathbf{\Phi}_{F_*}^*Y}{n}\right)\right\|_{\cH_X} \\ &\leq \lambda^{\frac{1}{2\gamma}c_F-1/2}\left(S_{-1} + S_0 \right),
\end{aligned}
\end{equation*}
\begin{align}
S_{-1} &\doteq \left\|(C_F+\lambda \operatorname{Id})^{1/2}\left( \hat C_{\hat F} + \lambda \operatorname{Id}\right)^{-1}\left(\frac{1}{n}\mathbf{\Phi}_{\hat F}^*Y -\frac{1}{n}\mathbf{\Phi}_{F_*}^*Y\right) \right\|_{\cH_X} \label{eq:s1a0}\\
S_0 &\doteq \left\|(C_F+\lambda \operatorname{Id})^{1/2}\left(\left( \hat C_{\hat F} + \lambda \operatorname{Id}\right)^{-1}\frac{1}{n}\mathbf{\Phi}_{F_*}^*Y-  \left( \hat C_F + \lambda \operatorname{Id}\right)^{-1}\frac{1}{n}\mathbf{\Phi}_{F_*}^*Y\right)\right\|_{\cH_X}.  \label{eq:s1b0}
\end{align}
$S_{-1}$ and $S_0$ are bounded respectively in Theorem~\ref{theo:s1e1} and Theorem~\ref{theo:s1e2}. Putting them together, we obtain the following bound for the stage 1 error. 

\begin{theorem}\label{theo:s1b}
Let Assumptions~\ref{asst:rkhs}, \eqref{asst:mom}, \eqref{asst:srcz} and \eqref{asst:embz}
hold with $\alpha_Z<\beta_Z$. Condition on the first-stage sample $\cD_1$ used to construct
$\hat F_\xi$. For $\tau\ge 1$ and sufficiently large $m$ and $n$ such that \Cref{eq:m_con}
holds, we have with $P^n(\cdot\mid \cD_1)$-probability at least $1-8e^{-\tau}$,
\[
\|\hat h_{\lambda,\xi}-\bar h_\lambda\|_{L_2}
\le
c_0\,\tau\,\lambda^{\frac{c_F}{2\gamma}-1}
\left(
\|F_*-\hat F_\xi\|_{L_2(Z;\cH_X)}
+
\frac{\|F_*-\hat F_\xi\|_{\alpha_Z}}{\sqrt n}
\right)
\bigl(\|\bar h_\lambda\|_{\cH_X}+1\bigr),
\]
with
\(
c_F\doteq 1_{\mathcal N(C_F)=\{0\}},
\)
and $c_0$ depending on $\sigma,L,A_Z,B_Z,\alpha_Z,\beta_Z$ and
$\|h_*\|_{\cH_X}$.
\end{theorem}

\begin{proof}
By \Cref{eq:outline_1}, \Cref{eq:outline_2}, \Cref{eq:s1a0}, and \Cref{eq:s1b0},
\(
\|\hat h_{\lambda,\xi}-\bar h_\lambda\|_{L_2}
\le
\lambda^{\frac{c_F}{2\gamma}-\frac12}\bigl(S_{-1}+S_0\bigr).
\)
The event used in Theorem~\ref{theo:s1e1} is $\mathcal E_{7}\cap \mathcal E_I$.
On the event $\mathcal E_{7}$, Theorem~\ref{theo:s1e2} also holds. Hence, on the
event of Theorem~\ref{theo:s1e1}, both Theorem~\ref{theo:s1e1} and
Theorem~\ref{theo:s1e2} are simultaneously valid. Therefore, with
$P^n(\cdot\mid \cD_1)$-probability at least $1-8e^{-\tau}$,
\begin{align*}
\|\hat h_{\lambda,\xi}-\bar h_\lambda\|_{L_2}
&\le
\lambda^{\frac{c_F}{2\gamma}-\frac12}
\left(
c\,\frac{\tau}{\sqrt\lambda}
\left(
\frac{\|F_*-\hat F_\xi\|_{\alpha_Z}}{\sqrt n}
+
\|F_*-\hat F_\xi\|_{L_2(Z;\cH_X)}
\right)\right.\\
&\qquad\left.
+
c'\,\frac{\tau}{\sqrt\lambda}
\left(
\frac{\|F_*-\hat F_\xi\|_{\alpha_Z}}{\sqrt n}
+
\|F_*-\hat F_\xi\|_{L_2(Z;\cH_X)}
\right)\|\bar h_\lambda\|_{\cH_X}
\right).
\end{align*}
Absorbing constants finishes the proof.
\end{proof}


\begin{remark}[On the sharpness of the stage-1 transfer bound]
The first-stage learning rate for \(\hat F_\xi\) is minimax optimal by Theorem~12. What is
not shown to be optimal is the transfer step from \(F_*-\hat F_\xi\) to
\(\hat h_{\lambda,\xi}-\bar h_\lambda\). The present argument uses:
(i) operator-norm control of \(\hat C_{\hat F}-\hat C_F\),
(ii) the \(L_\infty\)-envelope supplied by \eqref{asst:embz}, and
(iii) the generic bound \(\|(C_F+\lambda I)^{-1/2}\|\le \lambda^{-1/2}\).
Each of these steps is potentially lossy. In particular, the resulting lower bound on the
first-stage sample size should be interpreted as a sufficient condition rather than as a
proven minimax-sharp transition boundary. A sharper treatment would likely require a
projector-perturbation analysis for the identified subspace and mixed \(L_2(X)\)-metric
covariance perturbation bounds.
\end{remark}

\subsubsection{Stage 2 Error} \label{sec:sketch_stage2}

Recall the oracle stage-2 estimator (see \Cref{eq:iv_bar} and \Cref{eqn:iv_sol_bar})
\[
\bar h_\lambda = \argmin_{h \in \cH_X} \frac{1}{n} \sum_{i = 1}^n \left(y_i - \left\langle h , F_{*}(z_i)\right\rangle_{\mathcal{H}_X} \right)^2 + \lambda \|h\|_{\mathcal{H}_{X}}^2
=
\Big(\hat C_F+\lambda I\Big)^{-1}\frac1n \Phi_{F_*}^* Y,
\]
and its population counterpart from \Cref{eqn:iv_pop}
$$
h_\lambda = \argmin_{h \in \cH_X} \E\left[\left(Y - \langle h, F_*(Z) \rangle_{\cH_X}\right)^2\right] + \lambda \|h\|_{\cH_X}^2 = (C_F + \lambda I)^{-1} C_F h_*.
$$
We bound
\(
\|\bar h_\lambda-h_*\|_{L_2}
\le
\underbrace{\|h_\lambda-h_*\|_{L_2}}_{\text{approximation error}}
+
\underbrace{\|\bar h_\lambda-h_\lambda\|_{L_2}}_{\text{estimation error}}.
\)

\medskip \noindent
\textbf{Step 1: Source transfer and approximation error.}
Let $s\doteq(\beta_X-1)/\gamma\in[0,1]$. Using Proposition~\ref{prop:subsize_appendix}-$b)$ and \Cref{prop:op_1} we show in \Cref{lem:src_transfer_stage2} that the source condition \eqref{asst:src},
$\|C_X^{-(\beta_X-1)/2}h_*\|\le B_X$, implies a \emph{stage-2 source condition}
\(
\|C_F^{-s/2}h_*\|\le B_X.
\)
Then, using $P_F C_XP_F\preceq C_F^{1/\gamma}$ (again from Proposition~\ref{prop:subsize_appendix}-$b)$)
and spectral calculus, we obtain in \Cref{lem:stage2_bias_theta} the bias bound
\(
\|h_\lambda-h_*\|_{L_2}
\lesssim
\lambda^{\beta_X/(2\gamma)}.
\)

\medskip \noindent
\textbf{Step 2: Mixed effective dimension.}
Define the \emph{mixed effective dimension}
\begin{equation}
\label{eq:mxed_N_X_theta}
\mathcal N_{X}(\lambda)
\doteq
\operatorname{Tr}\!\Big(P_F C_XP_F\,(C_F+\lambda I)^{-1}\Big).
\end{equation}
Using \eqref{asst:evd} and the lower link, we show in \Cref{lem:mixed_dim_theta}
\(
\mathcal N_{X}(\lambda)
\lesssim
\lambda^{-(\gamma+p_X-1)/\gamma}.
\)

\medskip \noindent
\textbf{Step 3: Concentration event and stochastic term.}
On the standard covariance concentration event
\[
\mathcal E_\lambda
\doteq
\Big\|(C_F+\lambda I)^{-1/2}(C_F-\hat C_F)(C_F+\lambda I)^{-1/2}\Big\|
\le \frac23,
\]
we have $(\hat C_F+\lambda I)^{-1}\preceq 3(C_F+\lambda I)^{-1}$.
Using a conditional Bernstein inequality (from \eqref{asst:mom}) for the noise term
$\hat\zeta=\frac1n\sum_{i=1}^n \eta_i F_*(z_i)$, where $\eta_i\doteq y_i-\langle h_*,F_*(z_i)\rangle$,
yields
\[
\|\bar h_\lambda-h_\lambda\|_{L_2}
\lesssim
\sigma\sqrt{\frac{\tau\,\mathcal N_{X}(\lambda)}{n}}
+
L\,\frac{\tau}{n}\,\lambda^{-1+1/(2\gamma)}.
\]
Combining steps 1 to 3 gives the stage-2 bound,
and tuning $\lambda$ by balancing the squared bias and variance yields the optimal stage-2 rate. The detailed proof is given in \Cref{thm:stage2_oracle_theta}.

\subsection{Detailed Proof} \label{sec:detailed}
\subsubsection{Stage 2 Error: Detailed Results}
\label{sec:stage2_details}

We now formalize the above sketch. Throughout, let $P_F$ denote the orthogonal projection onto $\cN(C_F)^\perp$. We recall $\|f\|_{L_2}=\|C_X^{1/2}f\|_{\cH_X}$.

\begin{lemma}[Mixed effective dimension bound]
\label{lem:mixed_dim_theta}
Assume \eqref{asst:evd} with exponent $p_X\in(0,1]$ and \eqref{eq:link_up} with exponent $\gamma\ge 1$.
Define $\mathcal N_{X}(\lambda)$ as in \Cref{eq:mxed_N_X_theta}.
Then there exists a constant $c_{\mathcal N}>0$ depending only on $\overline{c}_X,p_X$ such that for all $\lambda\in(0,1]$,
\begin{equation}
\label{eq:mixed_dim_theta_rate}
\mathcal N_{X}(\lambda)
\le
c_{\mathcal N}\,\lambda^{-(\gamma+p_X-1)/\gamma}.
\end{equation}
\end{lemma}

\begin{proof}
Let $A\doteq P_F C_X^{\gamma}P_F$. By \eqref{eq:link_up}, $A\preceq C_F$.
By operator Jensen (as in Proposition~\ref{prop:subsize_appendix}-$b)$),
\(
P_F C_XP_F
=
P_F (C_X^{\gamma})^{1/\gamma}P_F
\preceq
(P_F C_X^{\gamma}P_F)^{1/\gamma}
=
A^{1/\gamma}.
\)
Moreover, since $A\preceq C_F$, Proposition~\ref{prop:op_1} implies the resolvent comparison
$P_F(C_F+\lambda I)^{-1}P_F\preceq P_F(A+\lambda I)^{-1}P_F$.
Therefore,
\(
\mathcal N_{X}(\lambda)
=
\operatorname{Tr}\!\bigl(P_F C_X P_F  (C_F+\lambda I)^{-1}\bigr)
\le
\operatorname{Tr}\!\bigl(A^{1/\gamma}(A+\lambda I)^{-1}\bigr).
\)
Let $(\alpha_j)_{j\ge 1}$ be the nonincreasing eigenvalues of $A$. We obtain 
\(
\mathcal N_{X}(\lambda)
\le
\sum_{j\ge 1}\frac{\alpha_j^{1/\gamma}}{\alpha_j+\lambda}.
\)
Set
\(
C\doteq \overline{c}_X^\gamma,
J\doteq\left\lceil (C/\lambda)^{p_X/\gamma}\right\rceil.
\)
Since $A$ is a compression of $C_X^{\gamma}$ to $\cN(C_F)^\perp$, Courant--Fischer and \eqref{asst:evd} yield
$\alpha_j\le \mu_{X,j}^{\gamma}\le Cj^{-\gamma/p_X}$. For $p_X=1$, we have
$
\mathcal N_{X}(\lambda) \le \lambda^{-1} \operatorname{Tr}(A^{1/\gamma} ) \leq \operatorname{Tr}(C_X)\lambda^{-1} \leq \lambda^{-1}.
$
This is \Cref{eq:mixed_dim_theta_rate}, with $c_{\mathcal N}\doteq 1$. For $p_X<1$, we split the series as
\[
\sum_{j\ge1}\frac{\alpha_j^{\frac{1}{\gamma}}}{\alpha_j+\lambda}
=
\sum_{j\le J}\frac{\alpha_j^{\frac{1}{\gamma}}}{\alpha_j+\lambda}
+
\sum_{j>J}\frac{\alpha_j^{\frac{1}{\gamma}}}{\alpha_j+\lambda} \doteq S_1+S_2.
\]
We first bound $S_1$. For every $x\ge 0$, setting $y\doteq x/\lambda\ge 0$, we have
\(
\frac{x^\frac{1}{\gamma}}{x+\lambda}
=
\lambda^{\frac{1}{\gamma}-1}\frac{y^{\frac{1}{\gamma}}}{1+y}.
\)
Since $\frac{1}{\gamma}\in[0,1]$, one has $y^{\frac{1}{\gamma}}\le \max(1,y)\le 1+y$, hence
\(
y^{\frac{1}{\gamma}}/(1+y)\le 1.
\)
Therefore
\(
\frac{x^{\frac{1}{\gamma}}}{x+\lambda}\le \lambda^{\frac{1}{\gamma}-1},
\)
and
\[
S_1
\le
J\lambda^{\frac{1}{\gamma}-1}
\le
\Bigl(1+(C/\lambda)^{p_X/\gamma}\Bigr)\lambda^{\frac{1}{\gamma}-1}.
\]
Let
\(
r\doteq1+\frac{p_X}{\gamma}-\frac{1}{\gamma}
=
\frac{\gamma+p_X-1}{\gamma}.
\)
Since $\lambda\in(0,1]$, we have $\lambda^{\frac{1}{\gamma}-1}\le \lambda^{-r}$, and also
\(
(C/\lambda)^{p_X/\gamma}\lambda^{\frac{1}{\gamma}-1}
=
C^{p_X/\gamma}\lambda^{-r}.
\)
Thus
\(
S_1
\le
\bigl(1+\overline{c}_X^{p_X}\bigr)\lambda^{-r}.
\)
We now bound $S_2$.
\(
\frac{\alpha_j^\frac{1}{\gamma}}{\alpha_j+\lambda}
\le
\lambda^{-1}\alpha_j^{\frac{1}{\gamma}}
\le
\lambda^{-1}C^{\frac{1}{\gamma}} j^{-1/p_X}.
\)
Assume first $p_X<1$. Using the integral test,
\[
\sum_{j>J}j^{-\frac{1}{p_X}}
\le
\int_J^\infty x^{-\frac{1}{p_X}}\,dx
=
\frac{J^{1-\frac{1}{p_X}}}{\frac{1}{p_X}-1}.
\]
Therefore
\(
S_2
\le
\frac{p_XC^{\frac{1}{\gamma}}}{1-p_X}\lambda^{-1}J^{1-\frac{1}{p_X}}.
\)
Since $J\ge (C/\lambda)^{p_X/\gamma}$,
\(
J^{1-\frac{1}{p_X}}
\le
(C/\lambda)^{\frac{p_X-1}{\gamma}}.
\)
Hence
\[
S_2
\le
\frac{p_XC^{\frac{1}{\gamma}}}{1-p_X}\lambda^{-1}
(C/\lambda)^{\frac{p_X-1}{\gamma}} = \frac{p_X\overline{c}_X^{p_X}}{1-p_X}\lambda^{-r}.
\]
Combining the bounds for $S_1$ and $S_2$, we obtain
\(
\mathcal N_{X}(\lambda)
\le
\left(
1+\overline{c}_X^{p_X}
+\frac{p_X\overline{c}_X^{p_X}}{1-p_X}
\right)\lambda^{-r}.
\)
This is exactly \Cref{eq:mixed_dim_theta_rate}, with
\(
c_{\mathcal N}
\doteq
1+\overline{c}_X^{p_X}
+\frac{p_X\overline{c}_X^{p_X}}{1-p_X}.
\)
\end{proof}

\begin{lemma}[Stage-2 source transfer]
\label{lem:src_transfer_stage2}
Assume \eqref{eq:link_up} with exponent $\gamma \ge 1$ and \eqref{asst:src} with $1\le \beta_X\le \gamma+1$, 
Let $s\doteq (\beta_X-1)/\gamma\in[0,1]$. Then $\|C_F^{-s/2}h_*\|_{\cH_X}\le B_X.$
\end{lemma}

\begin{proof}
By Proposition~\ref{prop:subsize_appendix}-$b)$ applied with $\tilde\theta=s$,
$P_F C_X^{\beta_X-1}P_F\preceq C_F^{s}$.
Apply \Cref{prop:op_1} with $A=C_X^{\beta_X-1}$ and $B=C_F^s$ to obtain
$\langle h_*,(C_F^s)^\dagger h_*\rangle\le \langle h_*,(C_X^{\beta_X-1})^\dagger h_*\rangle$,
which concludes the proof.
\end{proof}

\begin{lemma}[Approximation Error Bound]
\label{lem:stage2_bias_theta}
Assume the conditions of Lemma~\ref{lem:src_transfer_stage2}. Then for all $\lambda\in(0,1]$,
$
\|h_\lambda-h_*\|_{L_2}
\le
B_X\,\lambda^{\frac{\beta_X}{2\gamma}}.
$
\end{lemma}

\begin{proof}
We have $h_\lambda-h_*=-\lambda(C_F+\lambda I)^{-1}h_*$ and by Lemma~\ref{lem:src_transfer_stage2},
$h_*=C_F^{s/2}u$ for some $\|u\|_{\cH_X}\le B_X$.
Moreover, by Proposition~\ref{prop:subsize_appendix}-$b)$,
$P_F C_XP_F\preceq C_F^{1/\gamma}$, hence using $h_\lambda,h_*$ in $\cN(C_F)^\perp$,
\[
\|h_\lambda-h_*\|_{L_2}
=\|C_X^{1/2}(h_\lambda-h_*)\|_{\cH_X}
\le
\|C_F^{1/(2\gamma)}(h_\lambda-h_*)\|_{\cH_X}.
\]
Therefore,
\(
\|h_\lambda-h_*\|_{L_2}
\le
B_X \lambda\,
\big\|(C_F+\lambda I)^{-1}C_F^{1/(2\gamma)}C_F^{s/2}\big\| = B_X \lambda\,
\big\|(C_F+\lambda I)^{-1}C_F^{\frac{\beta_X}{2\gamma}}\big\|.
\)
By \Cref{lma:supl},
\(
\big\|(C_F+\lambda I)^{-1}C_F^{\frac{\beta_X}{2\gamma}}\big\| = \sup_{t\ge0}\frac{t^{\frac{\beta_X}{2\gamma}}}{t+\lambda}
\le
\lambda^{\frac{\beta_X}{2\gamma}-1}.
\)
and the claim follows.
\end{proof}

\begin{theorem}[Stage-2 oracle bound]
\label{thm:stage2_oracle_theta}
Assume Assumptions~\ref{asst:rkhs}, \ref{asst:exist_rkhs}, \eqref{eq:link_up}, \eqref{asst:evd}, \eqref{asst:src}, \eqref{asst:mom}
with $p_X\in(0,1]$ and $1\le \beta_X\le \gamma+1$.
Fix $\tau\ge 1$, $\lambda\in(0,1]$.
If $n\ge 8\tau g_\lambda \lambda^{-1}$, with $g_\lambda$ as in \Cref{eq:cte_fisher}, then with $P^n$-probability at least $1-4e^{-\tau}$,
\begin{equation*}
\|\bar h_\lambda-h_*\|_{L_2}
\le
C\Bigg[
\lambda^{\frac{\beta_X}{2\gamma}}
+
\sigma\sqrt{\frac{\tau\,\mathcal N_{X}(\lambda)}{n}}
+
L\,\frac{\tau}{n}\,\lambda^{-1+1/(2\gamma)}
\Bigg],
\end{equation*}
where $\mathcal N_{X}(\lambda)$ is defined in \Cref{eq:mxed_N_X_theta} and $C$ depends only on $B_X,\sigma,L$ and fixed problem constants.
Moreover, using Lemma~\ref{lem:mixed_dim_theta},
\begin{equation*}
\|\bar h_\lambda-h_*\|_{L_2}
\le
C\Bigg[
\lambda^{\frac{\beta_X}{2\gamma}}
+
\sqrt{\frac{\tau}{n}}\lambda^{-\frac{\gamma+p_X-1}{2\gamma}}
+
L\,\frac{\tau}{n}\,\lambda^{-1+1/(2\gamma)}
\Bigg].
\end{equation*}
\end{theorem}

\begin{proof}
Write
$F_i\doteq F_*(z_i), \hat C_F=\frac1n\sum_{i=1}^n F_i\otimes F_i, \eta_i\doteq y_i-\langle h_*,F_i\rangle,$
and define $\hat\zeta\doteq \frac1n\sum_{i=1}^n \eta_i F_i$. Then
\[
\bar h_\lambda=(\hat C_F+\lambda I)^{-1}(\hat C_F h_*+\hat\zeta),
\qquad
h_\lambda=(C_F+\lambda I)^{-1}C_F h_*.
\]
Since
\(
h_\lambda=h_*-\lambda(C_F+\lambda I)^{-1}h_*.
\)
We have
\begin{equation*}
    \begin{aligned}
(\hat C_F+\lambda I)&(\bar h_\lambda-h_\lambda) = \hat C_F h_* + \hat\zeta - (\hat C_F+\lambda I)(h_* - \lambda (C_F+\lambda I)^{-1}h_*) \\ &= \hat\zeta -\lambda h_* + \lambda (\hat C_F+\lambda I)(C_F+\lambda I)^{-1}h_* = \hat\zeta -\lambda h_* + (\hat C_F+\lambda I)(h_* - h_\lambda) \\ &= \hat\zeta -\lambda h_\lambda + \hat C_F(h_* - h_\lambda) = \hat\zeta + (\hat C_F - C_F)(h_* - h_\lambda).
\end{aligned}
\end{equation*}
Where in the last equality, we used $\lambda h_\lambda = C_F(h_* - h_\lambda)$. Hence
\[
\bar h_\lambda-h_\lambda
=
(\hat C_F+\lambda I)^{-1}\hat\zeta
-
(\hat C_F+\lambda I)^{-1}(\hat C_F-C_F)(h_\lambda-h_*).
\]
Set
\(
T_{1,\lambda}\doteq (\hat C_F+\lambda I)^{-1}\hat\zeta,
T_{2,\lambda}\doteq -(\hat C_F+\lambda I)^{-1}(\hat C_F-C_F)(h_\lambda-h_*).
\)

\paragraph{Preliminary facts.}
For every \(f\in \cN(C_F)\),
\(
0
=
\langle f,C_Ff\rangle_{\cH_X}
=
\E\!\left[\langle f,F_*(Z)\rangle_{\cH_X}^2\right].
\)
Therefore \(\langle f,F_*(Z)\rangle_{\cH_X}=0\) almost surely for every \(f\in \cN(C_F)\), and thus
\(
F_i\in \cN(C_F)^\perp
\)
a.s. Equivalently,
\(
P_FF_i=F_i
\)
a.s. Hence
\(
P_F\hat C_F=\hat C_F=\hat C_F P_F.
\)
Also, by the identification convention used throughout the paper, \(h_*\in \cN(C_F)^\perp\), and clearly
\[
h_\lambda=(C_F+\lambda I)^{-1}C_F h_*\in R(C_F)\subseteq \cN(C_F)^\perp.
\]
Thus
\(
P_F(h_\lambda-h_*)=h_\lambda-h_*.
\)
Next define
\[
\Delta_\lambda
\doteq
(C_F+\lambda I)^{-1/2}(C_F-\hat C_F)(C_F+\lambda I)^{-1/2}.
\]
We have
\(
\hat C_F+\lambda I
=
(C_F+\lambda I)^{1/2}(I-\Delta_\lambda)(C_F+\lambda I)^{1/2}.
\)
On the event
\[
\mathcal E_\lambda
\doteq
\left\{\|\Delta_\lambda\|_{\cH_X\to\cH_X}\le \frac23\right\},
\]
we have
\(
I-\Delta_\lambda \succeq \frac13 I,
\)
and therefore
\(
(I-\Delta_\lambda)^{-1}\preceq 3I.
\)
Consequently,
\begin{equation}
\label{eq:oracle_resolvent_compare}
(\hat C_F+\lambda I)^{-1}
=
(C_F+\lambda I)^{-1/2}(I-\Delta_\lambda)^{-1}(C_F+\lambda I)^{-1/2}
\preceq
3(C_F+\lambda I)^{-1}.
\end{equation}
$\mathcal E_\lambda$ holds with probability $\ge 1-2e^{-\tau}$ under $n\ge 8\tau g_\lambda \lambda^{-1}$ by \Cref{lma:con_cf}. We now prove two operator bounds that will be used repeatedly. Set
\(
B_\lambda
\doteq
C_X^{1/2}P_F(\hat C_F+\lambda I)^{-1/2}P_F.
\)
Because \(P_F\hat C_F=\hat C_F=\hat C_F P_F\), the projection \(P_F\) commutes with every bounded Borel function of \(\hat C_F\), and hence with \((\hat C_F+\lambda I)^{-1/2}\). Therefore
\(
B_\lambda B_\lambda^*
=
C_X^{1/2}P_F(\hat C_F+\lambda I)^{-1}P_FC_X^{1/2}.
\)
Using \Cref{eq:oracle_resolvent_compare},
\(
B_\lambda B_\lambda^*
\preceq
3\,C_X^{1/2}P_F(C_F+\lambda I)^{-1}P_FC_X^{1/2}.
\)
Thus
\begin{equation}
\label{eq:prelim_opnorm_start}
\|B_\lambda\|_{\cH_X\to\cH_X}^2
=
\|B_\lambda B_\lambda^*\|_{\cH_X\to\cH_X}
\le
3\,\bigl\|C_X^{1/2}P_F(C_F+\lambda I)^{-1}P_FC_X^{1/2}\bigr\|_{\cH_X\to\cH_X}.
\end{equation}
Now set
\(
M_\lambda^{(0)}\doteq C_X^{1/2}P_F(C_F+\lambda I)^{-1/2}.
\)
Then
\[
M_\lambda^{(0)}(M_\lambda^{(0)})^*
=
C_X^{1/2}P_F(C_F+\lambda I)^{-1}P_FC_X^{1/2},
\]
while
\(
(M_\lambda^{(0)})^*M_\lambda^{(0)}
=
(C_F+\lambda I)^{-1/2}P_FC_XP_F(C_F+\lambda I)^{-1/2}.
\)
Hence
\[
\bigl\|C_X^{1/2}P_F(C_F+\lambda I)^{-1}P_FC_X^{1/2}\bigr\|_{\cH_X\to\cH_X}
=
\bigl\|(C_F+\lambda I)^{-1/2}P_FC_XP_F(C_F+\lambda I)^{-1/2}\bigr\|_{\cH_X\to\cH_X}.
\]
By Proposition~\ref{prop:subsize_appendix}-$b)$ with \(\tilde\theta=1/\gamma\),
\(
P_FC_XP_F\preceq C_F^{1/\gamma}.
\)
Therefore
\[
(C_F+\lambda I)^{-1/2}P_FC_XP_F(C_F+\lambda I)^{-1/2}
\preceq
(C_F+\lambda I)^{-1/2}C_F^{1/\gamma}(C_F+\lambda I)^{-1/2}.
\]
Taking operator norms and using \Cref{lma:supl},
\[
\bigl\|(C_F+\lambda I)^{-1/2}P_FC_XP_F(C_F+\lambda I)^{-1/2}\bigr\|_{\cH_X\to\cH_X}
\le
\|C_F^{1/\gamma}(C_F+\lambda I)^{-1}\|_{\cH_X\to\cH_X}
\le
\lambda^{1/\gamma-1}.
\]
Returning to \Cref{eq:prelim_opnorm_start}, we obtain
\begin{equation}
\label{eq:prelim_opnorm}
\|C_X^{1/2}P_F(\hat C_F+\lambda I)^{-1/2}P_F\|_{\cH_X\to\cH_X}^2
\le
3\,\lambda^{1/\gamma-1}.
\end{equation}

\paragraph{Bound on \(T_{1,\lambda}\).}
Define
\(
v_i
\doteq
C_X^{1/2}P_F(\hat C_F+\lambda I)^{-1}F_i, i=1,\dots,n.
\)
Then
\[
C_X^{1/2}T_{1,\lambda}
=
\frac1n\sum_{i=1}^n \eta_i v_i.
\]
Conditionally on \(z_1,\dots,z_n\), the vectors \(v_i\) are deterministic. Moreover,
\[
\E[\eta_i\mid z_1,\dots,z_n]
=
\E[\eta_i\mid z_i]
=
0,
\]
and by \eqref{asst:mom}, for every integer \(q\ge2\),
\(
\E\!\left[|\eta_i|^q\mid z_1,\dots,z_n\right]
=
\E\!\left[|\eta_i|^q\mid z_i\right]
\le
\frac{q!}{2}\sigma^2L^{q-2}.
\)
Let
\(
\xi_i\doteq \eta_i v_i,
M_\lambda\doteq \max_{1\le i\le n}\|v_i\|_{\cH_X}.
\)
Then for every \(q\ge2\),
\begin{align*}
\E\!\left[\|\xi_i\|_{\cH_X}^q\mid z_1,\dots,z_n\right]
&=
 \|v_i\|_{\cH_X}^q \E\!\left[|\eta_i|^q\mid z_i\right]\le
\frac{q!}{2}\sigma^2L^{q-2} \|v_i\|_{\cH_X}^q\\
&\le
\frac{q!}{2}
\left(\sigma^2 \|v_i\|_{\cH_X}^2\right)
\left(LM_\lambda\right)^{q-2},
\end{align*}
because \(\|v_i\|^q\le M_\lambda^{q-2}\|v_i\|^2\). Hence the Hilbert-space Bernstein inequality (\Cref{theo:ope_con_steinwart}), applied conditionally on \(z_1,\dots,z_n\), gives an event \(\mathcal B_\lambda\) such that
\(
\Pr(\mathcal B_\lambda^c\mid z_1,\dots,z_n)\le 2e^{-\tau},
\)
and on \(\mathcal B_\lambda\),
\begin{equation}
\label{eq:T1_conditional_bernstein_full}
\|T_{1,\lambda}\|_{L_2}
\le
\frac{1}{n}\sqrt{2\sigma^2 \tau \sum_{i=1}^n \|v_i\|_{\cH_X}^2}
+
\frac{2L\tau}{n}M_\lambda.
\end{equation}
We now control the two quantities in \Cref{eq:T1_conditional_bernstein_full} on \(\mathcal E_\lambda\). For the quadratic term, using \(F_i=P_FF_i\) and \(\hat C_F=\frac1n\sum_{i=1}^n F_i\otimes F_i\),
\begin{align*}
\frac1n\sum_{i=1}^n \|v_i\|_{\cH_X}^2
&=
\frac1n\sum_{i=1}^n
\left\langle
F_i,
(\hat C_F+\lambda I)^{-1}P_FC_XP_F(\hat C_F+\lambda I)^{-1}F_i
\right\rangle_{\cH_X}\\
&=
\operatorname{Tr}\!\left(
(\hat C_F+\lambda I)^{-1}P_FC_XP_F(\hat C_F+\lambda I)^{-1}\hat C_F
\right)\\
&=
\operatorname{Tr}\!\left(
P_FC_XP_F(\hat C_F+\lambda I)^{-1}\hat C_F(\hat C_F+\lambda I)^{-1}
\right).
\end{align*}
Now the scalar inequality
\(
\frac{t}{(t+\lambda)^2}\le \frac{1}{t+\lambda}, t\ge0,
\)
implies, by functional calculus,
\[
(\hat C_F+\lambda I)^{-1}\hat C_F(\hat C_F+\lambda I)^{-1}
=
\hat C_F(\hat C_F+\lambda I)^{-2}
\preceq
(\hat C_F+\lambda I)^{-1}.
\]
Therefore, by trace monotonicity,
\(
\frac1n\sum_{i=1}^n \|v_i\|_{\cH_X}^2
\le
\operatorname{Tr}\!\left(
P_FC_XP_F(\hat C_F+\lambda I)^{-1}
\right).
\)
Using \Cref{eq:oracle_resolvent_compare} on \(\mathcal E_\lambda\),
\[
\frac1n\sum_{i=1}^n \|v_i\|_{\cH_X}^2
\le
3\,\operatorname{Tr}\!\left(
P_FC_XP_F(C_F+\lambda I)^{-1}
\right)
=
3\,\mathcal N_X(\lambda).
\]
Hence
\begin{equation}
\label{eq:T1_variance_proxy}
\frac{\sigma\sqrt{\tau}}{n}\sqrt{\sum_{i=1}^n \|v_i\|_{\cH_X}^2}
\le
\sqrt{3}\,\sigma\sqrt{\frac{\tau\,\mathcal N_X(\lambda)}{n}}.
\end{equation}
We next control \(M_\lambda\). Write
\(
v_i
=
\Big(C_X^{1/2}P_F(\hat C_F+\lambda I)^{-1/2}P_F\Big)
\Big((\hat C_F+\lambda I)^{-1/2}F_i\Big).
\)
Therefore
\[
\|v_i\|_{\cH_X}
\le
\|C_X^{1/2}P_F(\hat C_F+\lambda I)^{-1/2}P_F\|_{\cH_X\to\cH_X}\,
\|(\hat C_F+\lambda I)^{-1/2}F_i\|_{\cH_X}.
\]
By \Cref{eq:prelim_opnorm},
\(
\|C_X^{1/2}P_F(\hat C_F+\lambda I)^{-1/2}P_F\|_{\cH_X\to\cH_X}
\le
\sqrt{3}\,\lambda^{-1/2+1/(2\gamma)}.
\)
Also,
\[
\|(\hat C_F+\lambda I)^{-1/2}F_i\|_{\cH_X}^2
=
\langle F_i,(\hat C_F+\lambda I)^{-1}F_i\rangle_{\cH_X}
\le
\lambda^{-1}\|F_i\|_{\cH_X}^2.
\]
By Assumption~\ref{asst:rkhs} and Jensen's inequality,
\(
\|F_i\|_{\cH_X}=\|F_*(z_i)\|_{\cH_X}\le 1
\)
a.s. Hence
\[
\|(\hat C_F+\lambda I)^{-1/2}F_i\|_{\cH_X}\le \lambda^{-1/2},
\]
and thus
\begin{equation}
\label{eq:T1_max_bound}
M_\lambda
\le
\sqrt{3}\,\lambda^{-1+1/(2\gamma)}.
\end{equation}
Substituting \Cref{eq:T1_variance_proxy} and \Cref{eq:T1_max_bound} into \Cref{eq:T1_conditional_bernstein_full}, we obtain on \(\mathcal E_\lambda\cap\mathcal B_\lambda\),
\[
\|T_{1,\lambda}\|_{L_2}
\le
\sqrt{6}\,\sigma\sqrt{\frac{\tau\,\mathcal N_X(\lambda)}{n}}
+
2\sqrt{3}\,L\,\frac{\tau}{n}\lambda^{-1+1/(2\gamma)}.
\]
Absorbing the numerical constants into \(C\), this becomes
\begin{equation}
\label{eq:T1_final_full}
\|T_{1,\lambda}\|_{L_2}
\le
C\left[
\sigma\sqrt{\frac{\tau\,\mathcal N_X(\lambda)}{n}}
+
L\,\frac{\tau}{n}\lambda^{-1+1/(2\gamma)}
\right].
\end{equation}

\paragraph{Bound on \(T_{2,\lambda}\).}
Set
\(
d_\lambda\doteq h_\lambda-h_*.
\)
Since \(d_\lambda\in \cN(C_F)^\perp\), and \(\hat C_F-C_F=P_F(\hat C_F-C_F)P_F\), we may write
\(
T_{2,\lambda}
=
-P_F(\hat C_F+\lambda I)^{-1}P_F(\hat C_F-C_F)P_F d_\lambda.
\)
Therefore
\begin{align*}
\|&T_{2,\lambda}\|_{L_2}
=
\|C_X^{1/2}T_{2,\lambda}\|_{\cH_X} \le
\|C_X^{1/2}P_F(\hat C_F+\lambda I)^{-1/2}P_F\|_{\cH_X\to\cH_X}\\
&\qquad\times
\|P_F(\hat C_F+\lambda I)^{-1/2}P_F(\hat C_F-C_F)(C_F+\lambda I)^{-1/2}P_F\|_{\cH_X\to\cH_X}
\|(C_F+\lambda I)^{1/2}d_\lambda\|_{\cH_X}.
\end{align*}
We bound the three factors separately. The first one is already bounded by \Cref{eq:prelim_opnorm}:
\begin{equation}
\label{eq:T2_factor1}
\|C_X^{1/2}P_F(\hat C_F+\lambda I)^{-1/2}P_F\|_{\cH_X\to\cH_X}
\le
\sqrt{3}\,\lambda^{-1/2+1/(2\gamma)}.
\end{equation}
For the second one, recall that
\(
\hat C_F-C_F
=
-(C_F+\lambda I)^{1/2}\Delta_\lambda(C_F+\lambda I)^{1/2}.
\)
Hence
\begin{align*}
&\|P_F(\hat C_F+\lambda I)^{-1/2}P_F(\hat C_F-C_F)(C_F+\lambda I)^{-1/2}P_F\|_{\cH_X\to\cH_X}\\
&\qquad=
\|P_F(\hat C_F+\lambda I)^{-1/2}(C_F+\lambda I)^{1/2}\Delta_\lambda P_F\|_{\cH_X\to\cH_X}\\
&\qquad\le
\|P_F(\hat C_F+\lambda I)^{-1/2}(C_F+\lambda I)^{1/2}P_F\|_{\cH_X\to\cH_X}\,
\|\Delta_\lambda\|_{\cH_X\to\cH_X}.
\end{align*}
On \(\mathcal E_\lambda\), \(\|\Delta_\lambda\|\le 2/3\). Moreover,
\begin{align*}
&\|P_F(\hat C_F+\lambda I)^{-1/2}(C_F+\lambda I)^{1/2}P_F\|_{\cH_X\to\cH_X}^2\\
&\qquad=
\|(C_F+\lambda I)^{1/2}P_F(\hat C_F+\lambda I)^{-1}P_F(C_F+\lambda I)^{1/2}\|_{\cH_X\to\cH_X} \le 3,
\end{align*}
again by \Cref{eq:oracle_resolvent_compare}. Therefore, on \(\mathcal E_\lambda\),
\begin{equation}
\label{eq:T2_factor2}
\|P_F(\hat C_F+\lambda I)^{-1/2}P_F(\hat C_F-C_F)(C_F+\lambda I)^{-1/2}P_F\|_{\cH_X\to\cH_X}
\le
\frac{2\sqrt{3}}{3}.
\end{equation}
For the third factor, let
\(
s\doteq (\beta_X-1)/\gamma \in[0,1].
\)
By Lemma~\ref{lem:src_transfer_stage2}, there exists \(u\in\cH_X\) such that
\(
h_*=C_F^{s/2}u,
\|u\|_{\cH_X}\le B_X.
\)
Since
\[
d_\lambda
=
h_\lambda-h_*
=
-\lambda(C_F+\lambda I)^{-1}h_*
=
-\lambda(C_F+\lambda I)^{-1}C_F^{s/2}u,
\]
we get
\(
(C_F+\lambda I)^{1/2}d_\lambda
=
-\lambda(C_F+\lambda I)^{-1/2}C_F^{s/2}u.
\)
Hence
\[
\|(C_F+\lambda I)^{1/2}d_\lambda\|_{\cH_X}
\le
\lambda\,
\sup_{t\ge0}\frac{t^{s/2}}{(t+\lambda)^{1/2}}
\|u\|_{\cH_X}.
\]
For \(s\in[0,1]\), the inequality \(x^s\le 1+x\) for all \(x\ge0\) implies
\[
\frac{t^{s/2}}{(t+\lambda)^{1/2}}
=
\lambda^{s/2-1/2}\frac{(t/\lambda)^{s/2}}{(1+t/\lambda)^{1/2}}
\le
\lambda^{s/2-1/2}.
\]
Therefore
\begin{equation}
\label{eq:T2_factor3}
\|(C_F+\lambda I)^{1/2}d_\lambda\|_{\cH_X}
\le
B_X\,\lambda^{1/2+s/2}.
\end{equation}
Combining \Cref{eq:T2_factor1}, \Cref{eq:T2_factor2}, and \Cref{eq:T2_factor3}, we obtain on \(\mathcal E_\lambda\),
\[
\|T_{2,\lambda}\|_{L_2}
\le
\sqrt{3}\,\lambda^{-1/2+1/(2\gamma)}
\cdot
\frac{2\sqrt{3}}{3}
\cdot
B_X\,\lambda^{1/2+s/2}
=
2B_X\,\lambda^{1/(2\gamma)+s/2}.
\]
Since \(s=(\beta_X-1)/\gamma\),
\(
\frac{1}{2\gamma}+\frac{s}{2}
=
\frac{1}{2\gamma}+\frac{\beta_X-1}{2\gamma}
=
\frac{\beta_X}{2\gamma}.
\)
Thus
\begin{equation}
\label{eq:T2_final_full}
\|T_{2,\lambda}\|_{L_2}
\le
2B_X\,\lambda^{\beta_X/(2\gamma)}.
\end{equation}
\paragraph{Conclusion.}
On \(\mathcal E_\lambda\cap\mathcal B_\lambda\), by Lemma~\ref{lem:stage2_bias_theta}, \Cref{eq:T1_final_full}, and \Cref{eq:T2_final_full},
\begin{align*}
\|\bar h_\lambda-h_*\|_{L_2}
&\le
\|h_\lambda-h_*\|_{L_2}
+
\|T_{1,\lambda}\|_{L_2}
+
\|T_{2,\lambda}\|_{L_2}\\
&\le
C\Bigg[
\lambda^{\frac{\beta_X}{2\gamma}}
+
\sigma\sqrt{\frac{\tau\,\mathcal N_X(\lambda)}{n}}
+
L\,\frac{\tau}{n}\lambda^{-1+1/(2\gamma)}
\Bigg].
\end{align*}
By \Cref{lma:con_cf}, if \(n\ge 8\tau g_\lambda\lambda^{-1}\), then
\(
\Pr(\mathcal E_\lambda^c)\le 2e^{-\tau}.
\)
Also,
\(
\Pr(\mathcal B_\lambda^c)
=
\E\!\left[\Pr(\mathcal B_\lambda^c\mid z_1,\dots,z_n)\right]
\le
2e^{-\tau}.
\)
Therefore
\(
\Pr(\mathcal E_\lambda\cap\mathcal B_\lambda)
\ge
1-4e^{-\tau}.
\)
This proves
\[
\|\bar h_\lambda-h_*\|_{L_2}
\le
C\Bigg[
\lambda^{\frac{\beta_X}{2\gamma}}
+
\sigma\sqrt{\frac{\tau\,\mathcal N_X(\lambda)}{n}}
+
L\,\frac{\tau}{n}\lambda^{-1+1/(2\gamma)}
\Bigg].
\]
Finally, Lemma~\ref{lem:mixed_dim_theta} yields
\(
\mathcal N_X(\lambda)\le c_{\mathcal N}\lambda^{-(\gamma+p_X-1)/\gamma},
\)
so
\[
\|\bar h_\lambda-h_*\|_{L_2}
\le
C\Bigg[
\lambda^{\frac{\beta_X}{2\gamma}}
+
\sigma\sqrt{\frac{\tau}{n}}\lambda^{-\frac{\gamma+p_X-1}{2\gamma}}
+
L\,\frac{\tau}{n}\lambda^{-1+1/(2\gamma)}
\Bigg]. \qedhere
\]
\end{proof}
\begin{lemma}[Control of \(\|\bar h_\lambda\|_{\cH_X}\)] \label{lem:rkhs_bound_s2}
Assume the conditions of \Cref{thm:stage2_oracle_theta}. Fix \(\tau \ge 1\) and \(\lambda \in (0,1]\). If
\(n \ge 8\tau g_\lambda \lambda^{-1}\), then with \(P^n\)-probability at least \(1-4e^{-\tau}\),
\[
\|\bar h_\lambda\|_{\cH_X}
\le
c_0
\left(
1
+
\sigma\sqrt{\frac{\tau \cN_F(\lambda)}{n\lambda}}
+
L\frac{\tau}{n\lambda}
\right),
\]
where \(c_0\) depends only on \(\|h_\ast\|_{\cH_X},\sigma,L\) and fixed problem constants. Since \(C_F \preceq C_X\), \eqref{asst:evd} implies
\(
\cN_F(\lambda) \lesssim \lambda^{-p_X}.
\)
Hence, for any sequence \(\lambda_n = n^{-\ell}\) with \(\ell \in (0,1]\),
we have
\[
\|\bar h_{\lambda_n}\|_{\cH_X}
=
O_P\!\left(
1+\sqrt{\frac{\lambda_n^{-(1+p_X)}}{n}}
\right).
\]
\end{lemma}
\begin{proof}
Recall the decomposition from the proof of \Cref{thm:stage2_oracle_theta}:
\[
\bar h_\lambda-h_\lambda = T_{1,\lambda}+T_{2,\lambda},
\qquad
h_\lambda = (C_F+\lambda I)^{-1}C_F h_\ast,
\]
with
\[
T_{1,\lambda} = (\hat C_F+\lambda I)^{-1}\hat\zeta,
\qquad
T_{2,\lambda} = -(\hat C_F+\lambda I)^{-1}(\hat C_F-C_F)(h_\lambda-h_\ast),
\]
and
\[
\hat\zeta = \frac1n\sum_{i=1}^n \eta_i F_i,
\qquad
F_i = F_\ast(z_i),
\qquad
\eta_i = y_i-\langle h_\ast,F_i\rangle_{H_X}.
\]
Therefore
\(
\|\bar h_\lambda\|_{\cH_X}
\le
\|h_\lambda\|_{\cH_X}
+
\|T_{1,\lambda}\|_{\cH_X}
+
\|T_{2,\lambda}\|_{\cH_X}.
\)
For the deterministic term,
\[
\|h_\lambda\|_{\cH_X}
=
\|(C_F+\lambda I)^{-1}C_F h_\ast\|_{\cH_X}
\le \|h_\ast\|_{\cH_X},
\]
since \(t\mapsto t/(t+\lambda)\) is bounded by \(1\) on \([0,\infty)\).
Next, on the event \(\mathcal E_\lambda\) from the proof of \Cref{thm:stage2_oracle_theta}, we have
\(
(\hat C_F+\lambda I)^{-1} \preceq 3(C_F+\lambda I)^{-1},
\)
and therefore
\[
\|T_{1,\lambda}\|_{\cH_X}
\le
\sqrt 3\,\lambda^{-1/2}
\|(C_F+\lambda I)^{-1/2}\hat\zeta\|_{\cH_X}.
\]
Set
\(
v_i \doteq (C_F+\lambda I)^{-1/2}F_i.
\)
Then
\[
(C_F+\lambda I)^{-1/2}\hat\zeta
=
\frac1n\sum_{i=1}^n \eta_i v_i.
\]
Moreover,
\[
\|v_i\|_{\cH_X}^2
=
\langle F_i,(C_F+\lambda I)^{-1}F_i\rangle_{\cH_X}
\le
\lambda^{-1}\|F_i\|_{\cH_X}^2
\le
\lambda^{-1},
\]
so \(\|v_i\|_{\cH_X}\le \lambda^{-1/2}\). Also,
\[
\mathbb E\|v_i\|_{\cH_X}^2
=
\mathrm{Tr}\!\bigl(C_F(C_F+\lambda I)^{-1}\bigr)
=
\cN_F(\lambda).
\]
Applying \Cref{theo:ope_con_steinwart} to the independent centered Hilbert-valued variables
\(\eta_i v_i\), and using \eqref{asst:mom}, gives an event \(\mathcal B_\lambda\) with
\(P^n(\mathcal B_\lambda^c)\le 2e^{-\tau}\) such that on \(\mathcal B_\lambda\),
\[
\|(C_F+\lambda I)^{-1/2}\hat\zeta\|_{\cH_X}
\le
C
\left(
\sigma\sqrt{\frac{\tau \cN_F(\lambda)}{n}}
+
L\frac{\tau}{n\sqrt\lambda}
\right).
\]
Hence, on \(\mathcal E_\lambda\cap \mathcal B_\lambda\),
\[
\|T_{1,\lambda}\|_{\cH_X}
\le
C
\left(
\sigma\sqrt{\frac{\tau \cN_F(\lambda)}{n\lambda}}
+
L\frac{\tau}{n\lambda}
\right).
\]
For the perturbation term, still on \(\mathcal E_\lambda\),
\[
\|T_{2,\lambda}\|_{\cH_X}
\le
\sqrt 3\,\lambda^{-1/2}
\bigl\|(\hat C_F+\lambda I)^{-1/2}(\hat C_F-C_F)(C_F+\lambda I)^{-1/2}\bigr\|
\,
\|(C_F+\lambda I)^{1/2}(h_\lambda-h_\ast)\|_{\cH_X}.
\]
The operator norm in the middle is bounded by \(2/\sqrt 3\) on \(\mathcal E_\lambda\), exactly as in the
proof of \Cref{thm:stage2_oracle_theta}. Moreover,
\(
h_\lambda-h_\ast = -\lambda(C_F+\lambda I)^{-1}h_\ast,
\)
so
\[
\|(C_F+\lambda I)^{1/2}(h_\lambda-h_\ast)\|_{\cH_X}
=
\lambda\|(C_F+\lambda I)^{-1/2}h_\ast\|_{\cH_X}
\le
\sqrt\lambda\,\|h_\ast\|_{\cH_X}.
\]
Therefore
\(
\|T_{2,\lambda}\|_{\cH_X}
\le
2\|h_\ast\|_{\cH_X}.
\)
Combining the bounds for \(h_\lambda\), \(T_{1,\lambda}\), and \(T_{2,\lambda}\), we obtain on
\(\mathcal E_\lambda\cap \mathcal B_\lambda\),
\[
\|\bar h_\lambda\|_{\cH_X}
\le
C
\left(
1
+
\sigma\sqrt{\frac{\tau \cN_F(\lambda)}{n\lambda}}
+
L\frac{\tau}{n\lambda}
\right).
\]
By \Cref{lma:con_cf}, \(P^n(\mathcal E_\lambda^c)\le 2e^{-\tau}\) whenever \(n\ge 8\tau g_\lambda \lambda^{-1}\),
hence
\(
P^n(\mathcal E_\lambda\cap \mathcal B_\lambda)\ge 1-4e^{-\tau}.
\)
For the consequence, Jensen's inequality gives \(C_F \preceq C_X\), and thus
\[
\cN_F(\lambda)
=
\mathrm{Tr}\!\bigl(C_F(C_F+\lambda I)^{-1}\bigr)
\le
\mathrm{Tr}\!\bigl(C_X(C_X+\lambda I)^{-1}\bigr)
\le
C\lambda^{-p_X},
\]
where the last inequality follows from \eqref{asst:evd} (Lemma 11~\citet{fischer2020sobolev}). Hence
\[
\|\bar h_\lambda\|_{\cH_X}
\le
C
\left(
1
+
\sigma\sqrt{\frac{\tau}{n}}\lambda^{-(1+p_X)/2}
+
L\frac{\tau}{n}\lambda^{-1}
\right)
\]
with high probability. For $\lambda_n=n^{-\ell}$ with $\ell<1$, we have
$$
n^{-1}\lambda_n^{-1}
\le
n^{-1/2}\lambda_n^{-(1+p_X)/2},
\iff
\lambda_n^{1-p_X}\ge n^{-1},
$$
If $\lambda_n=n^{-\ell}$ with $\ell\in(0,1]$, then
$\lambda_n^{1-p_X}=n^{-\ell(1-p_X)}\ge n^{-1}$,
since $\ell(1-p_X)\le 1$.
\end{proof}
\subsubsection{Stage 1 Error: Detailed Results} \label{sec:s1_app}

\paragraph{Conditional convention for Stage 1.}
Throughout \Cref{sec:s1_app} we condition on the first-stage sample $\cD_1$ used to construct
$\hat F_\xi$. Accordingly, all probabilities in the stage-1 perturbation bounds are with
respect to the second-stage sample only, i.e. they are understood as
$P^n(\cdot \mid \cD_1)$-probabilities.

\medskip \noindent
The following theorem provides an upper bound on Eq.~\eqref{eq:s1a0}, term $S_{-1}$.

\begin{theorem}\label{theo:s1e1}
Let Assumptions~\ref{asst:rkhs}, \eqref{asst:mom}, \eqref{asst:srcz} and \eqref{asst:embz}
hold with $\alpha_Z<\beta_Z$. Condition on the first-stage sample $\cD_1$ used to construct
$\hat F_\xi$. For $\tau\ge 1$, and sufficiently large $m$ and $n$ such that \Cref{eq:m_con}
holds, we have with $P^n(\cdot\mid \cD_1)$-probability at least $1-8e^{-\tau}$,
\[
S_{-1}
\le
c\,
\frac{\tau}{\sqrt\lambda}
\left(
\frac{\|F_*-\hat F_\xi\|_{\alpha_Z}}{\sqrt n}
+
\|F_*-\hat F_\xi\|_{L_2(Z;\cH_X)}
\right),
\]
where $c$ depends on $\sigma,L,A_Z,B_Z,\alpha_Z,\beta_Z$ and $\|h_*\|_{\cH_X}$.
\end{theorem}

\begin{proof}
Set $\Delta F \doteq F_*-\hat F_\xi.$ Starting from \Cref{eq:s1a0},
\begin{align*}
S_{-1}
&=
\left\|
(C_F+\lambda I)^{1/2}
(\hat C_{\hat F}+\lambda I)^{-1}
\frac1n(\Phi_{\hat F}^*-\Phi_{F_*}^*)Y
\right\|_{\cH_X} \\
&\le
\left\|
(C_F+\lambda I)^{1/2}
(\hat C_{\hat F}+\lambda I)^{-1/2}
\right\|_{\cH_X\to\cH_X}
\left\|
(\hat C_{\hat F}+\lambda I)^{-1/2}
(\hat C_F+\lambda I)^{1/2}
\right\|_{\cH_X\to\cH_X}
\\
&\qquad\times
\left\|
(\hat C_F+\lambda I)^{-1/2}
\frac1n(\Phi_{\hat F}^*-\Phi_{F_*}^*)Y
\right\|_{\cH_X}.
\end{align*}
Let $\mathcal E_6$ be the event from \Cref{lma:emp_sub_app} and $\mathcal E_7$ be the event from \Cref{lma:chatc_2} that are such that $\mathcal E_7 \subseteq \mathcal E_6$ and
\(
P^n(\mathcal E_{7}^c\mid \cD_1)\le 6e^{-\tau}.
\)
On $\mathcal E_{7}$, \Cref{lma:chatc} yields
\(
\left\|
(C_F+\lambda I)^{1/2}
(\hat C_{\hat F}+\lambda I)^{-1/2}
\right\|_{\cH_X\to\cH_X}
\le 3,
\)
and on $\mathcal E_{6}$, Lemma~\ref{lma:chatc3} gives
\(
\left\|
(\hat C_{\hat F}+\lambda I)^{-1/2}
(\hat C_F+\lambda I)^{1/2}
\right\|_{\cH_X\to\cH_X}
\le \sqrt{\frac65}.
\)
Hence, on $\mathcal E_{7}$,
\[
\left\|
(C_F+\lambda I)^{1/2}
(\hat C_{\hat F}+\lambda I)^{-1/2}
\right\|_{\cH_X\to\cH_X}
\left\|
(\hat C_{\hat F}+\lambda I)^{-1/2}
(\hat C_F+\lambda I)^{1/2}
\right\|_{\cH_X\to\cH_X}
\le 4.
\]
We now bound the remaining factor. Write
\begin{align*}
&\left\|
(\hat C_F+\lambda I)^{-1/2}
\frac1n(\Phi_{\hat F}^*-\Phi_{F_*}^*)Y
\right\|_{\cH_X} \\
&\le
\underbrace{
\left\|
(\hat C_F+\lambda I)^{-1/2}
\frac1n(\Phi_{\hat F}^*-\Phi_{F_*}^*)
\bigl(Y-\Phi_{F_*}h_*\bigr)
\right\|_{\cH_X}
}_{=:I}
+
\underbrace{
\left\|
(\hat C_F+\lambda I)^{-1/2}
\frac1n(\Phi_{\hat F}^*-\Phi_{F_*}^*)\Phi_{F_*}h_*
\right\|_{\cH_X}
}_{=:II}.
\end{align*}
For $I$, define
\(
\theta_i \doteq ( \hat F_\xi(z_i)-F_*(z_i) )\,
\bigl(y_i-\langle h_*,F_*(z_i)\rangle_{\cH_X}\bigr), i=1,\dots,n.
\)
Then
\[
I
\le
\lambda^{-1/2}
\left\|
\frac1n\sum_{i=1}^n \theta_i
\right\|_{\cH_X}.
\]
Conditionally on $\cD_1$, the random variables $\theta_i$ are i.i.d. and centered, because
\[
\E\!\left[
y_i-\langle h_*,F_*(z_i)\rangle_{\cH_X}
\mid z_i, \cD_1
\right]
=
0.
\]
Moreover, by \Cref{lma:rhks_as} and \eqref{asst:mom}, for every integer $q\ge 2$,
\begin{align*}
\E\!\left[\|\theta_i\|_{\cH_X}^q \mid \cD_1\right]
&\le
\| \hat F_\xi-F_*\|_{L_\infty(Z;\cH_X)}^q\,
\E\!\left[
\bigl|y_i-\langle h_*,F_*(z_i)\rangle_{\cH_X}\bigr|^q
\mid \cD_1
\right] \\
&\le
\frac{q!}{2}
\bigl(\sigma A_Z\|\Delta F\|_{\alpha_Z}\bigr)^2
\bigl(LA_Z\|\Delta F\|_{\alpha_Z}\bigr)^{q-2}.
\end{align*}
Applying \Cref{theo:ope_con_steinwart} conditionally on $\cD_1$, there exists an event $\mathcal E_I$ such that
\(
P^n(\mathcal E_I^c\mid \cD_1)\le 2e^{-\tau},
\)
and on $\mathcal E_I$,
\[
\left\|
\frac1n\sum_{i=1}^n \theta_i
\right\|_{\cH_X}
\le
\sqrt{\frac{2\tau}{n}}\sigma A_Z\|\Delta F\|_{\alpha_Z}
+
\frac{2\tau}{n}\frac{L A_Z\|\Delta F\|_{\alpha_Z}}{\sqrt n}.
\]
Hence, on $\mathcal E_I$, using $\tau \geq 1$,
\(
I
\le
C\,
\frac{\tau}{\sqrt\lambda}
\frac{\|\Delta F\|_{\alpha_Z}}{\sqrt n},
\)
for a constant $C$ depending only on $\sigma,L,A_Z$.
For $II$, on $\mathcal E_{6}$, the auxiliary bound at the end of \Cref{lma:emp_sub_app} gives
\[
\left\|
\frac1n(\Phi_{\hat F}-\Phi_{F_*})^*\Phi_{F_*}
\right\|_{\cH_X\to\cH_X}
\le
A_ZB_Z
\left(
A_Z\|\Delta F\|_{\alpha_Z}\sqrt{\frac{\tau}{n}}
+
\|\Delta F\|_{L_2(Z;\cH_X)}
\right).
\]
Therefore, on $\mathcal E_{6}$,
\begin{align*}
II
&\le
\lambda^{-1/2}
\left\|
\frac1n(\Phi_{\hat F}-\Phi_{F_*})^*\Phi_{F_*}
\right\|_{\cH_X\to\cH_X}
\|h_*\|_{\cH_X} \\
&\le
\frac{A_ZB_Z\|h_*\|_{\cH_X}}{\sqrt\lambda}
\left(
A_Z\|\Delta F\|_{\alpha_Z}\sqrt{\frac{\tau}{n}}
+
\|\Delta F\|_{L_2(Z;\cH_X)}
\right) \\
&\le
C\,
\frac{\tau}{\sqrt\lambda}
\left(
\frac{\|\Delta F\|_{\alpha_Z}}{\sqrt n}
+
\|\Delta F\|_{L_2(Z;\cH_X)}
\right),
\end{align*}
where in the last step we used $\tau\ge 1$. Combining the preceding bounds, on $\mathcal E_{7}\cap \mathcal E_I$,
\[
S_{-1}
\le
4(I+II)
\le
c\,
\frac{\tau}{\sqrt\lambda}
\left(
\frac{\|\Delta F\|_{\alpha_Z}}{\sqrt n}
+
\|\Delta F\|_{L_2(Z;\cH_X)}
\right),
\]
with $c$ depending only on $\sigma,L,A_Z,B_Z,\alpha_Z,\beta_Z$ and
$\|h_*\|_{\cH_X}$. Since
\(
P^n((\mathcal E_{7}\cap \mathcal E_I)^c\mid \cD_1)\le 8e^{-\tau},
\)
the proof is complete.
\end{proof}
The following theorem provides an upper bound on Eq.~\eqref{eq:s1b0}, term $S_0$.

\begin{theorem}\label{theo:s1e2}
Let Assumptions~\ref{asst:rkhs}, \eqref{asst:srcz} and \eqref{asst:embz} hold with
$\alpha_Z<\beta_Z$. Condition on the first-stage sample $\cD_1$ used to construct
$\hat F_\xi$. For $\tau\ge 1$, and sufficiently large $m$ and $n$ such that \eqref{eq:m_con}
holds, we have with $P^n(\cdot\mid \cD_1)$-probability at least $1-6e^{-\tau}$,
\[
S_0
\le
c'\,
\frac{\tau}{\sqrt\lambda}
\left(
\frac{\|F_*-\hat F_\xi\|_{\alpha_Z}}{\sqrt n}
+
\|F_*-\hat F_\xi\|_{L_2(Z;\cH_X)}
\right)
\|\bar h_\lambda\|_{\cH_X},
\]
where $c'$ depends on $A_Z,B_Z,\alpha_Z,\beta_Z$.
\end{theorem}

\begin{proof}
Starting from \Cref{eq:s1b0} and using the resolvent identity
\(
A^{-1}-B^{-1}=A^{-1}(B-A)B^{-1},
\)
with
\(
A=\hat C_{\hat F}+\lambda I,
B=\hat C_F+\lambda I,
\)
we get
\begin{align*}
S_0
&=
\left\|
(C_F+\lambda I)^{1/2}
\Big(
(\hat C_{\hat F}+\lambda I)^{-1}
-
(\hat C_F+\lambda I)^{-1}
\Big)
\frac1n\Phi_{F_*}^*Y
\right\|_{\cH_X} \\
&=
\left\|
(C_F+\lambda I)^{1/2}
(\hat C_{\hat F}+\lambda I)^{-1}
(\hat C_F-\hat C_{\hat F})
(\hat C_F+\lambda I)^{-1}
\frac1n\Phi_{F_*}^*Y
\right\|_{\cH_X} \\
&\le
\lambda^{-1/2}
\left\|
(C_F+\lambda I)^{1/2}
(\hat C_{\hat F}+\lambda I)^{-1/2}
\right\|_{\cH_X\to\cH_X}
\,
\|\hat C_F-\hat C_{\hat F}\|_{\cH_X\to\cH_X}
\,
\|\bar h_\lambda\|_{\cH_X}.
\end{align*}
Let $\mathcal E_6$ be the event from \Cref{lma:emp_sub_app} and $\mathcal E_7$ be the event from \Cref{lma:chatc_2} that are such that $\mathcal E_7 \subseteq \mathcal E_6$ and
\(
P^n(\mathcal E_{7}^c\mid \cD_1)\le 6e^{-\tau}.
\)
On $\mathcal E_{7}$, \Cref{lma:chatc} yields
\[
\left\|
(C_F+\lambda I)^{1/2}
(\hat C_{\hat F}+\lambda I)^{-1/2}
\right\|_{\cH_X\to\cH_X}
\le 3,
\]
and on $\mathcal E_6$, \Cref{lma:emp_sub_app} gives
\[
\|\hat C_F-\hat C_{\hat F}\|_{\cH_X\to\cH_X}
\le
J\left(
\|F_*-\hat F_\xi\|_{L_2(Z;\cH_X)}
+
\|F_*-\hat F_\xi\|_{\alpha_Z}\sqrt{\frac{\tau}{n}}
\right).
\]
Therefore, on $\mathcal E_{7}$,
\[
S_0
\le
3J\,
\frac{1}{\sqrt\lambda}
\left(
\|F_*-\hat F_\xi\|_{L_2(Z;\cH_X)}
+
\|F_*-\hat F_\xi\|_{\alpha_Z}\sqrt{\frac{\tau}{n}}
\right)
\|\bar h_\lambda\|_{\cH_X}.
\]
Since $\tau\ge 1$, we may absorb $\sqrt{\tau}$ into $\tau$ and obtain the stated bound.
\end{proof}
\subsection{Proof of Corollary~\ref{cor:iv_rate_main}} \label{sec:proof_cor}
Corollary~\ref{cor:iv_rate_main} is a special case of the following more general result.
\begin{corollary}[Full sample-allocation regimes]
Assume the conditions of \Cref{theo:iv_l2} and Assumption~\eqref{asst:evdz}. Let
\(
 m=n^a, 
 \xi_m=\Theta\!\left(m^{-1/(\beta_Z+p_Z)}\right),
\)
for some $a>0$, and define
\(
 c_F \doteq \mathbf 1_{N(C_F)=\{0\}}, D \doteq \beta_X+\gamma+p_X-1, \Delta \doteq \beta_X+2\gamma-c_F.
\)
\[
 \overline\Delta \doteq \Delta+\gamma(1+p_X)
 \;=
 \beta_X+3\gamma+\gamma p_X-c_F,
\]
\[
 a_0 \doteq \frac{\beta_Z+p_Z}{\alpha_Z},
 \qquad
 a_{A,0} \doteq \frac{\beta_Z+p_Z}{\beta_Z}\,\frac{\Delta}{\gamma(1+p_X)},
 \qquad
 a_{B,0} \doteq \frac{\beta_Z+p_Z}{\beta_Z-\alpha_Z}\!\left(\frac{\Delta}{\gamma(1+p_X)}-1\right),
\]
\[
 \widetilde a_A \doteq \frac{\beta_Z+p_Z}{\beta_Z}
 \frac{\Delta+\bigl(1-\beta_X+(\gamma-1)p_X\bigr)_+}{D},
\qquad
 \widetilde a_B \doteq \frac{\beta_Z+p_Z}{\beta_Z-\alpha_Z}
 \frac{\Delta-D+\bigl(1-\beta_X+(\gamma-1)p_X\bigr)_+}{D}.
\]
With the convention that an interval of the form $[u,v)$ is empty when $u\ge v$, the following regimes hold.
\medskip
\noindent\emph{Case A:} $\widetilde a_A\le a_0$.
\begin{enumerate}
\item If $a\ge \widetilde a_A$, then taking
\(
 \lambda_n
 =
 \Theta\!\left(n^{-\gamma/D}\right)
\)
yields
\(
 \left\lVert \hat h_{\lambda_n,\xi_m}-h^*\right\rVert_{L_2(X)}^2
 =
 O_P\!\left(n^{-\beta_X/D}\right).
\)
\item If $a_{A,0}\le a<\widetilde a_A$, then taking
\(
 \lambda_n
 =
 \Theta\!\left(
 n^{-\frac{\gamma}{\overline\Delta}
 \left(1+a\frac{\beta_Z}{\beta_Z+p_Z}\right)}
 \right)
\)
yields
\[
 \left\lVert \hat h_{\lambda_n,\xi_m}-h^*\right\rVert_{L_2(X)}^2
 =
 O_P\!\left(
 n^{-\frac{\beta_X}{\overline\Delta}
 \left(1+a\frac{\beta_Z}{\beta_Z+p_Z}\right)}
 \right).
\]
\item If $a<\min\{a_{A,0},\widetilde a_A\}$, then taking
\(
 \lambda_n
 =
 \Theta\!\left(
 n^{-a\frac{\beta_Z}{\beta_Z+p_Z}\frac{\gamma}{\Delta}}
 \right)
\)
yields
\[
 \left\lVert \hat h_{\lambda_n,\xi_m}-h^*\right\rVert_{L_2(X)}^2
 =
 O_P\!\left(
 n^{-a\frac{\beta_Z}{\beta_Z+p_Z}\frac{\beta_X}{\Delta}}
 \right).
\]
\end{enumerate}
\medskip
\noindent\emph{Case B:} $\widetilde a_A>a_0$.
\begin{enumerate}
\item If $a\ge \widetilde a_B$, then taking
\(
 \lambda_n
 =
 \Theta\!\left(n^{-\gamma/D}\right)
\)
yields
\(
 \left\lVert \hat h_{\lambda_n,\xi_m}-h^*\right\rVert_{L_2(X)}^2
 =
 O_P\!\left(n^{-\beta_X/D}\right).
\)
\item If $\max\{a_0,a_{B,0}\}\le a<\widetilde a_B$, then taking
\(
 \lambda_n
 =
 \Theta\!\left(
 n^{-\frac{\gamma}{\overline\Delta}
 \left(2+a\frac{\beta_Z-\alpha_Z}{\beta_Z+p_Z}\right)}
 \right)
\)
yields
\[
 \left\lVert \hat h_{\lambda_n,\xi_m}-h^*\right\rVert_{L_2(X)}^2
 =
 O_P\!\left(
 n^{-\frac{\beta_X}{\overline\Delta}
 \left(2+a\frac{\beta_Z-\alpha_Z}{\beta_Z+p_Z}\right)}
 \right).
\]
\item If $a_0\le a<\min\{a_{B,0},\widetilde a_B\}$, then taking
\(
 \lambda_n
 =
 \Theta\!\left(
 n^{-\frac{\gamma}{\Delta}
 \frac{a(\beta_Z-\alpha_Z)+\beta_Z+p_Z}{\beta_Z+p_Z}}
 \right)
\)
yields
\[
 \left\lVert \hat h_{\lambda_n,\xi_m}-h^*\right\rVert_{L_2(X)}^2
 =
 O_P\!\left(
 n^{-\frac{\beta_X}{\Delta}
 \frac{a(\beta_Z-\alpha_Z)+\beta_Z+p_Z}{\beta_Z+p_Z}}
 \right).
\]
\item If $a_{A,0}\le a<a_0$, then taking
\(
 \lambda_n
 =
 \Theta\!\left(
 n^{-\frac{\gamma}{\overline\Delta}
 \left(1+a\frac{\beta_Z}{\beta_Z+p_Z}\right)}
 \right)
\)
yields
\[
 \left\lVert \hat h_{\lambda_n,\xi_m}-h^*\right\rVert_{L_2(X)}^2
 =
 O_P\!\left(
 n^{-\frac{\beta_X}{\overline\Delta}
 \left(1+a\frac{\beta_Z}{\beta_Z+p_Z}\right)}
 \right).
\]
\item If $a<\min\{a_{A,0},a_0\}$, then taking
\(
 \lambda_n
 =
 \Theta\!\left(
 n^{-a\frac{\beta_Z}{\beta_Z+p_Z}\frac{\gamma}{\Delta}}
 \right)
\)
yields
\[
 \left\lVert \hat h_{\lambda_n,\xi_m}-h^*\right\rVert_{L_2(X)}^2
 =
 O_P\!\left(
 n^{-a\frac{\beta_Z}{\beta_Z+p_Z}\frac{\beta_X}{\Delta}}
 \right).
\]
\end{enumerate}
\end{corollary}
\begin{proof}
Let
\(
 r_1(t,m)\doteq m^{-\frac{\beta_Z-t}{2(\beta_Z+p_Z)}},
 r_2(n,\lambda)
 \doteq
 \lambda^{\beta_X/(2\gamma)}
 +
 \frac{1}{\sqrt n}\lambda^{-(\gamma+p_X-1)/(2\gamma)}
 +
 \frac{1}{n}\lambda^{-1+1/(2\gamma)}.
\)
Set $\lambda=n^{-\ell}$ for some $\ell\in(0,1)$ to be selected later. By the pointwise bound on $\|\bar h_\lambda\|_{\mathcal H_X}$ in \Cref{lem:rkhs_bound_s2} and the proof of \Cref{theo:iv_l2}, there is an event of probability at least $1-Ce^{-\tau}$ such that, for every fixed $\tau\ge 1$ and all sufficiently large $m,n$,
\begin{equation}\label{eq:new-coro2-start}
\left\lVert \hat h_{\lambda,\xi_m}-h^*\right\rVert_{L_2(X)}
\lesssim
\lambda^{\frac{c_F}{2\gamma}-1}
\left(r_1(0,m)+\frac{r_1(\alpha_Z,m)}{\sqrt n}\right)
\left(1+\frac{1}{\sqrt n}\lambda^{-(1+p_X)/2}\right)
+
 r_2(n,\lambda).
\end{equation}
Set
\(
 u_A(a)\doteq\frac{a\beta_Z}{2(\beta_Z+p_Z)},
 u_B(a)\doteq\frac{a(\beta_Z-\alpha_Z)+\beta_Z+p_Z}{2(\beta_Z+p_Z)},
\)
\[
 \kappa\doteq\frac{2\gamma-c_F}{2\gamma},
 \quad
 \eta(\ell)\doteq\left(\frac{(1+p_X)\ell-1}{2}\right)_+, \quad
 \phi_1(\ell)\doteq\frac{\beta_X}{2\gamma}\ell,
 \quad
 \phi_2(\ell)\doteq\frac12-\frac{\gamma+p_X-1}{2\gamma}\ell,
\]
\[
 \psi_A(\ell,a)\doteq u_A(a)-\kappa\ell-\eta(\ell),
 \qquad
 \psi_B(\ell,a)\doteq u_B(a)-\kappa\ell-\eta(\ell).
\]
Eq.~\eqref{eq:new-coro2-start} implies
\begin{equation}\label{eq:new-coro2-rate-function}
\left\lVert \hat h_{\lambda,\xi_m}-h^*\right\rVert_{L_2(X)}
\lesssim
 n^{-\psi_A(\ell,a)}
+
 n^{-\psi_B(\ell,a)}
+
 n^{-\phi_1(\ell)}
+
 n^{-\phi_2(\ell)}.
\end{equation}
Moreover,
\(
 u_A(a)\le u_B(a)
 \iff
 a\le a_0,
\)
so the slower stage-1 term is $n^{-\psi_A(\ell,a)}$ when $a<a_0$, and $n^{-\psi_B(\ell,a)}$ when $a\ge a_0$.

\medskip
\noindent\textbf{Step 1: the stage-2 optimal tuning.}
Let
\(
 \ell_*\doteq \gamma/D.
\)
Then $\phi_1(\ell_*)=\phi_2(\ell_*)=\beta_X/(2D)$, and since $D=\beta_X+\gamma+p_X-1>\gamma$, we have $0<\ell_*<1$. Furthermore,
\[
 \eta(\ell_*)
 =
 \left(\frac{\gamma(1+p_X)-D}{2D}\right)_+
 =
 \frac{\bigl(1-\beta_X+(\gamma-1)p_X\bigr)_+}{2D}.
\]
Therefore
\(
 \psi_A(\ell_*,a)\ge \phi_1(\ell_*)
 \iff
 a\ge \widetilde a_A,
 \psi_B(\ell_*,a)\ge \phi_1(\ell_*)
 \iff
 a\ge \widetilde a_B.
\)
Also,
\(
 \widetilde a_A\le a_0
 \iff
 \widetilde a_B\le a_0,
\)
because both inequalities are equivalent to
\[
 \alpha_Z\Bigl(\Delta+\bigl(1-\beta_X+(\gamma-1)p_X\bigr)_+\Bigr)
 \le
 \beta_Z D.
\]
Hence, if $\widetilde a_A\le a_0$, every $a\ge\widetilde a_A$ is already in the stage-2-optimal regime; if $\widetilde a_A>a_0$, the same is true for every $a\ge\widetilde a_B$. In both cases,
\[
 \lambda_n=\Theta\!\left(n^{-\gamma/D}\right)
 \qquad\Longrightarrow\qquad
 \left\lVert \hat h_{\lambda_n,\xi_m}-h^*\right\rVert_{L_2(X)}^2
 =
 O_P\!\left(n^{-\beta_X/D}\right).
\]

\medskip
\noindent\textbf{Step 2: the breakpoint $\ell_0=(1+p_X)^{-1}$.}
Write
\(
 \ell_0\doteq 1/(1+p_X).
\)
For $\ell\le \ell_0$ one has $\eta(\ell)=0$, whereas for $\ell\ge \ell_0$ one has
\(
 \eta(\ell)=((1+p_X)\ell-1)/2.
\)
Since $\phi_1$ is increasing and the relevant stage-1 exponent is decreasing, the maximizer of the minimum in Eq.~\eqref{eq:new-coro2-rate-function} is obtained by balancing $\phi_1$ with the relevant stage-1 exponent, either below or above $\ell_0$.

\smallskip
\noindent\emph{A-branch ($a<a_0$).}
For $\ell\le \ell_0$, solving $\phi_1(\ell)=\psi_A(\ell,a)$ gives
\(
 \ell_{A,-}
 =
 \frac{\gamma}{\Delta}\,\frac{a\beta_Z}{\beta_Z+p_Z},
\)
which satisfies $\ell_{A,-}\le \ell_0$ exactly when $a\le a_{A,0}$. For $\ell\ge \ell_0$, solving $\phi_1(\ell)=\psi_A(\ell,a)$ gives
\[
 \ell_{A,+}
 =
 \frac{\gamma}{\overline\Delta}
 \left(1+a\frac{\beta_Z}{\beta_Z+p_Z}\right),
\]
which satisfies $\ell_{A,+}\ge \ell_0$ exactly when $a\ge a_{A,0}$. Moreover,
\[
 \phi_1(\ell_{A,-})
 =
 \frac{\beta_X}{2\Delta}
 a\frac{\beta_Z}{\beta_Z+p_Z},
 \qquad
 \phi_1(\ell_{A,+})
 =
 \frac{\beta_X}{2\overline\Delta}
 \left(1+a\frac{\beta_Z}{\beta_Z+p_Z}\right).
\]

\smallskip
\noindent\emph{B-branch ($a\ge a_0$).}
For $\ell\le \ell_0$, solving $\phi_1(\ell)=\psi_B(\ell,a)$ gives
\[
 \ell_{B,-}
 =
 \frac{\gamma}{\Delta}
 \frac{a(\beta_Z-\alpha_Z)+\beta_Z+p_Z}{\beta_Z+p_Z},
\]
which satisfies $\ell_{B,-}\le \ell_0$ exactly when $a\le a_{B,0}$. For $\ell\ge \ell_0$, solving $\phi_1(\ell)=\psi_B(\ell,a)$ gives
\[
 \ell_{B,+}
 =
 \frac{\gamma}{\overline\Delta}
 \left(2+a\frac{\beta_Z-\alpha_Z}{\beta_Z+p_Z}\right),
\]
which satisfies $\ell_{B,+}\ge \ell_0$ exactly when $a\ge a_{B,0}$. Moreover,
\[
 \phi_1(\ell_{B,-})
 =
 \frac{\beta_X}{2\Delta}
 \frac{a(\beta_Z-\alpha_Z)+\beta_Z+p_Z}{\beta_Z+p_Z}, \qquad
 \phi_1(\ell_{B,+})
 =
 \frac{\beta_X}{2\overline\Delta}
 \left(2+a\frac{\beta_Z-\alpha_Z}{\beta_Z+p_Z}\right).
\]

\medskip
\noindent\textbf{Step 3: assemble the regimes.}
If $\widetilde a_A\le a_0$, then every $a\ge\widetilde a_A$ is already optimal. For $a<\widetilde a_A$, necessarily $a<a_0$, so only the A-branch is relevant. Thus:
\begin{itemize}
\item if $a_{A,0}\le a<\widetilde a_A$, use $\ell_{A,+}$ and obtain item~2 of Case~A;
\item if $a<\min\{a_{A,0},\widetilde a_A\}$, use $\ell_{A,-}$ and obtain item~3 of Case~A.
\end{itemize}

If $\widetilde a_A>a_0$, then the optimal branch starts at $a\ge\widetilde a_B$. For $a_0\le a<\widetilde a_B$, the B-branch is relevant, hence:
\begin{itemize}
\item if $\max\{a_0,a_{B,0}\}\le a<\widetilde a_B$, use $\ell_{B,+}$ and obtain item~2 of Case~B;
\item if $a_0\le a<\min\{a_{B,0},\widetilde a_B\}$, use $\ell_{B,-}$ and obtain item~3 of Case~B.
\end{itemize}
For $a<a_0$, the A-branch is relevant, hence:
\begin{itemize}
\item if $a_{A,0}\le a<a_0$, use $\ell_{A,+}$ and obtain item~4 of Case~B;
\item if $a<\min\{a_{A,0},a_0\}$, use $\ell_{A,-}$ and obtain item~5 of Case~B.
\end{itemize}
In each subcase, the corresponding intersection lies below $\ell_*$ because the relevant allocation interval is strictly below the optimal threshold ($\widetilde a_A$ or $\widetilde a_B$). Hence $\phi_2$ is not active.

\medskip
\noindent\textbf{Step 4: verification of the size conditions in \eqref{eq:m_con}.}
All the tunings above satisfy $0<\ell\le \ell_*<1$. Hence $g_{\lambda_n}/(n\lambda_n)\to 0$, so the first constraint in \eqref{eq:m_con} holds eventually. Since
\[
 r_1(0,m)=n^{-u_A(a)},
 \qquad
 \frac{r_1(\alpha_Z,m)}{\sqrt n}=n^{-u_B(a)},
\]
it is enough to check that the selected exponent $\ell$ satisfies $\ell\le u_A(a)$ in the A-branches and $\ell\le u_B(a)$ in the B-branches.
For the unpenalized branches,
\[
 \ell_{A,-}=\frac{2\gamma}{\Delta}u_A(a)\le u_A(a),
 \qquad
 \ell_{B,-}=\frac{2\gamma}{\Delta}u_B(a)\le u_B(a),
\]
because $\Delta\ge 2\gamma$.
For the penalized A-branch,
\(
 \ell_{A,+}=\frac{\gamma}{\overline\Delta}\bigl(1+2u_A(a)\bigr),
\)
and at $a=a_{A,0}$ one has
\[
 \ell_{A,+}=\frac{1}{1+p_X},
 \qquad
 u_A(a_{A,0})=\frac{\Delta}{2\gamma(1+p_X)}\ge \frac{1}{1+p_X}.
\]
Since the slope of $u_A(a)$ is $\beta_Z/[2(\beta_Z+p_Z)]$ and the slope of $\ell_{A,+}$ is $\gamma\beta_Z/[\overline\Delta(\beta_Z+p_Z)]\le \beta_Z/[2(\beta_Z+p_Z)]$, it follows that $\ell_{A,+}\le u_A(a)$ for all $a\ge a_{A,0}$.
The penalized B-branch is identical:
\(
 \ell_{B,+}=\frac{\gamma}{\overline\Delta}\bigl(1+2u_B(a)\bigr),
\)
at $a=a_{B,0}$ one has
\(
 \ell_{B,+}=\frac{1}{1+p_X}, 
 u_B(a_{B,0})=\frac{\Delta}{2\gamma(1+p_X)}\ge \frac{1}{1+p_X},
\)
and the slope of $u_B(a)$ is $(\beta_Z-\alpha_Z)/[2(\beta_Z+p_Z)]$ whereas the slope of $\ell_{B,+}$ is $\gamma(\beta_Z-\alpha_Z)/[\overline\Delta(\beta_Z+p_Z)]\le (\beta_Z-\alpha_Z)/[2(\beta_Z+p_Z)]$. Hence $\ell_{B,+}\le u_B(a)$ for all $a\ge a_{B,0}$.
Therefore all constraints in \eqref{eq:m_con} hold eventually.
\end{proof}

\section{Proof of Theorem~\ref{theo:lower_bound}} \label{sec:proof_lb}

We prove the lower bound directly on a hard NPIV subclass built from a single admissible
channel. This avoids the auxiliary NPIR reduction as done by \cite{chen2011rate}. Let $(\cH_X,\pi_X)$ be a RKHS satisfying the assumptions given in Theorem~\ref{theo:lower_bound}. Fix parameters $\beta_X > 0, \gamma \geq 1$, and constants $B_X, \sigma^2, L > 0$.
By Assumption~\eqref{asst:link+}, fix a joint law $P^\dagger_{X,Z}$ with $X$-marginal $\pi_X$ and
associated conditional mean embedding
\[
F^\dagger(Z)\doteq E[\phi_X(X)\mid Z],
\qquad
C_F^\dagger \doteq E[F^\dagger(Z)\otimes F^\dagger(Z)],
\]
such that
\(
R_0 C_X^\gamma \preceq C_F^\dagger \preceq R_1 C_X^\gamma .
\)
Let $\widetilde T^\dagger:\cH_X\to L_2(Z)$ denote the corresponding conditional expectation
operator, so that
\(
(\widetilde T^\dagger h)(Z)=\langle h,F^\dagger(Z)\rangle_{\cH_X}.
\)
The two-sided comparison implies
\(
\cN(C_F^\dagger)= \cN(C_X).
\)
Indeed, if $f\in \cN(C_X)$ then
\(
\langle f,C_F^\dagger f\rangle
\le
R_1\langle f,C_X^\gamma f\rangle
=0,
\)
so $f\in \cN(C_F^\dagger)$. Conversely, if $f\in \cN(C_F^\dagger)$, then
\(
R_0\langle f,C_X^\gamma f\rangle
\le
\langle f,C_F^\dagger f\rangle
=0,
\)
hence $f\in \cN(C_X)$. Let $(\mu_{X,i},e_{X,i})_{i\ge 1}$ be the spectral decomposition from \Cref{eq:SVD}. For
$\ell \in\mathbb N$, $\varepsilon\in(0,1]$, and $\omega=(\omega_1,\ldots,\omega_\ell)\in\{0,1\}^\ell$,
define
\[
h_\omega
\doteq
2\Bigl(\frac{8\varepsilon}{\ell}\Bigr)^{1/2}
\sum_{i=1}^\ell \omega_i e_{X,i+\ell}.
\]
Choose $s_0>0$ sufficiently small, depending only on $(\sigma,L)$, so that a centered
Gaussian  $\cN(0,s_0^2)$ satisfies \eqref{asst:mom} with parameters $(\sigma,L)$.
Let $\xi\sim \cN(0,s_0^2)$ be independent of $(X,Z)\sim P^\dagger_{X,Z}$.
For each $\omega$, define
\(
Y_\omega \doteq \langle h_\omega,F^\dagger(Z)\rangle_{\cH_X}+\xi.
\)
Then, with
\[
U_\omega \doteq \langle h_\omega,F^\dagger(Z)\rangle_{\cH_X}-h_\omega(X)+\xi, \text{ we have } \quad Y_\omega = h_\omega(X)+U_\omega,
\quad
E[U_\omega\mid Z]=0.
\]
Thus each $\omega$ defines a genuine NPIV model $P_\omega$ with fixed channel
$P^\dagger_{X,Z}$ and structural target $h_\omega$. Since $h_\omega\in \cN(C_X)^\perp=\cN(C_F^\dagger)^\perp$, it is the unique minimum-$\cH_X$-norm
solution of the inverse equation
\(
\widetilde T^\dagger h_{\omega} = E[Y_\omega\mid Z]
\)
under the fixed channel $P^\dagger_{X,Z}$. Moreover, the first-stage sample
\(
\cD_1=((\tilde Z_i,\tilde X_i))_{i=1}^m \sim (P^\dagger_{Z,X})^{\otimes m}
\)
has the same law under all $\omega$. Therefore $\cD_1$ carries no information about the
index $\omega$, and all distinguishability comes from the second-stage sample $\cD_2$.

\medskip \noindent
We recall the definition of the Kullback-Leibler divergence. For two probability measures $P, P'$ on some measurable space $(\Omega, \mathcal{A})$ the Kullback-Leibler divergence is given by
$$
\mathrm{KL}\left(P, P'\right):=\int_{\Omega} \log \left(\frac{\mathrm{d} P}{\mathrm{~d} P'}\right) \mathrm{d} P, \qquad \text{ if $P \ll P'$ and otherwise $\mathrm{KL}\left(P, P'\right):=\infty$. }
$$
We distinguish the following steps to obtain the lower bound.
\begin{itemize}
    \item Step 1: Control the separation in $L_2(X)$ norm between the different $h_{\omega}$;
    \item Step 2: Control the KL divergence between models induced by the different $P_\omega$;
    \item Step 3: Check that \eqref{asst:src} with parameters $\beta_X$ and $B_X$ and \eqref{asst:mom} hold.
\end{itemize}
\textbf{Step 1: Separation.}
If $\rho(\omega,\omega')=\sum_{i=1}^{\ell} (\omega_i-\omega_i')^2 \ge \ell/8$ (this will be ensured later by Lemma~\ref{lma:gilbert_varshamov}), then
\(
\|h_\omega-h_{\omega'}\|_{L_2(X)}^2
= 32\varepsilon \ell^{-1}\rho(\omega,\omega')
\ge 4\varepsilon.
\)

\medskip \noindent
\textbf{Step 2: Kullback--Leibler control.} Let $P_\omega^{(m,n)}$ denote the joint law of $(\cD_1,\cD_2)$ under model $\omega$.
Since $\cD_1$ has the same law under all hypotheses,
\[
\mathrm{KL}\!\left(P_\omega^{(m,n)},P_{\omega'}^{(m,n)}\right)
=
\mathrm{KL}\!\left(P_{\omega,Z,Y}^{\otimes n},P_{\omega',Z,Y}^{\otimes n}\right).
\]
Recall that under $P_{\omega}$, for all almost all $z \in E_Z$, $\operatorname{Law}(Y|Z=z) = \cN(\langle h_\omega, F^{\dagger}(z) \rangle_{\cH_X}, s_0^2)$. For one second-stage observation, we therefore have,
\begin{equation*}
    \begin{aligned}
\mathrm{KL}(P_{\omega,Z,Y},P_{\omega',Z,Y})  &= \int_{E_Z} \mathrm{KL}(P_{\omega,Y\mid Z=z},P_{\omega',Y\mid Z=z}) d\pi_Z(z)\\
&= \frac{1}{2s_0^2}\int_{E_Z} \langle h_\omega-h_{\omega'}, F^\dagger(z) \rangle_{\cH_X}^2 d\pi_Z(z)= \frac{1}{2s_0^2}\|C_F^{\dagger\,1/2}(h_\omega-h_{\omega'})\|_{\cH_X}^2.
    \end{aligned}
\end{equation*}
Using $C_F^\dagger \preceq R_1 C_X^\gamma$,
\(
\|C_F^{\dagger\,1/2}(h_\omega-h_{\omega'})\|_{\cH_X}^2
\le
R_1\|C_X^{\gamma/2}(h_\omega-h_{\omega'})\|_{\cH_X}^2.
\)
Now
\[
\|C_X^{\gamma/2}(h_\omega-h_{\omega'})\|_{\cH_X}^2
=
\frac{32\varepsilon}{\ell}
\sum_{i=1}^\ell (\omega_i-\omega_i')^2 \mu_{X,i+\ell}^{\gamma-1} \leq 32 \varepsilon \mu_{X,\ell}^{\gamma - 1} 
\le 
32 \overline{c}_X^{\gamma-1}\varepsilon\, \ell^{-(\gamma-1)/p_X},
\]
where we used \eqref{asst:evd}.
Hence
\(
\mathrm{KL}\!\left(P_\omega^{(m,n)},P_{\omega'}^{(m,n)}\right)
\le
C_{\mathrm{KL}}\,n\,\varepsilon\, \ell^{-(\gamma-1)/p_X},
\)
for a constant $C_{\mathrm{KL}}>0$ depending only on $(R_1,\overline{c}_X,\gamma,s_0)$.

\medskip \noindent
\textbf{Step 3: Source condition and moment condition.}
By construction, each model satisfies \eqref{asst:mom} with parameters $(\sigma,L)$.
For the source condition,
\[
\|C_X^{-(\beta_X-1)/2}h_\omega\|_{\cH_X}^2
=
\frac{32\varepsilon}{\ell}\sum_{i=1}^\ell \omega_i^2 \mu_{X,i+\ell}^{-\beta_X}.
\]
By \eqref{asst:evd+},
\(
\|C_X^{-(\beta_X-1)/2}h_\omega\|_{\cH_X}^2
\le 32 \varepsilon \mu_{X, 2\ell}^{-\beta_X} \leq
32 \underline{c}_X^{-\beta_X}2^{\beta_X/p_X}\,
\varepsilon\, \ell^{\beta_X/p_X}.
\)
Let
\[
u\doteq \frac{p_X}{\beta_X},
\qquad
U\doteq \left(\frac{B_X^2 \underline{c}_X^{\beta_X}}{32\,2^{\beta_X/p_X}}\right)^u,
\qquad
\ell_\varepsilon \doteq \lfloor U\varepsilon^{-u}\rfloor.
\]
Then for all sufficiently small $\varepsilon$, every $h_\omega$ with
$\omega\in\{0,1\}^{\ell_\varepsilon}$ satisfies \eqref{asst:src} with parameters
$(B_X,\beta_X)$.

\medskip \noindent
\textbf{Putting everything together}
To conclude we use the following theorem that is derived from \citet[][Proposition 2.3]{tsybakov2009nonparametric} and \cite[][Theorem 20]{fischer2020sobolev}.
\begin{theorem}
Let $M \geq2,(\Omega, \mathcal{A})$ be a measurable space, $P_0, P_1, \ldots, P_M$ be probability measures on $(\Omega, \mathcal{A})$ with $P_j \ll P_0$ for all $j=1, \ldots, M$, and $0<\alpha_*<\infty$ with
$$
\frac{1}{M} \sum_{j=1}^M \mathrm{KL}\left(P_j, P_0\right) \leq \alpha_* \text {. }
$$
Then, for all measurable functions $\Psi: \Omega \rightarrow\{0,1, \ldots, M\}$, the following bound is satisfied
$$
\max _{j=0,1, \ldots, M} P_{j}(s \in \Omega: \Psi(s) \neq j) \geq\frac{\sqrt{M}}{1+\sqrt{M}}\left(1-\frac{3 \alpha_*}{\log (M)}-\frac{1}{2 \log (M)}\right) .
$$
\end{theorem}
To obtain the distributions $P_0, P_1, \ldots, P_M$ we use the following lemma \citep[][Lemma 2.9]{tsybakov2009nonparametric}.

\begin{lemma}[Gilbert-Varshamov Bound] \label{lma:gilbert_varshamov}For $\ell \geq8$ there exists some $M \geq2^{\ell / 8}$ and some binary strings $\omega^{(0)}, \ldots, \omega^{(M)} \in\{0,1\}^\ell$ with $\omega^{(0)}=(0, \ldots, 0)$ and
$
\sum_{i=1}^\ell\left(\omega_i^{(j)}-\omega_i^{(k)}\right)^2 \geq \ell / 8,
$
for all $j \neq k$, where $\omega^{(k)}=\left(\omega_1^{(k)}, \ldots, \omega_\ell^{(k)}\right)$.
\end{lemma}
Define $\varepsilon_0\doteq\min \left\{1,(U / 9)^{1 / u}\right\}$ and $\ell_{\varepsilon}\doteq\left\lfloor U \varepsilon^{-u}\right\rfloor$. Now, we fix an $n \geq1$ and a $0<\varepsilon \leq \varepsilon_0$. Since $\ell_{\varepsilon} \geq9$, the Gilbert-Varshamov Bound Lemma yields at least $M_{\varepsilon}\doteq\left\lceil 2^{\ell_{\varepsilon} / 8}\right\rceil \geq3$ binary strings $\omega^{(0)}, \omega^{(1)}, \ldots, \omega^{\left(M_{\varepsilon}\right)} \in\{0,1\}^{\ell_{\varepsilon}}$ satisfying the Gilbert-Varshamov Bound. For $j=0,1, \ldots, M_{\varepsilon}$, the corresponding functions $h_j\doteq h_{\omega^{(j)}}$ satisfy the bound $\|C_X^{-\frac{\beta_X-1}{2}}h_j \|_{\cH_X} \leq B_X$. Due to the definitions of $M_{\varepsilon}, \ell_{\varepsilon}$ and $\ell_{\varepsilon} \geq9$ we get $8 U / 9 \varepsilon^{-u} \leq \ell_{\varepsilon} \leq U \varepsilon^{-u}$ and
$$
2^{U / 9 \varepsilon^{-u}} \leq 2^{\ell_{\varepsilon} / 8} \leq M_{\varepsilon} \leq 2^{\ell_{\varepsilon} / 4} \leq 2^{U / 3 \varepsilon^{-u}}.
$$
We can simplify it as $2^{C_2 \varepsilon^{-u}} \leq M_{\varepsilon} \leq 2^{3 C_2 \varepsilon^{-u}}$ with $C_2\doteq U/9$.
Denote
\(
  P^{(m,n)}_j
  \;\doteq\; P^{(m,n)}_{\omega^{(j)}}, j=0,1,\dots,M_\varepsilon.
\)
We have,
$$
\frac{1}{M_{\varepsilon}} \sum_{j=1}^{M_{\varepsilon}} \mathrm{KL}(P^{(m,n)}_j,P^{(m,n)}_0) \leq \frac{n}{s_0^2}16\overline{c}_X^{\gamma - 1 } \varepsilon \ell_{\varepsilon}^{-\frac{\gamma - 1 }{p_X}}.
$$
Furthermore, using $\ell_{\varepsilon} \geq8 U / 9 \varepsilon^{-u}$ we find
$$
\frac{1}{M_{\varepsilon}} \sum_{j=1}^{M_{\varepsilon}} \mathrm{KL}(P^{(m,n)}_j,P^{(m,n)}_0) \leq C' n \varepsilon^{1+\frac{\gamma - 1 }{\beta_X }}=: \alpha_*
$$
with $C'\doteq\frac{16 \overline{c}_X^{\gamma-1} 9^{\frac{\gamma-1}{p_X}}}{s_0^2(8 U)^{\frac{\gamma-1}{p_X}}}$. For a measurable function 
$$
\Psi: \Omega \rightarrow\left\{0,1, \ldots, M_{\varepsilon}\right\}, \qquad \Omega \doteq (E_Z\times E_X)^m \times (E_Z\times \mathbb{R})^n,
$$
since $M_{\varepsilon} \geq2^{C_2 \varepsilon^{-u}}$, it yields
$$
\begin{aligned}
\max _{j=0,1, \ldots, M_{\varepsilon}} P^{(m,n)}_j(D: \Psi(D) \neq j) & \geq\frac{\sqrt{M_{\varepsilon}}}{1+\sqrt{M_{\varepsilon}}}\left(1-\frac{3 C' n \varepsilon^{1+\frac{\gamma - 1 }{\beta_X }}}{\log \left(M_{\varepsilon}\right)}-\frac{1}{2 \log \left(M_{\varepsilon}\right)}\right) \\
& \geq\frac{\sqrt{M_{\varepsilon}}}{1+\sqrt{M_{\varepsilon}}}\left(1-\frac{3 C'}{C_2 \log (2)} n \varepsilon^{1+ \frac{\gamma - 1 }{\beta_X  } +u}-\frac{1}{2 \log \left(M_{\varepsilon}\right)}\right) .
\end{aligned}
$$
Since $1+\frac{\gamma - 1 }{\beta_X } +u=\frac{\beta_X - 1 + \gamma +p_X}{\beta_X}$, we get
\begin{equation} \label{eq:reduc}
\max _{j=0,1, \ldots, M_{\varepsilon}} P^{(m,n)}_j(D: \Psi(D) \neq j)  \geq\frac{\sqrt{M_{\varepsilon}}}{1+\sqrt{M_{\varepsilon}}}\left(1-C_1 n \varepsilon^{\frac{ \beta_X - 1 + \gamma +p_X}{\beta_X}}-\frac{1}{2 \log \left(M_{\varepsilon}\right)}\right) .
\end{equation}
for $C_1\doteq\frac{3 C'}{C_2 \log (2)}$. To conclude the proof we follow the general reduction scheme from \citet[][Section 2.2]{tsybakov2009nonparametric}. Let $(\cD_1,\cD_2)\mapsto \widehat h(\cD_1,\cD_2)$ be an arbitrary (measurable) NPIV learning
method.\footnote{In our construction the first-stage sample $\cD_1$ has the same distribution
under all hypotheses. Hence $\cD_1$ carries no information about the index $j$; the argument
below nevertheless allows $\widehat h$ to depend on $(\cD_1,\cD_2)$.} Set
\[
r \doteq \frac{\beta_X}{\beta_X+\gamma+p_X-1},
\qquad
\varepsilon_n \doteq \tau\, n^{-r},
\]
and fix $\tau>0$ and $n\ge 1$ such that $\varepsilon_n\le \varepsilon_0$. For $\varepsilon=\varepsilon_n$, let $\{P^{(m,n)}_j\}_{j=0}^{M_n}$ with $M_n:=M_{\varepsilon_n}$
be the family of NPIV models constructed above, and denote by $\{h_j\}_{j=0}^{M_n}$ the
corresponding structural functions. Define the (measurable) classifier
\[
\Psi(\cD_1,\cD_2)
\doteq
\arg\min_{j\in\{0,1,\dots,M_n\}}
\big\|\widehat h(\cD_1,\cD_2)-h_j\big\|_{L_2(X)}.
\]
If $\Psi(\cD_1,\cD_2)\neq j$, then the separation property of the packing set gives
\[
2\sqrt{\varepsilon_n}
\;\le\;
\big\|h_{\Psi(\cD_1,\cD_2)}-h_j\big\|_{L_2(X)}.
\]
On the other hand, by the definition of $\Psi(\cD_1,\cD_2)$ and the triangle inequality,
$$
\begin{aligned}
\big\|h_{\Psi(\cD_1,\cD_2)}-h_j\big\|_{L_2(X)}
&\;\le\;
\big\|h_{\Psi(\cD_1,\cD_2)}-\widehat{h}(\cD_1,\cD_2)\big\|_{L_2(X)}
+\big\|\widehat{h}(\cD_1,\cD_2)-h_j\big\|_{L_2(X)}
\\&\;\le\;
2\big\|\widehat{h}(\cD_1,\cD_2)-h_j\big\|_{L_2(X)}.
\end{aligned}
$$
Consequently, for every $j=0,1,\dots,M_n$,
\[
P^{(m,n)}_j\!\left(
\big\|\widehat{h}(\cD_1,\cD_2)-h_j\big\|_{L_2(X)}^2 \ge \varepsilon_n
\right)
\;\ge\;
P^{(m,n)}_j\big(\Psi(\cD_1,\cD_2)\neq j\big).
\]
\medskip \noindent
\Cref{eq:reduc} yields
\[
\max_{j=0,1,\dots,M_n} P^{(m,n)}_j\big(\Psi(\cD_1,\cD_2)\neq j\big)
\;\ge\;
\frac{\sqrt{M_n}}{1+\sqrt{M_n}}
\left(1 - C_1 \tau^{1/r} - \frac{1}{2\log(M_n)}\right),
\]
where we used that $n\,\varepsilon_n^{(\beta_X-1+\gamma+p_X)/\beta_X}=\tau^{1/r}$.
Combining this with the reduction inequality
\[
P^{(m,n)}_{j}\!\left(\|\hat h(\cD_1,\cD_2)-h_j\|^2_{L_2(X)}\ge \varepsilon_n\right)
\;\ge\;
P^{(m,n)}_{j}\!\left(\Psi(\cD_1,\cD_2)\neq j\right),
\qquad j=0,1,\dots,M_n,
\]
we obtain
\[
\max_{j=0,1,\dots,M_n} 
P^{(m,n)}_{j}\!\left(\|\hat h(\cD_1,\cD_2)-h_j\|^2_{L_2(X)}\ge \varepsilon_n\right)
\;\ge\;
\frac{\sqrt{M_n}}{1+\sqrt{M_n}}
\left(1 - C_1 \tau^{1/r} - \frac{1}{2\log(M_n)}\right).
\]

\medskip
\noindent\textbf{Final high-probability simplification.}
Note that
\(
\frac{\sqrt{M_n}}{1+\sqrt{M_n}} \ge 1 - M_n^{-1/2}.
\)
Since $M_n=M_{\varepsilon_n}\to\infty$ as $n\to\infty$, we may choose $n$ large enough (depending on
$\tau$ and the fixed constants) so that
\(
M_n^{-1/2} \le C_1\tau^{1/r}
\)
and
\(
\frac{1}{2\log(M_n)}\le C_1\tau^{1/r}.
\)
For such $n$,
\[
\frac{\sqrt{M_n}}{1+\sqrt{M_n}}
\left(1 - C_1 \tau^{1/r} - \frac{1}{2\log(M_n)}\right)
\;\ge\;
\left(1-C_1\tau^{1/r}\right)\left(1-2C_1\tau^{1/r}\right)
\;\ge\;
1 - J_0\tau^{1/r},
\]
where we used $(1-a)(1-b)\ge 1-a-b$ and set $J_0:=3C_1$.
Hence, there exists an index $j_n\in\{0,1,\dots,M_n\}$ such that
\(
P^{(m,n)}_{j_n}\!\left(\|\hat h(\cD_1,\cD_2)-h_{j_n}\|^2_{L_2(X)}\ge \varepsilon_n\right)
\;\ge\;
1 - J_0\tau^{1/r}.
\)
Setting $h_\ast:=h_{j_n}$ and recalling $\varepsilon_n=\tau n^{-r}$ concludes the proof.


\medskip
\noindent\textbf{Comparison with \citet{chen2011rate}.}
A key distinction from classical NPIV lower bounds is methodological: \citet{chen2011rate}
derive their NPIV minimax lower bound by first reducing the NPIV experiment to an
auxiliary nonparametric indirect regression (NPIR) problem in which the conditional expectation
operator is treated as known, and then applying Assouad's cube, yielding a bound in expectation.
In contrast, our construction works \emph{directly} with the split-sample NPIV experiment and does
not pass through an NPIR reduction: the first-stage sample is explicitly made uninformative about
the hypothesis index, and the difficulty is shown to be already bottlenecked by the second-stage
experiment.
Moreover, we use Fano's method to obtain a lower bound with high probability,
in the same spirit as \citet{fischer2020sobolev} for least-squares regression.





\section{Some Bounds} \label{sec:more_bounds}

\begin{lemma}\label{lma:con_cf}
Let $g_\lambda$ be defined as in \Cref{eq:cte_fisher}. Then, for $\tau \geq 1, \lambda \in (0,1]$, and $n \geq 1$, the following operator norm bound is satisfied with $P^n$-probability at least $1-2 e^{-\tau}$,
\begin{equation} 
\left\|\left(C_F+\lambda \operatorname{Id} \right)^{-\frac{1}{2}}\left(C_F-\hat C_F\right)\left(C_F+\lambda \operatorname{Id} \right)^{-\frac{1}{2}}\right\|_{\cH_X \to \cH_X} \leq \frac{4 \tau g_\lambda}{3 n \lambda}+\sqrt{\frac{2 \tau g_\lambda}{n \lambda}} \label{eq:s2con}   
\end{equation}
In particular, for $n \geq 8 \tau g_\lambda \lambda^{-1}$, with probability at least $1-2e^{-\tau}$,
\begin{equation*} 
\left\|\left(C_F+\lambda \operatorname{Id} \right)^{-\frac{1}{2}}\left(C_F-\hat C_F\right)\left(C_F+\lambda \operatorname{Id} \right)^{-\frac{1}{2}}\right\|_{\cH_X \to \cH_X} \leq \frac{2}{3}.
\end{equation*}
\end{lemma}

\begin{proof} The bound is obtained directly from \citet[][Lemma 17]{fischer2020sobolev} applied to $C_F$, with $\alpha = 1$ in their notation, and using that almost surely $\|F_*(Z)\|_{\cH_X} \leq \E[\|\phi_X(X)\|_{\cH_X} \mid Z] \leq 1$. For $n \geq 8 \tau g_\lambda \lambda^{-1}$, we obtain that with probability at least $1-2e^{-\tau}$,
\begin{equation*} 
\left\|\left(C_F+\lambda \operatorname{Id}\right)^{-\frac{1}{2}}\left(C_F-\hat C_F\right)\left(C_F+\lambda \operatorname{Id}\right)^{-\frac{1}{2}}\right\|_{\cH_X \to \cH_X} \leq \frac{4 \tau g_\lambda}{3 n \lambda}+\sqrt{\frac{2 \tau g_\lambda}{n \lambda}} 
\leq \frac{4}{3}\cdot \frac{1}{8} + \sqrt{\frac{2}{8}} = \frac{2}{3}. \qedhere
\end{equation*}
\end{proof}

\begin{lemma}\label{lma:emp_sub_app}
Let Assumptions~\eqref{asst:srcz} and \eqref{asst:embz} hold with $\alpha_Z<\beta_Z$.
Condition on the first-stage sample $\cD_1$ used to construct $\hat F_\xi$.
Assume that $\|F_*-\hat F_\xi\|_{L_2(Z;\cH_X)}\le 1$ and $\|F_*-\hat F_\xi\|_{\alpha_Z}\le 1$.
Then for any $\tau \geq 1$, with $P^n(\cdot\mid \cD_1)$-probability at least $1-4e^{-\tau}$,
\[
\|\hat C_F-\hat C_{\hat F}\|_{\cH_X\to \cH_X}
\le J\Big(\sqrt{\tfrac{\tau}{n}}\|F_*-\hat F_\xi\|_{\alpha_Z}
      +\|F_*-\hat F_\xi\|_{L_2(Z;\cH_X)}\Big),
\]
where $J$ depends on $A_Z,B_Z,\alpha_Z,\beta_Z$.
\end{lemma}

\begin{proof}
Set
\(
\Delta F \doteq F_*-\hat F_\xi .
\)
For every $z\in E_Z$,
\[
F_*(z)\otimes F_*(z)-\hat F_\xi(z)\otimes \hat F_\xi(z)
=
F_*(z)\otimes \Delta F(z)
+
\Delta F(z)\otimes F_*(z)
-
\Delta F(z)\otimes \Delta F(z).
\]
Therefore,
\[
\hat C_F-\hat C_{\hat F}
=
\frac1n\sum_{i=1}^n
\Big(
F_*(z_i)\otimes \Delta F(z_i)
+
\Delta F(z_i)\otimes F_*(z_i)
-
\Delta F(z_i)\otimes \Delta F(z_i)
\Big),
\]
and hence
\begin{align*}
\|\hat C_F-\hat C_{\hat F}\|_{\cH_X\to\cH_X}
&\le
\frac{2}{n}\sum_{i=1}^n \|F_*(z_i)\|_{\cH_X}\,\|\Delta F(z_i)\|_{\cH_X}
+
\frac1n\sum_{i=1}^n \|\Delta F(z_i)\|_{\cH_X}^2 \\
&\le
\frac{2A_ZB_Z}{n}\sum_{i=1}^n \|\Delta F(z_i)\|_{\cH_X}
+
\frac1n\sum_{i=1}^n \|\Delta F(z_i)\|_{\cH_X}^2,
\end{align*}
where we used \Cref{lma:rhks_as} in the last step. Now define
\(
\xi_i \doteq \|\Delta F(z_i)\|_{\cH_X}, i=1,\dots,n.
\)
By \Cref{lma:rhks_as},
\(
0\le \xi_i \le A_Z\|\Delta F\|_{\alpha_Z}
\)
a.s. Hoeffding's inequality therefore gives, for $\tau\ge 1$, the event
\[
\mathcal E_{6,1}
\doteq
\left\{
\left|
\frac1n\sum_{i=1}^n \xi_i - \E[\xi_i\mid \cD_1]
\right|
\le
A_Z\|\Delta F\|_{\alpha_Z}\sqrt{\frac{\tau}{n}}
\right\},
\]
satisfying
\(
P^n(\mathcal E_{6,1}^c\mid \cD_1)\le 2e^{-\tau}.
\)
Likewise, since
\(
0\le \xi_i^2 \le A_Z^2\|\Delta F\|_{\alpha_Z}^2
\)
a.s., Hoeffding's inequality gives the event
\[
\mathcal E_{6,2}
\doteq
\left\{
\left|
\frac1n\sum_{i=1}^n \xi_i^2 - \E[\xi_i^2\mid \cD_1]
\right|
\le
A_Z^2\|\Delta F\|_{\alpha_Z}^2\sqrt{\frac{\tau}{n}}
\right\},
\]
with
\(
P^n(\mathcal E_{6,2}^c\mid \cD_1)\le 2e^{-\tau}.
\)
Set
\(
\mathcal E_6 \doteq \mathcal E_{6,1}\cap \mathcal E_{6,2}.
\)
Then
\(
P^n(\mathcal E_6^c\mid \cD_1)\le 4e^{-\tau}.
\)
On $\mathcal E_6$, using that by Jensen's inequality
\(
\E[\xi_i\mid \cD_1]\le \|\Delta F\|_{L_2(Z;\cH_X)},
\E[\xi_i^2\mid \cD_1]=\|\Delta F\|_{L_2(Z;\cH_X)}^2,
\)
we get
\begin{align*}
\|\hat C_F-\hat C_{\hat F}\|_{\cH_X\to\cH_X}
&\le
2A_ZB_Z
\left(
A_Z\|\Delta F\|_{\alpha_Z}\sqrt{\frac{\tau}{n}}
+
\|\Delta F\|_{L_2(Z;\cH_X)}
\right) \\
&\quad+
A_Z^2\|\Delta F\|_{\alpha_Z}^2\sqrt{\frac{\tau}{n}}
+
\|\Delta F\|_{L_2(Z;\cH_X)}^2.
\end{align*}
Under the standing assumptions
\(
\|\Delta F\|_{L_2(Z;\cH_X)}\le 1,
\|\Delta F\|_{\alpha_Z}\le 1,
\)
this simplifies to
\[
\|\hat C_F-\hat C_{\hat F}\|_{\cH_X\to\cH_X}
\le
J\left(
\|\Delta F\|_{L_2(Z;\cH_X)}
+
\|\Delta F\|_{\alpha_Z}\sqrt{\frac{\tau}{n}}
\right),
\]
for a constant $J$ depending only on $A_Z,B_Z,\alpha_Z,\beta_Z$. For later use, note also that on $\mathcal E_{6,1}$,
\begin{align*}
&\left\|
\frac1n(\Phi_{\hat F}-\Phi_{F_*})^*\Phi_{F_*}
\right\|_{\cH_X\to\cH_X}
=
\left\|
\frac1n\sum_{i=1}^n (\hat F_\xi(z_i)-F_*(z_i))\otimes F_*(z_i)
\right\|_{\cH_X\to\cH_X} \\
&\le
\frac1n\sum_{i=1}^n \|\Delta F(z_i)\|_{\cH_X}\,\|F_*(z_i)\|_{\cH_X} \le
A_ZB_Z
\left(
A_Z\|\Delta F\|_{\alpha_Z}\sqrt{\frac{\tau}{n}}
+
\|\Delta F\|_{L_2(Z;\cH_X)}
\right).
\end{align*}
This auxiliary bound will be used in the proof of \Cref{theo:s1e1}.
\end{proof}

\begin{lemma}\label{lma:chatc_2}
Let Assumptions~\eqref{asst:srcz} and \eqref{asst:embz} hold with $\alpha_Z<\beta_Z$.
Condition on the first-stage sample $\cD_1$ used to construct $\hat F_\xi$.
Assume that $\|F_*-\hat F_\xi\|_{L_2(Z;\cH_X)}\le 1$ and $\|F_*-\hat F_\xi\|_{\alpha_Z}\le 1$.
Then for any $\tau\ge 1$ and $\lambda \in (0,1]$, with $P^n(\cdot\mid \cD_1)$-probability at least $1-6e^{-\tau}$,
\begin{equation}
    \begin{aligned}
&\left\|\left( C_{ F} + \lambda \operatorname{Id}\right)^{-\frac{1}{2}} \left( C_F-\hat C_{\hat F} \right)\left( C_{ F} + \lambda \operatorname{Id}\right)^{-\frac{1}{2}}\right\|_{\cH_X \to \cH_X} \\
&\leq J\left(\frac{4 \tau g_\lambda}{3 n \lambda}+\sqrt{\frac{2 \tau g_\lambda}{n \lambda}} + \sqrt{\frac{\tau\|F_*-\hat F_{\xi}\|_{\alpha_Z}^2}{n\lambda^2}} + \frac{\|F_*-\hat F_{\xi}\|_{L_2(Z; \cH_X)}}{\lambda}\right)  \label{eq:s1s2con}   
    \end{aligned}
\end{equation}
In particular, if Eq.~\eqref{eq:m_con} holds, then with $P^n(\cdot\mid \cD_1)$-probability at least $1-6e^{-\tau}$,
\[
\left\|\left(C_F+\lambda \operatorname{Id}\right)^{-\frac{1}{2}}
\left(C_F-\hat C_{\hat F}\right)
\left(C_F+\lambda \operatorname{Id}\right)^{-\frac{1}{2}}\right\|_{\cH_X\to \cH_X}
\le \frac{5}{6}.
\]
\end{lemma}

\begin{proof}
Let $A_{\lambda} \doteq \left( C_{ F} + \lambda \operatorname{Id}\right)^{-\frac{1}{2}} \left( C_F-\hat C_{\hat F} \right)\left( C_{ F} + \lambda \operatorname{Id}\right)^{-\frac{1}{2}}$. We have,  
\begin{equation*}
    \begin{aligned}
&\left\|A_{\lambda}   \right\|_{\cH_X \to \cH_X} 
= \left\|\left( C_{ F} + \lambda \operatorname{Id}\right)^{-\frac{1}{2}} \left( C_F - \hat C_F + \hat C_F- \hat C_{\hat F} \right)\left( C_F + \lambda \operatorname{Id}\right)^{-\frac{1}{2}}\right\|_{\cH_X \to \cH_X} \\
&\leq \left\|\left( C_{ F} + \lambda \operatorname{Id}\right)^{-\frac{1}{2}} \left( C_F - \hat C_F\right)\left( C_F + \lambda \operatorname{Id}\right)^{-\frac{1}{2}}\right\| +  \left\|\left( C_{ F} + \lambda \operatorname{Id}\right)^{-\frac{1}{2}} \left( \hat C_{\hat F} - \hat C_F \right)\left( C_F + \lambda \operatorname{Id}\right)^{-\frac{1}{2}}\right\| \\ &\leq \left\|\left( C_{ F} + \lambda \operatorname{Id}\right)^{-\frac{1}{2}} \left( C_F - \hat C_F\right)\left( C_F + \lambda \operatorname{Id}\right)^{-\frac{1}{2}}\right\|_{\cH_X \to \cH_X} + \lambda^{-1}  \left\| \left( \hat C_{\hat F} - \hat C_F \right)\right\|_{\cH_X \to \cH_X}.
    \end{aligned}
\end{equation*}
Apply Eq.~\eqref{eq:s2con} to the first term and Lemma~\ref{lma:emp_sub_app} to the second term, we obtain Eq.~\eqref{eq:s1s2con}. In addition, for the first term, for $\tau \geq 1, \lambda \in (0,1]$ and $n \geq 8 \tau g_\lambda \lambda^{-1}$, with probability at least $1-2e^{-\tau}$,
\begin{equation*}
    \left\|\left(C_F+\lambda \operatorname{Id}\right)^{-1 / 2}\left(C_F-\hat C_F\right)\left(C_F+\lambda \operatorname{Id}\right)^{-1 / 2}\right\|_{\cH_X \to \cH_X} \leq \frac{2}{3}.
\end{equation*}
For the second term, under the assumptions that \( \|F_* - \hat F_{\xi}\|_{L_2(Z;\cH_X)} \leq 1\) and $\|F_* - \hat F_{\xi}\|_{\alpha_Z} \leq 1$, with $P^n$-probability at least $1- 4e^{-\tau}$, it holds
\[ \left\| \hat C_F - \hat C_{\hat F}\right\|_{\cH_X \to \cH_X} \leq J\left(\sqrt{\frac{\tau}{n}} \|F_*-\hat F_{\xi}\|_{\alpha_Z} + \|F_*-\hat F_{\xi}\|_{L_2(Z; \cH_X)}\right).\]  
Under the constraints of Eq.~\eqref{eq:m_con}, it implies that with probability at least $1-6e^{-\tau}$, $\left\| A_{\lambda}   \right\|_{\cH_X \to \cH_X}  \leq \frac{5}{6}.$
\end{proof}

\begin{lemma}\label{lma:chatc3} Let Assumptions~\eqref{asst:srcz} and \eqref{asst:embz} hold with $\alpha_Z < \beta_Z$. Condition on the first-stage sample $\cD_1$ used to construct $\hat F_\xi$.
Assume that $\|F_*-\hat F_\xi\|_{L_2(Z;\cH_X)}\le 1$ and $\|F_*-\hat F_\xi\|_{\alpha_Z}\le 1$.
For any $\tau\ge 1$ and $\lambda \in (0,1]$, if the constraints in Eq.~\eqref{eq:m_con} hold, then with
$P^n(\cdot\mid \cD_1)$-probability at least $1-4e^{-\tau}$,
 \[ \left\|\left( \hat C_{\hat F} + \lambda \operatorname{Id}\right)^{-\frac{1}{2}}\left( \hat C_F + \lambda \operatorname{Id}\right)^{\frac{1}{2}}\right\|_{\cH_X \to \cH_X}  \leq \sqrt{\frac{6}{5}}.\]    
\end{lemma}

\begin{proof} By Lemma~\ref{lma:AB_norm}, we obtain that \[\left\|\left( \hat C_{\hat F} + \lambda \operatorname{Id}\right)^{-\frac{1}{2}}\left( \hat C_F + \lambda \operatorname{Id}\right)^{\frac{1}{2}}\right\|_{\cH_X \to \cH_X} \leq (1-t)^{-\frac{1}{2}},\] where $t = \left\|\left( \hat C_F + \lambda \operatorname{Id}\right)^{-\frac{1}{2}} \left(\hat C_F - \hat C_{\hat F} \right) \left( \hat C_F + \lambda \operatorname{Id}\right)^{-\frac{1}{2}}\right\|_{\cH_X \to \cH_X} \leq \lambda^{-1}\left\|\hat C_F - \hat C_{\hat F} \right\|_{\cH_X \to \cH_X} $. 
By Lemma~\ref{lma:emp_sub_app}, under the assumptions that \( \|F_* - \hat F_{\xi}\|_{L_2(Z;\cH_X)} \leq 1\) and $\|F_* - \hat F_{\xi}\|_{\alpha_Z} \leq 1$, with $P^n$-probability at least $1- 4e^{-\tau}$, it holds
\[ \left\| \hat C_F - \hat C_{\hat F}\right\|_{\cH_X \to \cH_X} \leq J\left(\sqrt{\frac{\tau}{n}} \|F_*-\hat F_{\xi}\|_{\alpha_Z} + \|F_*-\hat F_{\xi}\|_{L_2(Z; \cH_X)}\right).\]
Under the constraints of Eq.~\eqref{eq:m_con}, it implies that with probability at least $1-4e^{-\tau}$, $t \leq \frac{1}{6}$, which concludes the proof.
\end{proof}

\begin{lemma}\label{lma:chatc} Let Assumptions~\eqref{asst:srcz} and \eqref{asst:embz} hold with $\alpha_Z<\beta_Z$.
Condition on the first-stage sample $\cD_1$ used to construct $\hat F_\xi$.
For any $\tau\ge 1$ and $\lambda \in (0,1]$, if the constraints in Eq.~\eqref{eq:m_con} hold, then with
$P^n(\cdot\mid \cD_1)$-probability at least $1-6e^{-\tau}$,
 \[ \left\|\left( C_{ F} + \lambda \operatorname{Id}\right)^{\frac{1}{2}}\left( \hat C_{\hat F} + \lambda \operatorname{Id}\right)^{-\frac{1}{2}}\right\|_{\cH_X \to \cH_X} \leq 3.\] 
\end{lemma}

\begin{proof}
By Lemma~\ref{lma:AB_norm}, \(B_{\lambda} \doteq \left\|\left( C_{ F} + \lambda \operatorname{Id}\right)^{\frac{1}{2}}\left( \hat C_{\hat F} + \lambda \operatorname{Id}\right)^{-\frac{1}{2}}\right\|_{\cH_X \to \cH_X} \leq (1-t)^{-\frac{1}{2}},\) when $t \doteq \|A_{\lambda}\|_{\cH_X \to \cH_X}<1$, with $A_{\lambda} \doteq \left( C_{ F} + \lambda \operatorname{Id}\right)^{-\frac{1}{2}} \left( C_F-\hat C_{\hat F} \right)\left( C_{ F} + \lambda \operatorname{Id}\right)^{-\frac{1}{2}}$. By Lemma~\ref{lma:chatc_2}, under the constraints of Eq.~\eqref{eq:m_con}, with probability at least $1-6e^{-\tau}$, $t \leq 5/6$, and therefore,
\( B_{\lambda}\leq \sqrt{6} \leq 3.\)
\end{proof}

\section{Auxiliary Results}\label{sec:auxiliary}


\begin{lemma}\label{lma:rhks_as}
Let $F \in \cG$, then, for $\pi_Z$-almost all $z \in E_Z$, $\|F(z)\|_{\cH_X} \leq \|F\|_{\cG}.$ Alternatively, if \eqref{asst:embz} holds and $F$ satisfies \eqref{asst:srcz} with $\alpha_Z < \beta_Z$, then for $\pi_Z$-almost all $z \in E_Z$, $\|F(z)\|_{\cH_X} \leq A_Z\|F\|_{\alpha_Z} \leq  A_Z B_Z$.
\end{lemma}
\begin{proof}
By Theorem~\ref{theo:operep}, since $F \in \cG$, there is an operator $C \in S_2(\cH_Z, \cH_X)$ such that for all $z \in E_Z$, $F(z) = C \phi_Z(z)$ and $\|F\|_{\cG} = \|C\|_{S_2}$. Therefore, for $\pi_Z$-almost all $z \in E_Z$
$$
\|F(z)\|_{\cH_X} = \|C\phi_Z(z)\|_{\cH_X} \leq \|C\|_{\cH_Z \to \cH_X} \leq \|C\|_{S_2} = \|F\|_{\cG},
$$
where we used Assumption~\ref{asst:rkhs}: $k_Z(z,z) \leq 1$ for $\pi_Z$-almost all $z \in E_Z$. 
Under \eqref{asst:embz}, it is shown in \citet[Lemma 4]{lietal2022optimal} that for all functions $F: E_Z \to \cH_X$ such that $\|F\|_{\alpha_Z} < +\infty$,
$
\|F\|_{L_{\infty}(Z;\cH_X)} \leq A_Z \|F\|_{\alpha_Z}.
$
To conclude we show that since $F$ satisfies \eqref{asst:srcz} with $\alpha_Z < \beta_Z$ then $\|F\|_{\alpha_Z} \leq \|F\|_{\beta_Z}$. As $F \in L_2(Z; \cH_X)$, by \Cref{eq:vv_interpolation_norm}, there is an operator $C \in S_2(\overline{\cR(L_Z)}, \cH_X)$ such that
$
\|F\|_{\theta} = \|CL_Z^{-\theta / 2}\|_{S_2(L_2(Z), \mathcal{H}_X)}.
$
Exploiting the spectral decomposition of $L_Z$ (see Eq.~\eqref{eq:SVD}) and using the fact that $\{\sqrt{\mu_{X,i}}e_{X,i} \otimes [e_{Z,j}]\}_{i \geq 1, j \geq 1}$ is an ONB of $S_2(\overline{\cR(L_Z)}, \cH_X)$ (see Definition~\ref{def:appendix_tensor}), we have
\begin{align}
    \|F\|_{\alpha_Z}^2 &= \sum_{i \geq 1} \sum_{j \geq 1} \mu_{Z,i}^{-\alpha_Z} \langle C, \sqrt{\mu_{X,i}}e_{X,i} \otimes [e_{Z,j}] \rangle_{S_2}^2 \nonumber \\
     &\leq \sum_{i \geq 1}  \sum_{j \geq 1} \left(\frac{1}{\mu_{Z,i}}\right)^{\beta_Z} \langle C, \sqrt{\mu_{X,i}}e_{X,i} \otimes [e_{Z,j}] \rangle_{S_2}^2 = \|F\|_{\beta_Z}^2, \nonumber
\end{align} 
which concludes the proof.
\end{proof}
\noindent
The following theorem provides convergence guarantees for learning the CME, $F_*$.
\begin{theorem}[Theorem 4 \cite{meunier2024optimal}]\label{theo:cme_rate}
Let $g_{\xi}$ be a filter function with qualification $\rho \geq1$ used to build the estimator $\hat F_{\xi}$ on $\cD_1$ with Eq.~\eqref{eqn:vkrr_func}. Let Assumptions~\ref{asst:rkhs}, \eqref{asst:evdz} and \eqref{asst:embz} hold with  $0< p_Z \leq \alpha_Z \leq 1$. With $0 \leq \theta \leq 1$, if \eqref{asst:srcz} is satisfied with $\max\{\theta,\alpha_Z\} < \beta_Z \leq 2\rho$, we have, taking $\xi_m = \Theta \left(m^{-\frac{1}{\beta_Z + p_Z}}\right)$, that there is a constant $J > 0$ independent of $m \geq 1$ and $\tau \geq 1$ such that \[\left\|\hat{F}_{\xi} - F_*\right\|^2_{\theta} \leq \tau^2 J m^{-\frac{\beta_Z-\theta}{\beta_Z + p_Z}}\] is satisfied for sufficiently large $m \geq 1$ with $P^m$-probability not less than $1-4e^{-\tau}$. In particular, by Assumption~\eqref{asst:embz},
\[\left\|\hat{F}_{\xi} - F_*\right\|^2_{L_{\infty}(Z; \cH_X)} \leq A_Z^2\left\|\hat{F}_{\xi} - F_*\right\|^2_{\alpha_Z} \leq \tau^2 A_Z^2 J m^{-\frac{\beta_Z-\alpha_Z}{\beta_Z + p_Z}}.\]
\end{theorem}

\begin{lemma}[Lemma $25$ \cite{fischer2020sobolev}]\label{lma:supl} For $\lambda>0$ and $0 \leq \alpha \leq 1$, let $f_{\lambda, \alpha}:[0, \infty) \rightarrow \mathbb{R}$ be defined by $f_{\lambda, \alpha}(t)\doteq t^\alpha /(\lambda+t)$. Then,
$
\sup _{t \geq 0} f_{\lambda, \alpha}(t) \leq \lambda^{\alpha-1}.
$
\end{lemma}
In the remainder of this section, \textbf{we fix $H$ a separable Hilbert space}.
The following bound is a Bernstein-like concentration inequality for Hilbert space-valued random variables. It
can be deduced from Corollary~1, \cite{pinelis1986remarks}.

\begin{theorem}[Bernstein's Inequality]\label{theo:ope_con_steinwart}
Let $\xi_1,\dots,\xi_n$ be independent random variables with values in $H$, and assume that $\E[\xi_i]=0$ for all $i$. Suppose that there exist
constants $\tilde{\sigma},\tilde{L}>0$ such that, for every integer $m\ge 2$,
\[
\sum_{i=1}^n \E\|\xi_i\|^m_H \le \frac{m!}{2} \tilde{\sigma}^2 \tilde{L}^{m-2}.
\]
Then, for every $\tau >0$, with probability at least $1-2e^{-\tau}$,
\[
\left\|\frac1n\sum_{i=1}^n \xi_i\right\|
\le
\frac{\sqrt{2\tilde{\sigma}^2\tau}}{n}
+\frac{2\tilde{L}\tau}{n}.
\]
\end{theorem}

\begin{lemma}[Proposition 7, \cite{rudi2015less}] \label{lma:AB_norm}
Let A, B be two bounded positive semidefinite operators acting on $H$ and $\lambda>0$. Then,
$$
\left\|(A+\lambda \operatorname{Id}_{H})^{-1 / 2} B^{1 / 2}\right\|_{H \to H} \leq\left\|(A+\lambda \operatorname{Id}_{H})^{-1 / 2}(B+\lambda \operatorname{Id}_{H})^{1 / 2}\right\|_{H \to H} \leq(1-\beta)^{-1 / 2},
$$
when
$
\beta=\left\|(B+\lambda \operatorname{Id}_{H})^{-1 / 2}(B-A)(B+\lambda \operatorname{Id}_{H})^{-1 / 2}\right\|_{H \to H}<1.
$
\end{lemma}

\begin{proposition}\label{prop:op_1}
    Let $A,B$ be two compact self-adjoint positive semi-definite operators acting on $H$ and let $P$ be the orthogonal projection on $\overline{\cR(B)}$. If $PAP \leq B$, then for all $\delta > 0$, 
    $$
    P\left(B + \delta Id_{H}\right)^{-1}P \leq  P\left(A + \delta Id_{H}\right)^{-1}P.
    $$
    Furthermore, if $f \in \overline{\cR(B)}$ and $f \in \cR(A^{1/2})$, we have \(\langle f, B^{\dagger} f\rangle_{H} \leq \langle f, A^{\dagger} f\rangle_{H}.\)
\end{proposition}

\begin{proof}
For any $t,\alpha >0$ define $C_{t,\alpha}\doteq B + tP + \alpha P_{\perp}$. Then if $t(\alpha-\|A\|) \geq \|A\|^2$, we have $A \leq C_{t,\alpha}$. Indeed, for all $f \in H$,
$$
\begin{aligned}
&\langle f, (C_{t,\alpha}-A)f \rangle_{H} = \langle Pf + P_{\perp}f , (C_{t,\alpha}-A)(Pf + P_{\perp}f) \rangle_{H}\\
&= \langle f, Bf \rangle_{H}+t\langle Pf, Pf \rangle_{\cH}-\langle f, PAPf \rangle_{\cH} - 2\langle Pf, AP_{\perp}f \rangle_{H} + \alpha\langle P_{\perp}f, P_{\perp}f \rangle_{H} - \langle P_{\perp}f, AP_{\perp}f \rangle_{H}\\
&\ge t\|Pf\|^2_{H} -2\|A\|\|Pf\|_{H}\|P_{\perp}f\|_{H} + (\alpha-\|A\|)\|P_{\perp}f\|^2_{H}\\
&= t\|Pf\|^2_{H} - 2\|A\|\|Pf\|_{H}\|P_{\perp}f\|_{H} +\frac{\|A\|^2}{t}\|P_{\perp}f\|_{H}^2 - \frac{\|A\|^2}{t}\|P_{\perp}f\|_{H}^2+(\alpha-\|A\|)\|P_{\perp}f\|^2_{H}\\
&= \left(\sqrt{t}\|Pf\|_{H} - \frac{\|A\|}{\sqrt{t}}\|P_{\perp}f\|_{H}\right)^2 - \frac{\|A\|^2}{t}\|P_{\perp}f\|_{H}^2+(\alpha-\|A\|)\|P_{\perp}f\|^2_{H}\geq 0.
\end{aligned}
$$
where the last inequality follows from $t(\alpha-\|A\|) \geq \|A\|^2$. Since $B$ is compact self-adjoint positive semi-definite, it admits a decomposition
$
B = \sum_{i \geq 1} \omega_i b_i \otimes b_i,
$
where for all $i \geq 1$, $(\omega_i, b_i)$ are pairs of eigenvalues and eigenvectors of $B$ such that $\omega_i > 0$ and $\{b_i\}_{i \geq 1}$ forms a orthonormal basis of $\overline{\cR(B)}$. Therefore, on one hand,
$$
P(B+tP+\delta Id_{\cH})^{-1}P = P\left( \sum_{i \geq 1} \frac{1}{\delta + t + \omega_i} b_i \otimes b_i +\frac{1}{\delta} P_{\perp} \right)P = \sum_{i \geq 1} \frac{1}{\delta + t + \omega_i} b_i \otimes b_i,
$$
and on the other hand,
$$
P(C_{t,\alpha} + \delta Id_{\cH})^{-1}P = P\left( \sum_{i \geq 1} \frac{1}{\delta + t + \omega_i} b_i \otimes b_i + \frac{1}{\delta + \alpha}P_{\perp}\right)P = \sum_{i \geq 1} \frac{1}{\delta + t + \omega_i} b_i \otimes b_i.
$$
It follows that, for $t(\alpha-\|A\|) \geq \|A\|^2$, 
$$
P(B+tP+\delta Id_{H})^{-1}P=P(C_{t,\alpha} + \delta Id_{H})^{-1}P\le P(A + \delta Id_{H})^{-1}P.
$$ Let $t\to0^+$, the result follows: $P(B+\delta Id_{H})^{-1}P \le P(A + \delta Id_{H})^{-1}P.$ For the second part, let us consider $f \in \overline{\cR(B)}$. Then $Pf = f$ and
\[\langle f, (B+\delta Id_{H})^{-1}f \rangle_H \leq \langle f, (A+\delta Id_{H})^{-1}f \rangle_H,
\]
by the first part of the proposition. Under the assumption that $f \in \cR(A^{1/2})$, $\|(A^{1/2})^{\dagger}f\|_H < +\infty$ and taking the limit with $\delta \to 0^+$ gives the final result.
\end{proof}

\end{document}